\documentclass[letterpaper]{article}
\usepackage{fullpage}
\usepackage[utf8]{inputenc}

\usepackage{amsmath}
\usepackage{amsmath, graphicx,amssymb, hyperref, bbm, multirow, listings}
\usepackage{enumitem}
\usepackage{todonotes}
\usepackage{float}
\let\vec\mathbf
\usepackage{soul}
\usepackage{wrapfig}

\usepackage{algorithmic}
\usepackage{algorithm}

\usepackage{amsmath,bm}
\usepackage{amssymb}
\usepackage{mathtools}
\usepackage{amsthm}

\usepackage{hyperref}
\hypersetup{
  colorlinks,
  citecolor=blue,
  linkcolor=red,
  urlcolor=violet}
\usepackage{cleveref}

\crefname{equation}{}{}

\usepackage{subcaption}
\usepackage{booktabs}       

\usepackage[toc,page,header]{appendix}
\usepackage{minitoc}

\newtheorem{theo}{Theorem}[section]

\newtheorem{lem}{Lemma}[section]
\newtheorem{prop}{Proposition}[section]
\newtheorem{rem}{Remark}[section]
\newtheorem{cor}{Corollary}[section]

\newtheorem{nota}{Notation}[section]
\newtheorem{de}{Definition}[section]
\newtheorem{exa}{Example}[section]
\newtheorem{as}{Assumption}[section]
\newtheorem{alg}{Algorithm}[section]

\newcommand{\btheo}{\begin{theos}}
\newcommand{\bde}{\begin{de}}
\newcommand{\ble}{\begin{lem}}
\newcommand{\bpr}{\begin{props}}
\newcommand{\bno}{\begin{nota}}
\newcommand{\bex}{\begin{exa}}
\newcommand{\bcor}{\begin{cors}}
\newcommand{\spro}{\begin{proof}}
\newcommand{\bas}{\begin{as}}
\newcommand{\balg}{\begin{alg}}
\newcommand{\bremark}{\begin{remark}}

\newcommand{\etheo}{\end{theos}}
\newcommand{\ede}{\end{de}}
\newcommand{\ele}{\end{lem}}
\newcommand{\epr}{\end{props}}
\newcommand{\eno}{\end{nota}}
\newcommand{\eex}{\end{exa}}
\newcommand{\ecor}{\end{cors}}
\newcommand{\fpro}{\end{proof}}
\newcommand{\eas}{\end{as}}
\newcommand{\ealg}{\end{alg}}
\newcommand{\eremark}{\end{remark}}
\newcommand{\reals}{\mathbb{R}}

\newtheorem{theos}{Theorem}
\newtheorem{props}{Proposition}
\newtheorem{lems}{Lemma}
\newtheorem{cors}{Corollary}
\newtheorem{rems}{Remark}

\newtheorem{exas}{Example}
\newtheorem{algs}{Algorithm}
\newtheorem{asss}{Assumption}
\newtheorem{defns}{Definition}

\newcommand{\btheos}{\begin{theos}}
\newcommand{\etheos}{\end{theos}}
\newcommand{\brems}{\begin{rems}}
\newcommand{\erems}{\end{rems}}
\newcommand{\bprops}{\begin{props}}
\newcommand{\eprops}{\end{props}}
\newcommand{\bdes}{\begin{defns}}
\newcommand{\edes}{\end{defns}}
\newcommand{\blems}{\begin{lems}}
\newcommand{\elems}{\end{lems}}
\newcommand{\bcors}{\begin{cors}}
\newcommand{\ecors}{\end{cors}}
\newcommand{\bexs}{\begin{exas}}
\newcommand{\eexs}{\end{exas}}
\newcommand{\balgs}{\begin{algs}}
\newcommand{\ealgs}{\end{algs}}
\newcommand{\bass}{\begin{asss}}
\newcommand{\eass}{\end{asss}}
\newcommand{\bit}{\begin{itemize}}
\newcommand{\eit}{\end{itemize}}

\usepackage{xcolor}

\newcommand{\real}{\mathbb{R}}
\newcommand{\dual}{\vec{v}}
\newcommand{\dualmat}{\vec{V}}

\newcommand{\rectset}{\mathcal{Q}_\data}
\newcommand{\relu}[1]{\big( #1 \big)_+}

\newcommand{\ball}{\mathcal{B}}

\newcommand{\sign}{\text{sign}}

\newcommand{\diag}{\vec{D}}

 \DeclareMathOperator*{\argmin}{\arg\!\min} 
  \DeclareMathOperator*{\argmax}{\arg\!\max} 

\let\vec\mathbf
\usepackage{hyperref}
\usepackage{url}

\newcommand{\datascalar}{x}
\newcommand{\data}{\vec{X}}
\newcommand{\datavec}{\vec{x}}
\newcommand{\labelscalar}{y}
\newcommand{\labelvec}{\vec{y}}
\newcommand{\labelmat}{\vec{Y}}

\newcommand{\firstw}{\vec{w}^{(1)}}
\newcommand{\firstwb}{\bar{\vec{w}}^{(1)}}
\newcommand{\firstwh}{\hat{\vec{w}}^{(1)}}

\newcommand{\firstwmat}{\vec{W}^{(1)}}
\newcommand{\firstwmath}{\hat{\vec{W}}^{(1)}}

\newcommand{\secondw}{w^{(2)}}
\newcommand{\secondwb}{\bar{w}^{(2)}}
\newcommand{\secondwh}{\hat{w}^{(2)}}

\newcommand{\secondwvec}{\vec{w}^{(2)}}

\newcommand{\secondwvech}{\hat{\vec{w}}^{(2)}}

\newcommand{\secondwmat}{\vec{W}^{(2)}}

\newcommand{\dataf}{\mathcal{A}(\data)} 

\newcommand{\act}{\phi}
\newcommand{\actinf}{\pi}

\newcommand{\actt}[1]{\phi\big( #1 \big)}

\newcommand{\bias}{b}
\newcommand{\biasvec}{\vec{b}}

\newcommand{\weight}{\vec{w}}
\newcommand{\weightmat}{\vec{W}}

\DeclareUnicodeCharacter{2212}{-}
\let\vec\mathbf
\usepackage{bbm}
\usepackage{pifont}
\usepackage{enumitem}

\newcommand{\defn}{\ensuremath{\!:=}}

\usepackage[numbers,sort&compress]{natbib}

\title{The Convex Landscape of Neural Networks: Characterizing Global Optima and Stationary Points via Lasso Models}

\usepackage{authblk}

\author[1]{Tolga Ergen\thanks{The research for this paper was conducted while the author was affiliated with Stanford University.}}
\author[2]{Mert Pilanci}
\affil[1]{LG AI Research}

\affil[2]{Stanford Universtiy}

\date{}

\begin{document}
\doparttoc 
\faketableofcontents 

\maketitle

\begin{abstract}
Due to the non-convex nature of training Deep Neural Network (DNN) models, their effectiveness relies on the use of non-convex optimization heuristics.
Traditional methods for training DNNs often require costly empirical methods to produce successful models and do not have a clear theoretical foundation. In this study, we examine the use of convex optimization theory and sparse recovery models to refine the training process of neural networks and provide a better interpretation of their optimal weights. We focus on training two-layer neural networks with piecewise linear activations and demonstrate that they can be formulated as a finite-dimensional convex program. These programs include a regularization term that promotes sparsity, which constitutes a variant of group Lasso. We first utilize semi-infinite programming theory to prove strong duality for finite width neural networks and then we express these architectures equivalently as high dimensional convex sparse recovery models. Remarkably, the worst-case complexity to solve the convex program is polynomial in the number of samples and number of neurons when the rank of the data matrix is bounded, which is the case in convolutional networks. To extend our method to training data of arbitrary rank, we develop a novel polynomial-time approximation scheme based on zonotope subsampling that comes with a guaranteed approximation ratio. We also show that all the stationary of the nonconvex training objective can be characterized as the global optimum of a subsampled convex program. Our convex models can be trained using standard convex solvers without resorting to heuristics or extensive hyper-parameter tuning unlike non-convex methods. Due to the convexity, optimizer hyperparameters such as initialization, batch sizes, and step size schedules have no effect on the final model. Through extensive numerical experiments, we show that convex models can outperform traditional non-convex methods and are not sensitive to optimizer hyperparameters. The code for our experiments is available at \href{https://github.com/pilancilab/convex_nn}{\texttt{https://github.com/pilancilab/convex\_nn}}.
\end{abstract}


\section{Introduction}
Convex optimization has been a topic of interest due to several desirable properties that make it attractive for use in machine learning models. First and foremost, convex optimization problems in standard form are computationally tractable and typically admit a unique global optimum. In addition, standard convex optimization problems can be solved efficiently using well-established numerical solvers. This is in contrast to non-convex optimization problems, which can have multiple local minima and require heuristics to obtain satisfactory solutions. 

The distinction between convex and non-convex optimization is of great practical importance for machine learning problems. In non-convex optimization, the choice of optimization method and its internal parameters such as initialization, mini-batching, and step sizes have a significant effect on the quality of the learned model. This is in sharp contrast to convex optimization, for which these hyperparameters have no effect, and solutions are often unique and are determined by the data and the model, as opposed to being a function of the training trajectory and hyperparameters as in the case of neural networks. Moreover, convex optimization solutions can be obtained in a robust, reproducible, and transparent manner.

\subsection{Related work}
Existing works on convex neural networks \cite{bengio2006convex,bach2017breaking,fang2019convex} consider neural networks of infinite width to enable convexification over a set of measures. Hence, these results do not apply to finite width neural networks that are used in practice. 
\cite{bengio2006convex} proved that infinite width neural network training problems can be cast as a convex optimization problem with infinitely many variables. They also introduced an incremental algorithm that inserts a hidden neuron at a time by solving a maximization problem to obtain a linear classifier at each step. However, even though the algorithm may be used to achieve a global minimum for small datasets, it does not scale to high dimensional cases. In addition, \cite{bach2017breaking} investigated infinite width convex neural network training, however, did not provide a computationally tractable algorithm. In particular, strategies based on Frank-Wolfe \cite{bach2017breaking} require solving an intractable problem in order to train only a single neuron and do not optimize finite width networks. Similarly, \cite{fang2019convex} proved that in the infinite width limit, neural networks can be approximated as infinite dimensional convex learning models with an appropriate reparameterization. 

Sparse recovery models have become an essential tool in a wide variety of disciplines such as signal processing and statistics and forms the foundation of compressed sensing \cite{Tropp04, CandesTao05, Donoho06}. The key idea behind these models is to leverage the sparsity inherent in many data sources to enable more efficient and accurate processing. A notable technique used in sparse recovery is the l1-norm minimization, also known as the Lasso, which encourages sparsity in the solution vector \cite{Chen98,CandesTao06}. Further developments, such as group l1-norm minimization, extend this idea to incorporate structured sparsity, where groups of variables are either jointly included or excluded from the model \cite{yuan2006model}. These sparse recovery models provide an elegant framework for finding meaningful and parsimonious representations of complex data, thereby allowing more effective analysis and interpretation.

In contrast to existing work on convexifying neural networks, in this paper, we introduce a novel approach to derive exact finite dimensional convex program representations for finite width networks. Our characterization parallels sparse recovery models studied in the compressed sensing literature. The principal innovation lies in our analysis of hyperplane arrangements, an area of study originating from Cover's work on linear classifiers (Cover, 1965). Our results are applicable to any piecewise linear activations such as the Rectified Linear Unit (ReLU). 

\subsection{Our contributions} 
{A preliminary work on convex formulations of ReLU networks appeared in \cite{pilanci2020neural}. Our contributions over this work and other previous studies can be summarized as follows:
\begin{itemize}
\item We introduce a convex analytic framework to describe the training of two-layer neural networks with piecewise linear activations (including ReLU, leaky ReLU, and absolute value activation) as equivalent finite dimensional convex programs that perform sparse recovery. We prove the polynomial-time trainability of these architectures by standard convex optimization solvers when the data matrix has bounded rank, as is the case of Convolutional Neural Networks (CNNs).

\item In Theorem \ref{theo:subsampled_GD}, we prove that all of the stationary points of nonconvex neural networks correspond to the global optimum of a subsampled convex program. Therefore, we characterize all critical points of the nonconvex training problem which may be found via local heuristics as a global minimum of our subsampled convex program.

\item  We introduce a simple randomized algorithm to generate the hyperplane arrangements which are required to solve the convex program.  We prove a theoretical bound on the number of required samples (Theorem \ref{theo:effcient_sampling}) by relating it to sampling vertices of zonotopes. This approach significantly simplifies convex neural networks, since prior work assumed that the arrangement patterns are computed through enumeration. 

\item One potential limitation of solving our convex program exactly is the exponential worst-case complexity when applied to data with unbounded rank, as is often the case in Fully Connected (FC) neural networks.
In order to address this, we introduce an approximation algorithm (Theorem \ref{theo:lowrank_approx}) and prove strong polynomial-time approximation guarantees with respect to the global optimum. Combined with Theorem \ref{theo:effcient_sampling}, this enables a highly practical and simple method with strong guarantees.

\item We show that the optimal solution of the convex program is typically extremely sparse due to a small number of effective hyperplane arrangements in practical applications. We propose a novel hyperplane arrangement sampling technique utilizing convolutions and achieve substantial performance improvements in standard benchmarks.

\item Proposed convex models reveal novel interpretations of neural network models through diverse convex regularization mechanisms. The regularizers range from group $\ell_p$-norm to nuclear norm depending on the network architecture such as the connection structure and the number of outputs.

\item Our derivations are extended to various neural network architectures including convolutional networks, piecewise linear activation functions, vector outputs, arbitrary convex losses, and $\ell_p$-norm regularizers. We study vector output networks and derive exact convex programs for different regularizers.

\item We extend the analysis to several practically relevant variants of the NN training problem. In particular, we examine networks with bias terms, $\ell_p$-norm regularization of weights, and the interpolation regime.
\end{itemize}
}

\begin{table*}
  \caption{List of the neural network architectures that we study in this paper and the corresponding non-convex and convex training objectives. }
  \label{tab:models}
  \centering
   \resizebox{1\columnwidth}{!}{%
  \begin{tabular}{|c|c|c|c|c|}
    \toprule
    & \textbf{Model } & \textbf{Non-convex Objective} & \textbf{Convex Objective}  & \textbf{Result}     \\
    \cmidrule(r){1-5}
    \textbf{FC scalar output NN}  & $ f_{\theta}(\data)=\actt{\data\firstwmat}\secondwvec$ & \eqref{eq:twolayer_objective_generic}&   \eqref{eq:twolayerconvexprogram} & Theorem \ref{theo:mainconvex}  \\
   \textbf{Nonlinear CNN}  & $  f_{\theta}(\data)=\frac{1}{K} \sum_{k,j}  \act(\data_k\firstw_{j})\secondw_{j}$ & \eqref{eq:training_globalavg}& \eqref{eq:twolayerconvexprogram} & Section \ref{sec:cnn_avgpool} \\  
   \textbf{Linear CNN}  & $   f_{\theta,c}(\{\data_k\})= \sum_{k,j}\data_k \firstw_j \secondw_{jk}$ & \eqref{eq:CNN_linear}& \eqref{eq:CNN_linear_nuclear} & Section \ref{sec:cnn_linear} \\    
 \textbf{Circular linear CNN}  & $      f_{\theta,c}(\data)= \sum_{j}\data \firstwmat_j\secondwvec_j$ & \eqref{eq:linear_cnn}& \eqref{eq:linear_cnn_l1} & Section \ref{sec:circular_cnn_linear} \\    
    \textbf{FC vector output NN}  & $ f_{\theta}(\data)=\actt{\data\firstwmat}\secondwmat$ & \eqref{eq:twolayer_objective_generic_vector}&  \eqref{eq:twolayerconvexprogram_vector} & Theorem \ref{theo:mainconvex_vector}  \\
    \bottomrule
  \end{tabular}}
\end{table*}

\subsection{Notation}
We use uppercase and lowercase bold letters to denote matrices and vectors, respectively, throughout the paper. We use subscripts to index  entries (columns) of vectors (matrices). We use $\vec{I}_k$ for the identity matrix of size $k \times k$. We denote the set of integers from $1$ to $n$ as $[n]$. Moreover, $\|\cdot \|_{F}$ and $\| \cdot\|_{*}$ are Frobenius and nuclear norms and $\ball_p:=\{\vec{u} \in \mathbb{R}^d:\|\vec{u}\|_p\le 1\}$ is the unit $\ell_p$ ball. We also use $\mathbbm{1}[x\geq0]$ as an element-wise 0-1 valued indicator function. Furthermore, we use $\sigma_{max}(\cdot)$ to represent the maximum singular value of its argument. Finally, $\diag(\cdot)$ (or $\diag$) denotes a diagonal matrix. We use $\mathrm{Conv}(S)$ to denote the convex hull of a subset $S\subseteq\reals^{d}$.
\subsection{Preliminaries}
\label{sec:preliminaries}
We consider a two-layer neural network architecture $f_{\theta}(\data)\,:\,\reals^{n\times d}\rightarrow\reals^{n\times C}$ with $m$ hidden neurons and $C$ outputs as follows
\begin{align}
    f_{\theta}(\data) :=  \act(\data \firstwmat) \secondwmat  \,, \label{eq:twolayer_network}
\end{align}
where $\data \in \real^{n \times d}$ is a data matrix containing $n$ training samples in $\reals^d$, $\firstwmat \in \mathbb{R}^{d \times m}$ and $\secondwmat \in \mathbb{R}^{m \times C}$ are the hidden and output layer weights respectively. Here, $\act(\cdot)$ is the non-linear activation function. We consider positive homogeneous activations of degree one, i.e., $\act(t\datascalar)=t\act(\datascalar),\, \forall t \in \mathbb{R}_+$ such as ReLU, leaky ReLU, and absolute value. In addition, we denote all trainable parameters by $\theta:=\{\firstwmat,\secondwmat\}$ and the corresponding parameter space $\Theta:=\{\theta: \firstwmat \in \mathbb{R}^{d \times m},\,\secondwmat \in \mathbb{R}^{m \times C}\}$. 

Due to the nondifferentiability of the piecewise linear activations, we also review the definition of the Clarke subdifferential \cite{clarke1975generalized} of a given function $f$. Let $ D \subset \mathbb{R}^d$ be the set of points at which \( f \) is differentiable. We assume that \( D \) has (Lebesgue) measure \( 1 \), meaning that \( f \) is differentiable \emph{almost everywhere}.
The Clarke subdifferential of $f$ at $\vec{x}$ is then defined as
\begin{align*} 
  \partial_C f(\vec{x}) = \mathrm{Conv} \left\{\lim_{k \to \infty} \nabla f(\vec{x}_k) \mid \lim_{k \to \infty }\vec{x}_k \to \vec{x},\,\vec{x}_k \in D\right\}.
\end{align*}
Then, we say that $\vec{x} \in \mathbb{R}^d$ is Clarke
stationary with respect to $f$ if $\vec{0} \in   \partial_C f(\vec{x})$.

\begin{figure*}[t]
\centering
\captionsetup[subfigure]{oneside,margin={1cm,0cm}}
	\begin{subfigure}[t]{0.32\textwidth}
	\centering
	\includegraphics[width=1\textwidth]{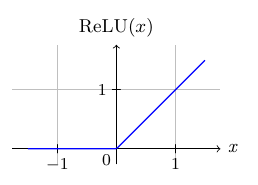}
	\caption{$\kappa=0$\centering} \label{fig:relu}
\end{subfigure} \hspace*{\fill}
	\begin{subfigure}[t]{0.32\textwidth}
	\centering
	\includegraphics[width=1\textwidth]{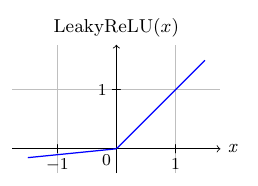}
	\caption{$\kappa=0.1$\centering} \label{fig:leaky}
\end{subfigure} \hspace*{\fill}
	\begin{subfigure}[t]{0.32\textwidth}
	\centering
	\includegraphics[width=1\textwidth]{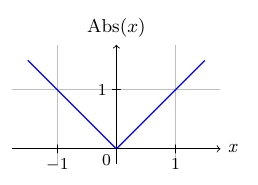}
	\caption{$\kappa=-1$\centering} \label{fig:abs}
\end{subfigure} \hspace*{\fill}
\caption{Examples of piecewise linear activations $\act$ satisfying the definition in \eqref{eq:activationdefn}.
 }\label{fig:activation}
\end{figure*}

%
Given a matrix of labels $\labelmat \in \real^{n \times C}$, the regularized training problem for the network in \eqref{eq:twolayer_network} is given by
\begin{align*} 
&\min_{\theta \in \Theta} \mathcal{L}(f_{\theta}(\data),\labelmat)+\beta\mathcal{R}(\theta)\,,
\end{align*}
where $\mathcal{L}(\cdot,\cdot)$ is an arbitrary convex loss function, $\mathcal{R}(\cdot)$ is a regularization term, and $\beta>0$ is the corresponding regularization parameter. We focus on the standard supervised regression/classification framework with conventional squared $\ell_2$-norm, i.e., weight decay, regularization denoted as $\mathcal{R}(\theta)=\frac{1}{2} (\|\firstwmat\|_F^2+\|\secondwmat\|_F^2)$. We consider the following family of piecewise linear activation functions
\begin{align}\label{eq:activationdefn}
\act(\datascalar):=\begin{cases}\datascalar&\text{ if } x\geq 0\\
\kappa\datascalar&\text{ if } x< 0\end{cases}
\end{align}
for some fixed scalar $\kappa < 0.5$. We note that the definition above includes a set of commonly used activation functions including ReLU, Leaky ReLU, and absolute value (see Figure \ref{fig:activation}). 
With these definitions, we have the following training problem
\begin{align}\label{eq:twolayer_objective_generic}
     p^*:=\min_{\theta \in \Theta} \mathcal{L}(f_{\theta}(\data),\labelmat)+\frac{\beta}{2} (\|\firstwmat\|_F^2+\|\secondwmat\|_F^2)\,.
\end{align}

Notice that the objective function above is highly non-convex due to the nested minimization of first and second layer weights and the composition of the nonlinearity $\act$ with the loss function $\mathcal{L}$. 

One of our key contributions is in developing an alternative parameterization of the same neural network and the corresponding training objective that enables significantly more efficient optimization. 

To illustrate the challenges involved in optimizing the original formulation, we will examine the combinatorial nature of the original parameterization and why straightforward attempts to convexify the objective function fail. Consider rewriting the non-convex optimization problem \eqref{eq:twolayer_objective_generic} with ReLU activations and a scalar output, i.e., $\kappa=0$ and $C=1$, via enumerating all activation patterns of all ReLU neurons as
\begin{align} \label{eq:nonconvex_combinatorial}
    &\min_{\vec{d}_j \in \mathcal{H}_{d_j}}\min_{\theta \in \Theta} \mathcal{L}\left(\sum_{j=1}^m \left[\vec{d}_j \odot (\data \firstw_j)\right]\secondw_j,\labelvec \right) +\frac{\beta}{2} (\|\firstwmat\|_F^2+\|\secondwvec\|_2^2)\,,
\end{align}
where $\odot$ denotes element-wise multiplication, and $\mathcal{H}_{d_j}:=\{\vec{d}_j \in \{0,1\}^n\, :\, (2\vec{d}_j-\vec{1}) \odot (\data \firstw_j)\geq 0\}$ is a discrete parameterization of the piecewise parameterization of the set of ReLU activation patterns. A brute-force search would involve enumerating all possible combinations of $m$ different length-$n$ binary vectors, $\{\vec{d}_j\}_{j=1}^m$, which takes exponential time in the number of neurons $m$ and the number of samples $n$. In fact, the best known algorithm for directly optimizing the training objective \eqref{eq:twolayer_objective_generic} with ReLU activations is a brute-force search over all possible piecewise linear regions of ReLU activations of $m$ neurons and sign patterns for the output layer, which has complexity $\mathcal{O}(2^m n^{dm})$ (see Theorem 4.1 in \cite{arora2018understanding}). In fact, known algorithms for approximately learning $m$ hidden neuron ReLU networks have complexity $\mathcal{O}(2^{\sqrt{m}})$ (see Theorem 5 of \cite{goel17a}) due to similar combinatorial hardness with respect to the number of neurons. Since the number of hidden neurons in practical networks is typically in the order of hundreds or thousands, existing methods are computationally intractable even in small feature dimensions, e.g., $d=2$.

\begin{figure*}[t]
        \centering
            \centering
            \includegraphics[width=\textwidth,height=0.55\textwidth]{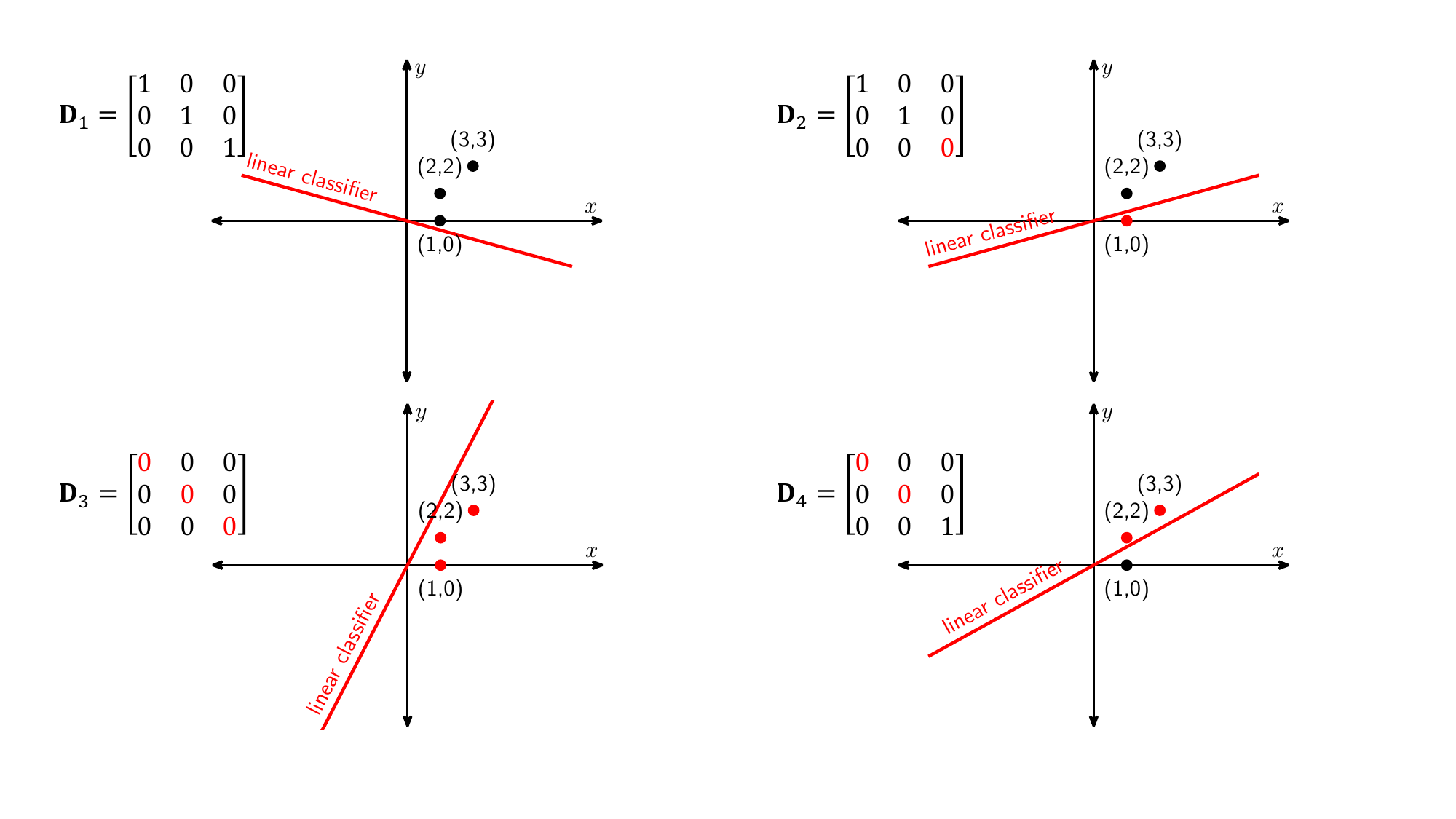}
            \caption{Two dimensional illustration of all possible hyperplane arrangements that determine the diagonal matrices $\{\diag_i\}_{i=1}^P$ for a toy dataset with dimensions $n=3$, $d=2$. In this example, we consider the ReLU activation, i.e., $\act(\datascalar)=\max\{\datascalar,0\}$. {Note that the hyperplanes pass through the origin, as there is no bias term included in the neurons.}}\label{fig:hyperplanes}
    \end{figure*}

\begin{figure*}[t]
        \centering
            \centering
            \includegraphics[width=\textwidth,height=0.5\textwidth]{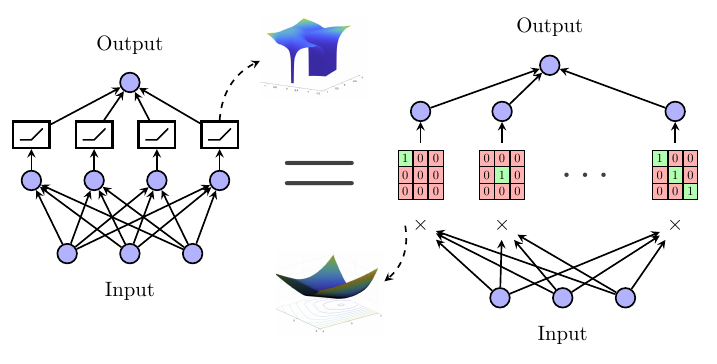}
            \caption{An illustration of the equivalence between the non-convex ReLU network (\textbf{left}) and equivalent convex model \eqref{eq:twolayerconvexprogram} (\textbf{right}) along with their corresponding training losses.}\label{fig:convex_model}
    \end{figure*}

\section{An equivalent convex program for two-layer neural networks} \label{sec:main_results}
We first introduce the notion of hyperplane arrangements of the data matrix $\data$ and then introduce an exact convex program as an alternative to the non-convex problem \eqref{eq:twolayer_objective_generic}.
Next, we note that piecewise linear activations can be equivalently represented via linear inequality constraints when their activation patterns are fixed since
\begin{align*}
    \act(\data\firstw)=\diag \data \firstw \iff (2\diag-\vec{I}_n)\data \firstw \geq 0,
\end{align*}
where $\diag \in \mathbb{R}^{n \times n}$ is a fixed diagonal matrix of activation patterns  defined as
\begin{align} \label{eq:diagonal_matrix}
\diag_{ii} := \begin{cases} 1 & \mbox{ if } \datavec_i^T\firstw \geq 0\\
\kappa & \mbox{ otherwise}  \end{cases}.
\end{align}
It can be seen that each activation pattern corresponds to a hyperplane arrangement of the data matrix $\data$. We now enumerate all such distinct diagonal matrices that can be obtained for all possible $\firstw \in \real^d$, and denote them as $\diag_1,...,\diag_P$ (see Figure \ref{fig:hyperplanes} for the visualization of a two-dimensional case). Here, $P$ denotes the number of regions in a partition of $\real^d$ by hyperplanes passing through the origin, and are perpendicular to the rows of $\data$. It is well known that 
\begin{align}\label{eq:complexity}
  P\le 2 \sum_{k=0}^{r-1} {n-1 \choose k}\le 2r\Big(\frac{e(n-1)}{r}\Big)^r \,  
\end{align}
 for $r\le n$, where $r:=\mbox{rank}(\data)$ \cite{stanley2004introduction,cover1965geometrical} (see Appendix \ref{sec:hyperplane_arrangments_appendix}). Thus, for a given data matrix of bounded rank, the number of hyperplane arrangements $P$ is upper-bounded by an expression that is polynomial in both $n$ and $d$. In Section \ref{sec:convolutions}, we show that convolutional networks used in practice have data rank bounded by a small constant that is equal to the spatial length of a filter.
 
 A crucial observation is that the number $P$ corresponds to the number of distinct activation patterns generated by $\textbf{only a single neuron}$ on the training data. This number is polynomial in $n$ when the rank of the data matrix is bounded by a constant. On the other hand, optimizing the non-convex formulation \eqref{eq:twolayer_objective_generic} or \eqref{eq:nonconvex_combinatorial} requires searching over all activation patterns of $m$ neurons jointly, which results in \textbf{computational complexity exponential in $m$} as discussed in Section \ref{sec:preliminaries}.

 With this observation, we next introduce an exact polynomial-time solvable convex program that solves \eqref{eq:twolayer_objective_generic} optimally by using $dP$ variables.
\begin{theo}
\label{theo:mainconvex}
Given a scalar output network, i.e., $C=1$, consider the convex program
\begin{align}
&p_{\mathrm{cvx}}=\min_{ \weight  \in \mathcal{C}(\data)}\,  \mathcal{L}(\dataf\weight ,\labelvec)  +\beta \sum_{i=1}^{2P}\|\weight_i\|_{2} \label{eq:twolayerconvexprogram}   
\end{align}
where $\vec{w}:=[\vec{w}_1^T,\ldots, \vec{w}_{2P}^T]^T \in \mathbb{R}^{2dP}$, and let $\vec{w}^*$ be the minimum-norm optimal solution.
We have $p^*=p_\mathrm{cvx}$ using the equivalent formulation in Lemma \ref{lemma:scaling} when $m\ge m^*:=\sum_{\tiny i=1}^{2P} \mathbbm{1}[ \weight_i^*\neq 0]$, i.e., when the number of neurons exceeds the critical threshold $m^*$. Here, $\dataf \in \mathbb{R}^{n \times 2dP}$ and the constraint set of the convex program denoted as $\mathcal{C}(\data)$ are defined as
\begin{align*}
& \mathcal{C}(\data):=\left\{\weight\in \mathbb{R}^{ 2dP} \;: \; 
   (2\diag_i-\vec{I}_n)\data\begin{bmatrix}\weight_i&\weight_{i+P}\end{bmatrix} \geq 0,  \forall i \in [P] \right\} \\
&\dataf:= \begin{bmatrix}  \diag_1 \data  & \ldots & \diag_P\data& -\diag_1 \data  & \ldots & -\diag_P \data
  \end{bmatrix}.
\end{align*}

\end{theo}
Theorem \ref{theo:mainconvex} shows that a standard two-layer neural network with piecewise linear activations can be described as a convex mixture of locally linear models $\{\diag_i\data\weight_i\}_{i=1}^{P}$ and $\{-\diag_i\data\weight_{i+P}\}_{i=1}^{P}$, where the fixed diagonal matrices $\{\diag_i\}_{i=1}^P$ control the data samples interacting with the local model as fixed gates. Therefore, optimal two-layer networks can be viewed as sparse convex mixtures of locally linear functions, where sparsity is enforced via the group Lasso regularization. 

Next, we prove that Clarke stationary points correspond to the global optimum of a subsampled convex programs studied in the previous section. This result explains the neural network models found by first order optimization methods such as (Stochastic) Gradient Descent, which converge to a neighborhood of a stationary point.

\begin{theo}\label{theo:subsampled_GD}
Suppose that $\theta$ is a Clarke stationary point of the nonconvex training objective in \eqref{eq:twolayer_objective_generic}. Then, $\theta$ corresponds to a global optimum of the subsampled form of  the convex program in \eqref{eq:twolayerconvexprogram} with $\tilde{P}=m$ arrangement patterns.
\end{theo}
Theorem \ref{theo:subsampled_GD} implies that any local minimum of the nonconvex training objective in \eqref{eq:twolayer_objective_generic} can be characterized as a global minimum of a subsampled form of the convex program in \eqref{eq:twolayerconvexprogram}, for which the sampling procedure is in Section \ref{sec:efficient_arrangement}. Therefore, we can characterize all stationary points of the nonconvex training objective in \eqref{eq:twolayer_objective_generic} by sampling the arrangement patterns for the convex optimization problem in \eqref{eq:twolayerconvexprogram}.

Importantly, the proposed convex program trains two-layer neural networks \emph{optimally}. In contrast, local search heuristics such as backpropagation may converge to suboptimal solutions, which is illustrated with numerical examples in Section \ref{sec:numerical}.  To the best of our knowledge, our results provide the first polynomial-time algorithm to train optimal neural networks when the data rank (or feature dimension) is fixed.

In the light of the results above, a weight decay, i.e.,  squared $\ell_2$ norm, regularized two-layer neural network with piecewise linear activations is a high-dimensional feature selection method that seeks sparsity. More specifically, training the non-convex model can be considered as transforming the data to the higher dimensional feature matrix $\dataf$, and then seeking a parsimonious convex model through the group Lasso regularization. The optimal model is very concise due to the group sparsity induced by the sum of Euclidean norms. This fact, however, is not obvious from the non-convex formulations of these neural network models.

The following result shows that one can construct a classical two-layer network as in \eqref{eq:twolayer_network} from the solution of the convex program \eqref{eq:twolayerconvexprogram}.
\begin{prop}
\label{prop:mapping}
An optimal solution to the non-convex problem in \eqref{eq:twolayer_objective_generic}, i.e., denoted as $\{{\firstw_j}^*,{\secondw_j}^*\}_{j=1}^{m^*}$, can be constructed from the optimal solution to the convex program as follows
\begin{align*}
    ({\firstw_{j_{i}}}^*,{\secondw_{j_{i}}}^*) = \begin{cases} \left(\frac{\weight^{*}_i}{\sqrt{\|\weight^{*}_i\|_2}}, \sqrt{\|\weight^{*}_i\|_2}\right) \,\quad  &\mbox{  if  }  \weight_i^{*}\neq 0 \mbox{ and } i\leq P\\
    \left(\frac{\weight^{*}_i}{\sqrt{\|\weight^{*}_i\|_2}}, -\sqrt{\|\weight^{*}_i\|_2}\right) \,\quad  &\mbox{  if  }  \weight_i^{*}\neq 0 \mbox{ and } i> P
    \end{cases},
\end{align*}
where $\{\weight_i^*\}_{i=1}^{2P}$ are the optimal solutions to \eqref{eq:twolayerconvexprogram}, and $j_{i} \in [\vert\mathcal{J}\vert]$ given the definitions $\mathcal{J}:=\{i\;:\; \|\weight_{i}\|> 0\}$.
\end{prop}
\begin{rem}\label{rem:complexity}
Theorem \ref{theo:mainconvex} shows that two-layer networks with $m$ hidden neurons and the activation $\act$ can be globally optimized via the second order cone program \eqref{eq:twolayerconvexprogram} with $2dP$ variables and $2nP$ linear inequalities where  $P \leq 2r\Big(\frac{e(n-1)}{r}\Big)^r$, and $r=\mbox{rank}(\data)$. The computational complexity is at most $\mathcal{O}\Big(d^3r^3\big(\frac{n}{r}\big)^{3r}\Big)$ using standard interior-point solvers. For fixed rank $r$ (or dimension $d$), the complexity is polynomial in $n$ and $m$, which is an exponential improvement over the state of the art \cite{arora2018understanding,bienstock2018principled}. However, for fixed $n$ and $\mbox{rank}(\data)=d$, the complexity is exponential in $d$, which can not be improved unless $\mathrm{P}=\mathrm{NP}$ even for $m=2$ \cite{boob2018complexity}. Note that the convex program and the non-convex problem differ in terms of the hardness of the optimization problem they present. While the non-convex problem has fewer decision variables, it does not have a known systematic method for solving it 
 (apart from the one presented in this work) or verifying the optimality of a given solution. In contrast, the convex program has a larger number of decision variables, but it can be solved to global optimality and the optimality of any candidate solution can be checked. Thus, the convex program provides a trade-off between the difficulties of high-dimensionality and non-convexity.
\end{rem}
\begin{rem}\label{rem:active_set_heuristic}  Popular non-convex heuristics such as gradient descent and variants applied to the non-convex problem \eqref{eq:twolayer_objective_generic} can be viewed as local active set solvers for the convex program \eqref{eq:twolayerconvexprogram}. In this active set strategy, only a small subset of variables are maintained in the current solution, which corresponds to a small subset of hyperplane arrangements, i.e., column blocks of $\hat \data$. The variables in the active set solver enter and exit the active set as the ReLU activation patterns change.
\end{rem}
%

The proofs of the theorems and other claims (including Theorem \ref{theo:mainconvex}) can be found in Supplementary Material.

To gain a better understanding of the convex program \eqref{eq:twolayerconvexprogram} and the resulting convex ReLU neural network model, we will next consider a toy example. This example will provide insight into the underlying mechanisms and allow for a more intuitive interpretation of the equivalent non-convex neural network model.

\bex \label{example3by2}Let us consider ReLU activations and the training data matrix 
\begin{align*}
    \data=\begin{bmatrix} \datavec_1^T\\\datavec_2^T\\\datavec_3^T\end{bmatrix}=\begin{bmatrix} 2&2\\3&3\\1&0\end{bmatrix}.
\end{align*}
Even though there exist $2^3=8$ distinct binary sequences of length $3$, the number of hyperplane arrangements in this case is $4$ as shown in Figure \ref{fig:hyperplanes}. Considering an arbitrary label vector $\labelvec \in \mathbb{R}^3$ and the squared loss, we formulate the convex program in \eqref{eq:twolayerconvexprogram} as follows
\begin{align*}
    &\min_{\{\weight_i\}_{i=1}^6} \frac{1}{2}\left\| f_\weight(\data)-\labelvec \right\|_2^2+ \beta \sum_{i=1}^6 \|\weight_i\|_2 \\
    &\mbox{s.t. } \datavec_i^T [\weight_1\; \weight_4] \geq 0,\, i=1,2,3 \\
&\datavec_i^T [\weight_2\; \weight_5] \geq 0,\,i=1,2,~~~\,\datavec_3^T [\weight_2\; \weight_5] \leq 0 \\
&\datavec_i^T [\weight_3\; \weight_6] \leq 0,\,i=1,2,~~~\,\datavec_3^T [\weight_3\; \weight_6] \geq 0,
\end{align*}
where
\begin{align*}
   f_\weight(\data)&:= \begin{bmatrix} \datavec_1^T\\\datavec_2^T\\\datavec_3^T\end{bmatrix} (\weight_1-\weight_4)+\begin{bmatrix} \datavec_1^T\\\datavec_2^T\\\vec{0}^T\end{bmatrix} (\weight_2-\weight_5)+\begin{bmatrix} \vec{0}^T\\\vec{0}^T\\\datavec_3^T\end{bmatrix} (\weight_3-\weight_6)\\
   &=\diag_1\data(\weight_1-\weight_4)+\diag_2\data(\weight_2-\weight_5)+\diag_3\data(\weight_3-\weight_6),
\end{align*}
provided that 
\begin{align*}
    \diag_1=\begin{bmatrix}
        1 & 0 & 0 \\
        0 & 1 & 0 \\
        0 & 0 & 1
    \end{bmatrix},\;
    \diag_2=\begin{bmatrix}
        1 & 0 & 0 \\
        0 & 1 & 0 \\
        0 & 0 & 0
    \end{bmatrix},\;
    \diag_2=\begin{bmatrix}
        0 & 0 & 0 \\
        0 & 0 & 0 \\
        0 & 0 & 1
    \end{bmatrix}.
\end{align*}
\eex
Interestingly, we obtain a convex programming description of the neural network model that is interpretable: we are looking for a group sparse model to explain the response $\vec{y}$ via a convex mixture of linear models. To give an example, the linear term $\weight_2-\weight_5$ is responsible for predicting on the subset $\{ \datavec_1, \datavec_2\}$ of the dataset, and the linear term $\weight_3-\weight_6$ is responsible for predicting on the subset $\{\datavec_3\}$ of the dataset, etc. Due to the regularization term $\sum_{i=1}^6 \|\weight_i\|_2$, only a few of these linear terms will be non-zero at the optimum, which shows a strong bias towards simple solutions among all piecewise linear models. Here, we may ignore the arrangement that corresponds to all-zeros since it does not contribute to the objective. It is important to note that the above objective is equivalent to the non-convex neural network training problem. Although the non-convex training process given in \eqref{eq:twolayer_objective_generic} is hard to interpret, the equivalent convex optimization formulation shows the structure of the optimal neural network through a fully transparent convex model.

\begin{rem}
In general, we expect the number of hyperplane arrangements $P$ to be small when the data matrix is of small rank as in Figure \ref{fig:hyperplanes}. In Section \ref{sec:rank_approx}, we show that a similar result applies to near low-rank matrices which are frequently encountered in practice: a relatively small number of arrangement patterns is sufficient to approximate the global optimum up to a small relative error.
\end{rem}
\subsection{Networks with a bias term in hidden neurons}\label{sec:network_bias}
We now modify the neural network architecture in \eqref{eq:twolayer_network} as
\begin{align*}
    f_{\theta}(\data)=\act(\data \firstwmat+\vec{1}\biasvec^T)\secondwmat,
\end{align*}
where $\biasvec \in \mathbb{R}^m$ denotes the trainable bias vector. For this architecture, the training problem is the same with \eqref{eq:twolayer_objective_generic} except $\theta=\{\firstwmat, \secondwmat,\biasvec\}$.

\begin{cor}
\label{cor:mainconvex_bias}
As a result of Theorem \ref{theo:mainconvex}, the non-convex training problem with bias term can be cast as a finite dimensional convex program as follows
\begin{align*}
\min_{ \theta_c  \in \mathcal{C}(\data)}\,  \mathcal{L}(f_{\theta_c}(\data) ,\labelvec)  +\beta \sum_{i=1}^{2P}\big\|[\weight_i;\bias_i]\big\|_{2}  , 
\end{align*}
where $\theta_c:=\{\weight,\biasvec \}$. Moreover, $f_{\theta_c}(\data) ,\labelvec)$ and $\mathcal{C}(\data)$ are defined as
\begin{align*}
&f_{\theta_c}(\data) = \sum_{i=1}^P \diag_i((\data\weight_i+\vec{1}\bias_i)- (\data\weight_{i+P}+\vec{1}\bias_{i+P}))\\
& \mathcal{C}(\data):=\big\{\weight\in \mathbb{R}^{ 2dP},\biasvec \in \mathbb{R}^{2P} : (2\diag_i-\vec{I}_n)\left(\data\begin{bmatrix}\weight_i&\weight_{i+P}\end{bmatrix} +\vec{1}\begin{bmatrix}\bias_i & \bias_{i+P} \end{bmatrix}\right)\geq 0,  \forall i \in [P] \big\}.
\end{align*}
\end{cor}

\begin{rem}%
We note that including bias may improve the expressive power of the neural network \eqref{eq:twolayer_network} in small feature dimensions. This operation corresponds to augmenting a column of all-ones to the data matrix, which implies a slight increase in the number of hyperplane arrangements. The rank of the data matrix increases from $r$ to at most $r+1$, therefore, the new upperbound on the number of arrangements \eqref{eq:complexity} is obtained by simply replacing $r$ with $r+1$. As an example, in Figure \ref{fig:hyperplanes}, $(\firstw,\bias)=([1;1],-5)$ can separate $\datavec_1$ and $\datavec_2$, therefore we obtain an additional hyperplane arrangement and a corresponding variable vector in the convex program.
\end{rem}

\subsection{Convex duality of two-layer neural networks}\label{sec:convex_duality}
In this section, we provide a high-level overview of the mathematical proof technique that we use to obtain the convex program. It is worth noting that these results are of independent interest, as they are not solely motivated by global optimization of neural networks. We start with convex duality for \eqref{eq:twolayer_objective_generic}, which is essential for deriving the convex program in Theorem \ref{theo:mainconvex}.

\begin{figure}[t!]
        \centering
            \centering
            \includegraphics[width=0.5\textwidth,height=0.3\textwidth]{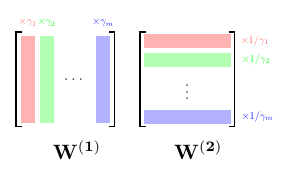}
            \caption{An illustration of the scaling technique in Lemma \ref{lemma:scaling}. This is instrumental in obtaining the $\ell_1$-norm penalized problem \eqref{eq:twolayer_objective_generic_l1}, which leads to a strong dual formulation.}\label{fig:scaling}
    \end{figure}

Since the piecewise linear activation $\act$ is a positive homogeneous function of degree one, we can apply a rescaling (see Figure \ref{fig:scaling}) to equivalently state the problem in \eqref{eq:twolayer_objective_generic} as an $\ell_1$-norm minimization problem.

\begin{lem}\label{lemma:scaling}
The problem in \eqref{eq:twolayer_objective_generic} can be equivalently formulated as the following $\ell_1$-norm minimization problem
\begin{align}\label{eq:twolayer_objective_generic_l1}
     p^*:=\min_{\theta \in \Theta_s} \mathcal{L}(f_{\theta}(\data),\labelvec)+\beta \|\secondwvec\|_1 \,, 
\end{align}
where $\Theta_s:=\{\theta \in \Theta: \|\firstw_j\|_2\leq 1,\, \forall j \in [m]\}$.
\end{lem}

We note that is important to obtain an $\ell_1$ regularized form of the non-convex problem in order to obtain strong duality. It can be easily verified that a straightforward application of Lagrange duality applied to the original weight-decay regularized objective does not lead to a strong dual.

We now use Lemma \ref{lemma:scaling} to obtain the convex dual form of \eqref{eq:twolayer_objective_generic}. We first take the dual of \eqref{eq:twolayer_objective_generic_l1} with respect to $\secondwvec$ and then change the order of min-max to obtain the following dual problem
\begin{align}\label{eq:twolayer_dual_generic}
        &p^*\geq d^* :=\max_{\vec{\dual}}\min_{\substack{ \theta \in \Theta_s}} - \mathcal{L}^*(\dual)   \\ \nonumber
        &\text{s.t. } \left\vert \dual^T \actt{\data \firstw_j} \right\vert\leq \beta,\, \forall j \in [m],
\end{align}
where $\mathcal{L}^*$ is the Fenchel conjugate function defined as \cite{boyd_convex}
\begin{align*}
\mathcal{L}^*(\dual) := \max_{\vec{z}}~ \vec{z}^T \dual - \mathcal{L}(\vec{z},\vec{y})\,.
\end{align*}
Since $\min_x\max_y f(x,y) \geq \max_y\min_x f(x,y)$, \eqref{eq:twolayer_dual_generic} is a lower bound for \eqref{eq:twolayer_objective_generic_l1}. However, at this point it is not clear whether the lower-bound is tight, i.e., $p^* = d^*$. 

Using the dual characterization in \eqref{eq:twolayer_dual_generic}, we first find a set of hidden layer weights via the optimality conditions and active constraints of \eqref{eq:twolayer_dual_generic}. We then prove strong duality, i.e., $p^*=d^*$, to verify the optimality of the hidden layer weight found via the dual problem. A complete proof of this result can be found in Appendix \ref{sec:appendix_convexduality}.

\begin{algorithm}[!t]
		\caption{\texttt{Polynomial-time convex neural network training algorithm}}
		\begin{algorithmic}[1]
        \STATE{Set the desired rank $k$ based on the bound in \eqref{eq:rank_bound}}
        \STATE{Compute the rank-$k$ approximation of the data matrix: $\hat{\data}_k$}
        \STATE{Set the number of arrangements to be sampled via Theorem \ref{theo:effcient_sampling}: $\tilde P$ }
        \STATE{Sample hyperplane arrangements from $\hat{\data}_k$: $\{\diag_{i}^k\}_{i=1}^{\tilde P}$}
        \STATE{Solve the convex training problem in \eqref{eq:twolayerconvexprogram} using the original data $\data$ and rank-$k$ arrangements $\{\diag_{i}^k\}_{i=1}^{\tilde P}$}
		\end{algorithmic}\label{alg:convex_rank}	
	\end{algorithm}

\section{Scalable optimization of the neural network convex program}\label{sec:rank_approx} 
Notice that the worst-case computational complexity to solve the convex program in Theorem \ref{theo:mainconvex} is exponential in the feature dimension $d$ for full-rank training data as detailed in Remark \ref{rem:complexity}. Therefore, globally optimizing the training objective \eqref{eq:twolayerconvexprogram} may not be feasible for large $d$.

To avoid the complexity of enumerating exponentially many hyperplane arrangements and to effectively scale to high-dimensional datasets, we consider a low-rank approximation of the data to approximate the arrangements and subsequently obtain an approximation of \eqref{eq:twolayer_objective_generic}. We denote the rank-$k$ approximation of $\data$ as $\hat{\data}_k$ such that $\|\data-\hat{\data}_k\|_2 \leq \sigma_{k+1}$, where $\sigma_{k+1}$ is the $(k+1)^{th}$ largest singular value of $\data$. Then, we have the following result.

\begin{theo}\label{theo:lowrank_approx}
Consider the following variant of the convex program \eqref{eq:twolayerconvexprogram} with rank-$k$ approximated hyperplane arrangements 
\begin{align}
      \weight^{(k)}&\in \argmin_{ \weight  \in \mathcal{C}(\hat{\data}_k)}\,  \mathcal{L}\left(\sum_{i=1}^{\hat{P}}\diag_i^k\data(\weight_i-\weight_{i+\hat{P}}) ,\labelvec \right)  +\beta \sum_{i=1}^{2\hat{P}}\|\weight_i\|_{2}  ,
      \label{eq:convex_program_lowrank}
\end{align}
where $\{ \diag_i^k\}_{i=1}^{\hat P}$ denotes the set of hyperplane arrangements generated by the rank-$k$ approximation $\hat{\data}_k $. Let us define $p_{\mathrm{cvx}-k}$ as the value of the non-convex objective \eqref{eq:twolayer_objective_generic} evaluated at any minimizer $\weight^{(k)}$ defined above. Then, given an $L$-Lipschitz convex loss $\mathcal{L}(\cdot,\labelvec)$ and an $R$-Lipschitz activation function $\phi(\cdot)$, we have the following approximation guarantee
\begin{align}\label{eq:rank_bound}
    p^*\leq p_{\mathrm{cvx}-k} \leq p^*  \left(1+\frac{L R\sigma_{k+1}}{\beta}\right)^2.
\end{align}
\end{theo}


\begin{rem} \label{rem:rank_approx}
Theorem \ref{theo:mainconvex} and Theorem \ref{theo:lowrank_approx} imply that for a given rank-$r$ data matrix $\data$, the regularized training problem in \eqref{eq:twolayer_objective_generic} can be approximately solved via convex optimization solvers to achieve an approximation with objective value $p^*\left(1+\frac{L R\sigma_{k+1}}{\beta}\right)^2$ in $\mathcal{O}\Big(d^3k^3\big(\frac{n}{k}\big)^{3k}\Big)$ time complexity, where $p^*$ is the optimal value and $k \leq r$. Therefore, even for full rank data matrices for which the worst-case complexity of solving \eqref{eq:twolayerconvexprogram} is exponential in $d$, this method approximately solves the convex program in \eqref{eq:twolayerconvexprogram} in polynomial-time with strong guarantees.
\end{rem}
\begin{figure*}[t]
\centering
\captionsetup[subfigure]{oneside,margin={1cm,0cm}}
	\begin{subfigure}[t]{0.45\textwidth}
	\centering
	\includegraphics[width=1\textwidth]{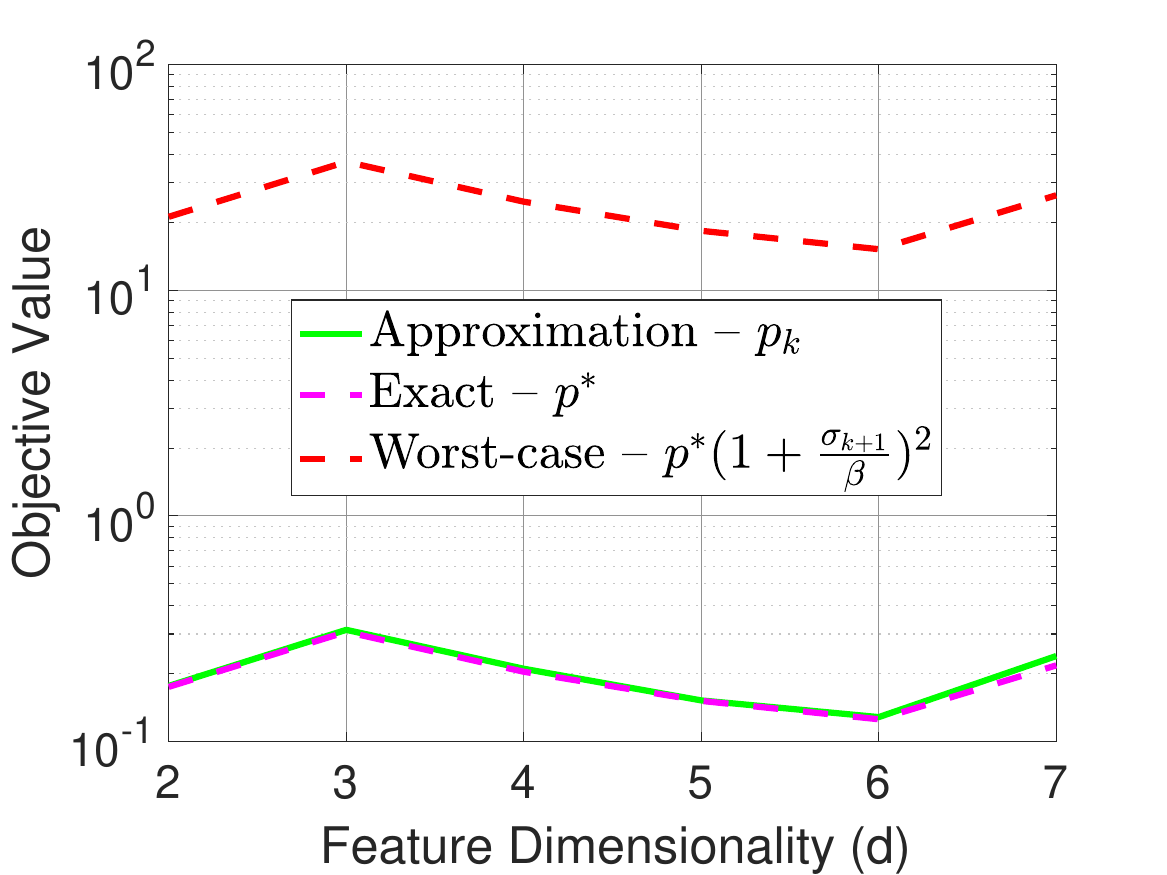}
	\caption{Objective value\centering} \label{fig:rank_error}
\end{subfigure} \hspace*{\fill}
	\begin{subfigure}[t]{0.45\textwidth}
	\centering
	\includegraphics[width=1\textwidth]{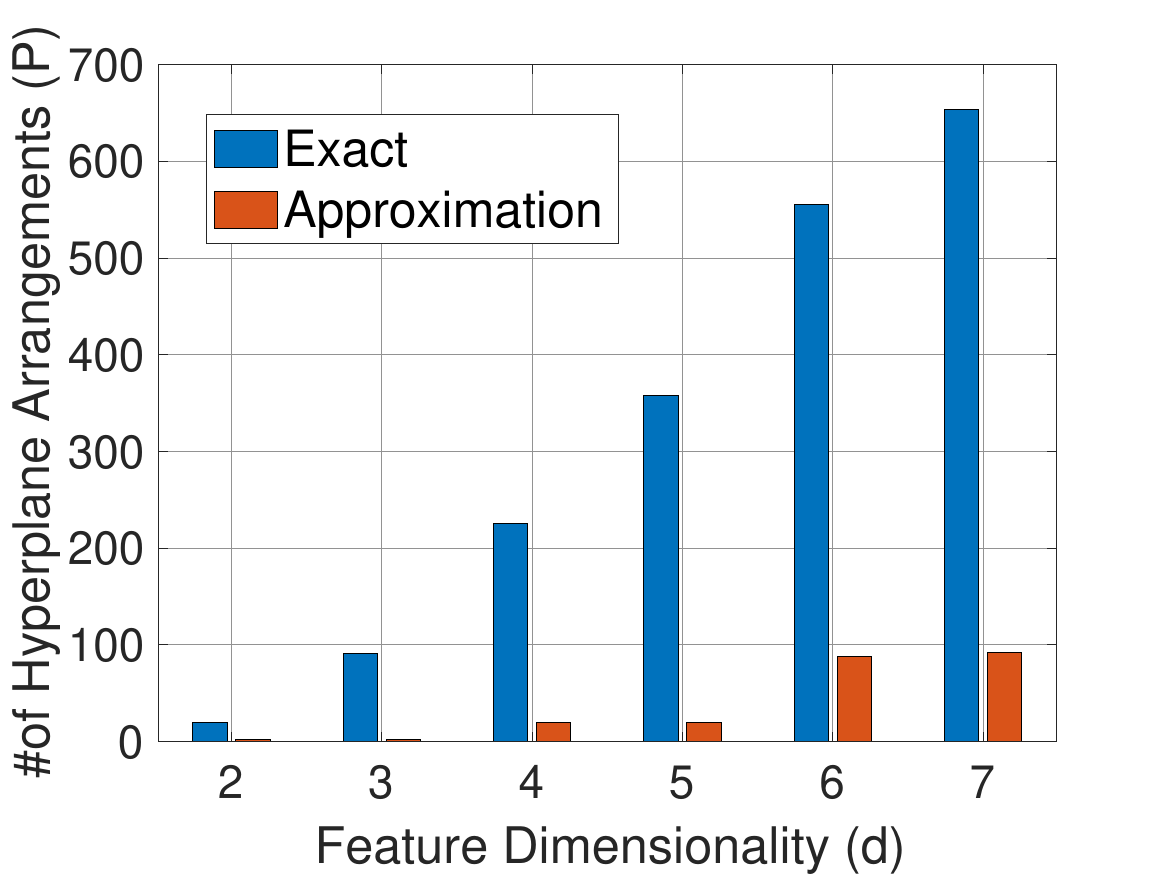}
	\caption{Number of hyperplane arrangements ($P$)\centering} \label{fig:rank_p}
\end{subfigure} \hspace*{\fill}
\caption{Verification of the approximation guarantees in Theorem \ref{theo:lowrank_approx}. Here, we train a two-layer ReLU network using the convex program in Theorem \ref{theo:mainconvex} with $\ell_2$ loss on a synthetic dataset with $n=10$, $\beta=0.1$, and the low-rank approximation $k= \lfloor \frac{d}{2} \rfloor$. To obtain a rank-deficient model, we first generate a random data matrix using a multivariate Gaussian distribution with $\bm{\mu}=\vec{0} $ and $\bm{\Sigma}=\vec{I}_d$ and then explicitly set $\sigma_{k+1}=\ldots=\sigma_d=1$.}\label{fig:low_rank}
\end{figure*}

As an illustration of Theorem \ref{theo:lowrank_approx}, consider a ReLU network training problem with $\ell_2$ loss. The approximation ratio becomes $(1+\frac{\sigma_{k+1}}{\beta})^2$, which is typically close to $1$ due to fast decaying singular values of training data matrices encountered in practice. In Figure \ref{fig:low_rank}, we present a numerical example on i.i.d. Gaussian synthetic data matrices and the low-rank approximation strategy. Figure \ref{fig:rank_error} shows that\footnote{We provide the details of this experiment in \ref{sec:supp_exps}.}  the low-rank approximation of the objective $p_k$ is closer to $p^*$ than the worst-case upper-bound predicted by Theorem \ref{theo:lowrank_approx}. However, in Figure \ref{fig:rank_p}, we observe that the low-rank approximation provides a significant reduction in the number of hyperplane arrangements, and therefore in the complexity of solving the convex program.

\subsection{Efficient sampling of hyperplane arrangements with guarantees} \label{sec:efficient_arrangement}
The convex program in \eqref{eq:convex_program_lowrank} can be globally optimized with a polynomial-time complexity, however, it is not obvious how to generate the hyperplane arrangement matrices $\{ \diag_i\}_{i=1}^P$ in practice. Although there exist algorithms to construct these arrangements, e.g., \cite{edelsbrunner1986constructing}, they can become computationally challenging in high dimensions. In this section, we first show how to efficiently sample these hyperplane arrangements for the convex programs \eqref{eq:twolayerconvexprogram} and  \eqref{eq:convex_program_lowrank}, and then provide probabilistic approximation guarantees.

We first note that the convex program \eqref{eq:twolayerconvexprogram} can be approximated by sampling a set of diagonal matrices $
\{\diag_i\}_{i=1}^{\tilde P}$. For example, we can generate vectors from the standard multivariate Gaussian, or some other distribution, as $\firstw \sim N(\vec{0},\vec{I}_d)$ i.i.d. $\tilde P$ times, and then construct diagonal matrices via \eqref{eq:diagonal_matrix} to solve the reduced convex problem. This is essentially a type of random coordinate descent strategy applied to the convex objective \eqref{eq:twolayerconvexprogram}.


\begin{figure*}[t]
\centering
\captionsetup[subfigure]{oneside,margin={1cm,0cm}}
	\begin{subfigure}[t]{0.49\textwidth}
            \centering
            \includegraphics[width=1\textwidth]{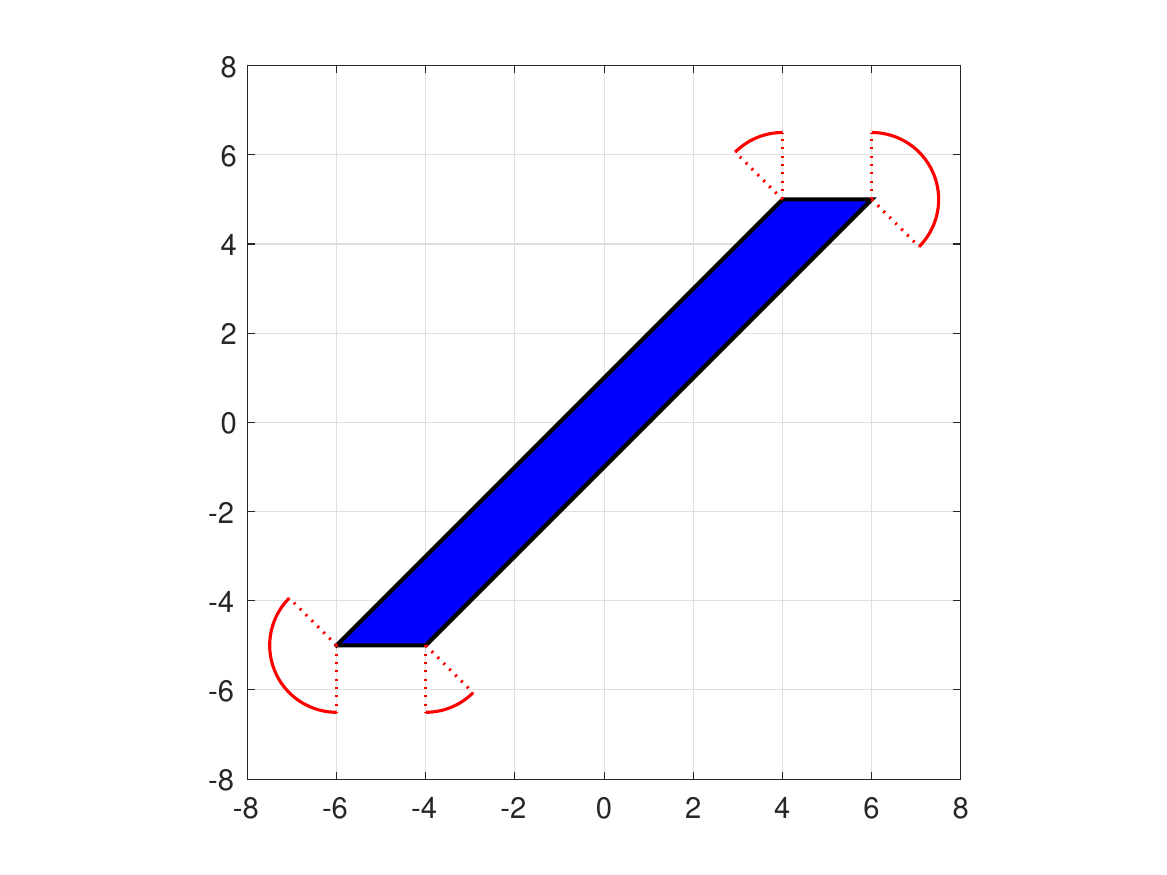}
    \end{subfigure} \hspace*{\fill}
	\begin{subfigure}[t]{0.49\textwidth}
            \centering
            \includegraphics[width=1\textwidth]{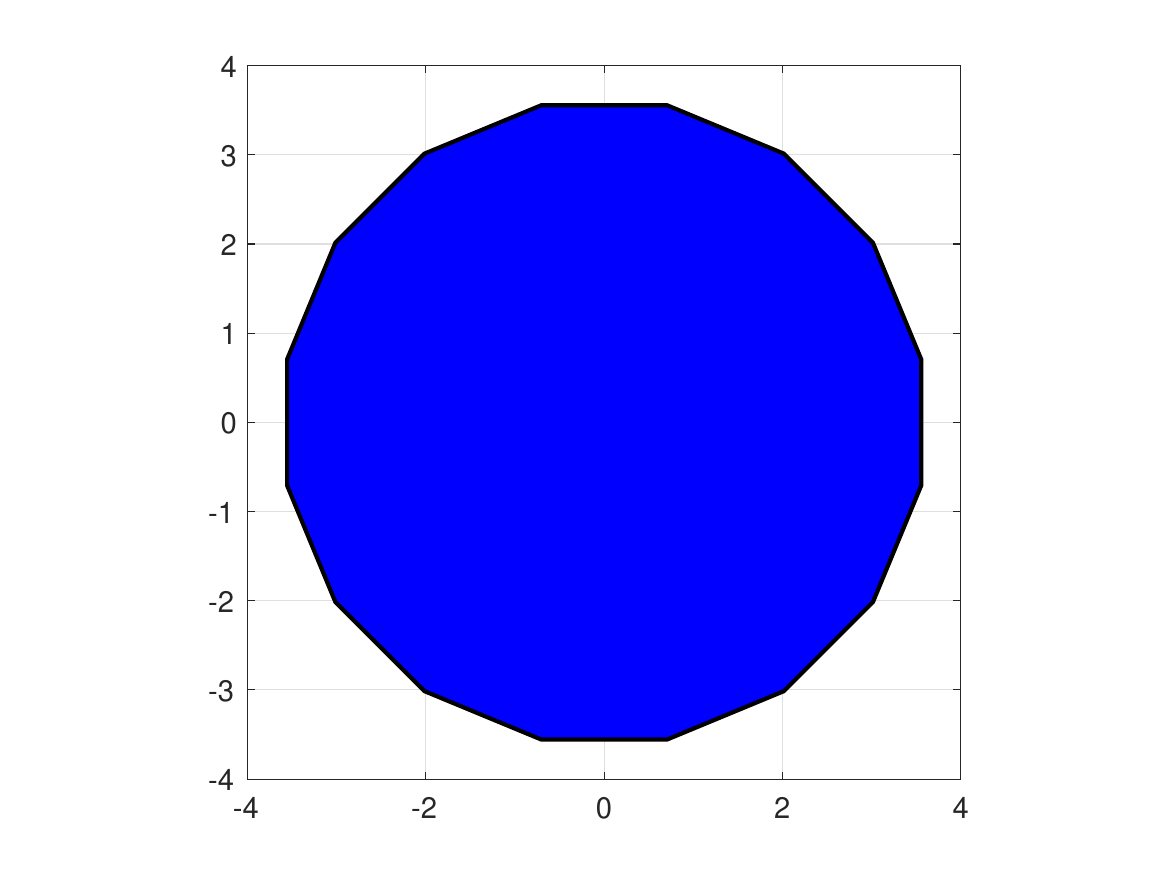}
    \end{subfigure} \hspace*{\fill}
\caption{\textbf{Left:} An illustration of the zonotope $\mathcal{Z}(\data)$ for the data in Example \ref{example3by2} and vertex solid angles $\theta_i$, where $\data=[2 \; 2; 3\; 3; 1 \; 0]$. Note that here the minimum angle is $\theta_{min}=\pi/4$. \textbf{Right:} An illustration of a zonotope $\mathcal{Z}(\data)$ generated with uniformly on the unit sphere $\data \in \mathbb{R}^{16\times 2}$. It can be verified that all the angles are equal, i.e., $\theta_i=\frac{2\pi}{n}=\frac{\pi}{8}\,\forall i$.  }\label{fig:zonotope}
\end{figure*}

We next show that hyperplane arrangement matrices $\{ \diag_i\}_{i=1}^P$ have a one-to-one correspondence to the vertices of a zonotope whose generators are the training data samples. We define the data zonotope $\mathcal{Z}(\data)$, which is a low-dimensional linear projection of a hypercube as follows.
\begin{align*}
    \mathcal{Z}(\data) &:= \{\data^T \vec{u}:\vec{u} \in [0,1]^n\}=\mathrm{Conv}\left\{\sum_{i=1}^n \datavec_i u_i\,:u_i \in \{0,1\},\; \forall i\in[n]\right\},
\end{align*}
where $\mathrm{Conv}$ denotes the convex hull operation. Then, we observe that the extreme points of the zonotope defined above are linked to the hyperplane arrangements that appear in our convex program \eqref{eq:twolayerconvexprogram} since  
\begin{align}
   \vec{u}^*:= \argmax_{\vec{u}\in [0,1]^n} \vec{v}^T \data^T \vec{u} \implies u^*_i=\mathbbm{1}[\vec{x}_i^T \vec{v}\geq0],\,\forall i\,:\,\vec{x}_i^T\vec{v}\neq 0,
   \label{eq:zonotope_support_function}
\end{align}
and we may pick $u^*_i\in [0,1]$ whenever $\vec{x}_i^T\vec{v}=0$ for any $i\in[n]$.
Therefore, for every direction $\vec{v} \in \mathbb{R}^d$, there is an extreme point of $\mathcal{Z}(\data)$ of the form $\vec{e}=\data^T \mathbbm{1}[\data\vec{v}\geq 0] = \sum \vec{x}_i\mathbbm{1}[\vec{x}_i^T\vec{v}\ge 0]$. In particular, an extreme point $\vec{e}$ is optimal when $-\vec{v}$ is in the normal cone of the zonotope at $\vec{e}$, i.e., $\vec{v} \in N_{\mathcal{Z}(\data)}(\vec{e}):=\big\{\vec{w}:\vec{w}^T(\vec{x}-\vec{e})\le 0\,\,\forall \vec{x} \in \mathcal{Z}(\data)\big\}$ due to convex optimality conditions for the problem \eqref{eq:zonotope_support_function}. An alternative representation of the normal cone is given by $ N_{\mathcal{Z}(\data)}(\vec{e}):=\big\{\vec{w}:\sign(\data\vec{w})=\sign(\data\vec{v})\big \}$, where $\vec{e}=\data^T \mathbbm{1}[\data\vec{v}\geq 0]$. The number of extreme points of $\mathcal{Z}(\data)$ is equal to the number of hyperplane arrangements of $\data$, which we denote by $P$. We refer the reader to \cite{grunbaum1967convex} for further details on zonotopes and hyperplane arrangements.

Next, we will now prove that this approach in fact solves the exact convex program in \eqref{eq:twolayerconvexprogram} with high probability provided that $\tilde P$ exceeds a certain threshold. We define the solid angle of a convex cone $\mathcal{C}(\data)$ (see e.g., \cite{ball1997elementary}) as the probability that a randomly drawn standard multivariate Gaussian lies inside $\mathcal{C}(\data)$, i.e.,
\begin{align*}
    \Omega (\mathcal{C}(\data)) := \mathbb{P}_{\vec{w}\sim \mathcal{N}(\vec{0},\vec{w})} \big[  \vec{w} \in \mathcal{C}(\data) \big ] = \frac{1}{(2\pi)^{d/2}} \int_{\mathcal{C}(\data)} e^{-\frac{1}{2}\|\vec{x}\|_2^2}d\vec{x}\,.
\end{align*}
Suppose that the normal cones $N_{\mathcal{Z}(\data)}(\vec{e})$
of $\mathcal{Z}(\data)$ at its extreme points have solid angles given by $\{\theta_i\}_{i=1}^P$. More precisely, we define
\begin{align*}
    \theta_i := \Omega(N_{\mathcal{Z}(\data)}(\vec{e}_i)),\;\forall i \in [P]\,,
\end{align*}
where $\vec{e}_i=\data^T \mathbbm{1}[\data\vec{v}_i\geq 0],  \; \forall i \in [P]$ are $P$ distinct extreme points generated by directions $\{\vec{v}_i\}_{i=1}^P$. Note that all extreme points have strictly positive solid angle since otherwise those points may be removed from the set of extreme points while maintaining the same convex hull. We also include a two-dimensional illustration of the zonotope $\mathcal{Z}(\data)$ for the data in Example \ref{example3by2} and the corresponding solid angles of normal cones $\{\theta_i\}_{i=1}^{P}$ at extreme points in Figure \ref{fig:zonotope}.

The next result shows that all arrangement patterns will be sampled under the assumption that the minimum solid angle of the data zonotope is bounded by a positive constant.
\begin{theo}\label{theo:effcient_sampling}
Let $P$ be the number of hyperplane arrangements for the training data matrix $\data$. Then, in order to sample all $P$ arrangements with probability $1-\epsilon$, it is sufficient to let the number of random samples $\tilde P$ satisfy
 $\tilde P \geq \bar\theta^{-1} P\log\left( P / \epsilon\right)$, where $\bar\theta:=P\min_{i\in[P]:\theta_i>0} \theta_i$ is the minimum solid angle of the normal cones of the zonotope $\mathcal{Z}(\data)$ multiplied by the number of vertices $P$.
\end{theo}
\begin{rem}
We note that the multiplicative factor $P$ is introduced in the definition of $\bar\theta$ in order to remove its inverse dependence on the number of vertices $P$. To give concrete examples, the zonotope $\mathcal{Z}(\data)$ seen in the right panel of Figure \ref{fig:zonotope} is a regular $n$-gon which has $\theta_i = \frac{2\pi}{n}\,\forall i\in[n]$ and therefore $\bar\theta = n\frac{2\pi}{n}=2\pi$ since $P=n$. Similarly, the zonotope  $\mathcal{Z}(\data)$ in the left panel of Figure \ref{fig:zonotope} has $\bar\theta = 4\frac{\pi}{4}=\pi$ since $P=4$.
\end{rem}
Along with the low-rank approximation in Theorem \ref{theo:lowrank_approx} reducing the number of arrangements to $P=\mathcal{O}((n/k)^k)$ for any target rank $k$, the efficient sampling approach in Theorem \ref{theo:effcient_sampling} proves that we can solve the convex program in \eqref{eq:twolayerconvexprogram} with a polynomial-time complexity in all problem parameters. The pseudocode for this training approach is presented in Algorithm \ref{alg:convex_rank}.

\section{Convolutional neural networks} \label{sec:convolutions}
Here, we introduce extensions of our approach to CNNs. Two-layer convolutional networks with $m$ hidden neurons (filters) of dimension $d$ and fully connected output layer weights can be described by patch matrices $\data_k \in \real^{n\times d},\,k=1,...,K$. With this notation, $\data_k\firstw_{j}$ represents the $k^{th}$ spatial dimension of the convolution with the filter $\firstw_{j}$ across the dataset.

{\subsection{Standard convolutional networks}\label{sec:cnn_avgpool}
We first analyze convolutional neural networks with global average pooling, which is a commonly used technique for reducing the dimensionality of the feature maps in a convolutional neural network. Global average pooling involves taking the average of all the values in each spatial feature map. Using the notation above, the output of a CNN with global average pooling is given by  
$$\frac{1}{K}\sum_{k=1}^K \sum_{j=1}^m  \act(\data_k\firstw_{j})\secondw_{j}.$$ 
For this architecture, we consider the following training problem
\begin{align} \label{eq:training_globalavg}
    \min_{\theta \in \Theta}  \mathcal{L}\left( \frac{1}{K}\sum_{k=1}^K f_{\theta}(\data_k),\labelvec \right) +\frac{\beta}{2} (\|\firstwmat\|_F^2+\|\secondwvec\|_2^2)\,,
\end{align}
where $f_{\theta}(\data_k)=\sum_{j=1}^m  \act(\data_k\firstw_{j})\secondw_{j}$. We first define an augmented data matrix by concatenating the patch matrices as $\hat{\data}:= \begin{bmatrix} \data_1^T& \data_2^T & \ldots & \data_K^T \end{bmatrix}^T \in \mathbb{R}^{nK \times d}$. Then, \eqref{eq:training_globalavg} can be equivalently written as
\begin{align} \label{eq:training_globalavgv2}
    \min_{\theta \in \Theta}  \tilde{\mathcal{L}}(f_{\theta}(\hat{\data}),\labelvec) +\frac{\beta}{2} (\|\firstwmat\|_F^2+\|\secondwvec\|_2^2)\,,
\end{align} 
where we define the loss function as
\begin{align*}
    \tilde{\mathcal{L}}(f_{\theta}(\hat{\data}),\labelvec):=\mathcal{L}\left( \frac{1}{K}\sum_{k=1}^K f_{\theta}(\data_k),\labelvec \right).
\end{align*}
This shows that the convolutional network training problem with global average pooling in \eqref{eq:training_globalavg} can be cast as a standard fully connected network training problem as in \eqref{eq:training_globalavgv2} using the training data $\hat{\data}$ and modified convex loss function $\tilde{\mathcal{L}}$. Therefore, the convex program \eqref{eq:twolayerconvexprogram} solves the above problem exactly in $\mathcal{O}\Big(d^3r^3\big(\frac{nK}{r}\big)^{3r}\Big)$ complexity, where $d$ is the number of variables in a single filter and $r$ is the rank of $\hat{\data}$. It holds that $r\le d$ since $\hat{\data}\in\mathbb{R}^{nK\times d}$. Note that typical CNNs employ $m$ filters of constant size, e.g., $3\times 3\times m$ ($d$=9) in the first layer \cite{he2016deep}. As a result of this small feature dimension (or filter size), our result implies that globally optimizing a CNN architecture can be done in a polynomial-time, i.e., polynomial in all dimensions when the filter size $d$ is a constant.}

\subsection{Linear convolutional network training as a Semi-Definite Program (SDP)} \label{sec:cnn_linear}
We now analyze CNNs linear activations $\act(\datascalar)=\datascalar$ trained via the following optimization problem
\begin{align}
\label{eq:CNN_linear}
    &\min_{\theta \in \Theta}  \mathcal{L}(f_{\theta,c}(\{\data_k\}_{k=1}^K),\labelvec) +\frac{\beta}{2} \sum_{j=1}^m (\|\firstw_{j}\|_2^2+\|\secondwvec_j\|_2^2)\,,
\end{align}
where
\begin{align*}
    f_{\theta,c}(\{\data_k\}_{k=1}^K)= \sum_{k=1}^K \sum_{j=1}^m \data_k \firstw_j \secondw_{jk}.
\end{align*}
The corresponding dual problem is given by
\begin{align}
\label{eq:linearconvdual}
    &\max_{\dual} -\mathcal{L}^*(\dual)\,\,\mbox{s.t.} \max_{\firstw \in \ball_2}\, \sqrt{\sum_{k=1}^K\big( \dual^T \data_k \firstw \big)^2} \le \beta.
\end{align}
By similar arguments to those used in the proof of Theorem \ref{theo:mainconvex}, strong duality holds. Furthermore, the maximizers of the constraint are the maximal eigenvectors of $\sum_k \data_k^T \dual \dual^T \data_k$, which are optimal neurons (filters). Thus, we can express \eqref{eq:linearconvdual} as the following SDP
\begin{align}
    &\max_{\dual} -\mathcal{L}^*(\dual) \mbox{ s.t.  } \sigma_{\max}\left([\data_1^T \dual\, ...\, \data_K^T \dual ]\right)\le \beta.\label{eq:SDP}
\end{align}
The dual of the above SDP is a nuclear norm penalized convex optimization problem (see \ref{sec:sdp_appendix})
\begin{align}
\label{eq:CNN_linear_nuclear}
    &\min_{\vec{z}_k\in\real^d   } \mathcal{L}(\hat{f}_{\theta,c}(\{\data_k\}_{k=1}^K),\labelvec)+ \beta\Big \|[\vec{z}_1,\ldots,\vec{z}_K] \Big \|_{*},
\end{align}
where 
\begin{align*}
    \hat{f}_{\theta,c}(\{\data_k\}_{k=1}^K)= \sum_{k=1}^K \data_k \vec{z}_k
\end{align*}
and $\Big\|[\vec{z}_1,\ldots,\vec{z}_K] \Big \|_{*}=\|\vec{Z}\|_*:=\sum_i \sigma_i(\vec{Z})$ is the nuclear norm, i.e, sum of singular values, of $\vec{Z}$. In convex optimization, the nuclear norm is often used as a convex surrogate for the rank of a matrix, with the rank being a non-convex function.  \cite{recht2010guaranteed,Fa02}.

%

\subsection{Linear circular convolutional networks}\label{sec:circular_cnn_linear}
Now, suppose that the patches are padded with zeros and extracted with stride one, and we have full-size filters that can be represented by circular convolution. Then the circular convolution version of \eqref{eq:CNN_linear} can be written as
\begin{align}\label{eq:linear_cnn}
    \min_{\theta \in \Theta} \mathcal{L}(  f_{\theta,c}(\data),\labelvec)+\frac{\beta}{2}\sum_{j=1}^m \left( \|\firstw_j\|_2^2+\|\secondwvec_j\|_2^2 \right),
\end{align}
where 
\begin{align*}
    f_{\theta,c}(\data)= \sum_{j=1}^m\data \firstwmat_j\secondwvec_j,
\end{align*}
and $\firstwmat_j \in \mathbb{R}^{d \times d}$ is a circulant matrix generated by a circular shift modulo $d$ using the elements $\firstw_j \in \mathbb{R}^h$. Then, the SDP in \eqref{eq:SDP} reduces to (see Appendix \ref{sec:circular_linear_cnn_appendix})
\begin{align}\label{eq:linear_cnn_l1}
    \min_{\vec{z} \in \mathbb{C}^d} \mathcal{L}(\hat{f}_{\theta,c}(\hat{\data}),\labelvec)+ \beta \|\vec{z}\|_1,
\end{align}
where $\hat{\data}=\data \vec{F}$, $\vec{F} \in \mathbb{C}^{d \times d}$ is the Discrete Fourier Transform (DFT) matrix, and
$\hat{f}_{\theta,c}(\hat{\data})= \hat{\data} \vec{z}$. {We note that certain linear CNNs trained via gradient descent exhibit similar spectral regularization properties \cite{gunasekar2019implicit}.}

\section{$\ell_p$-norm regularization of hidden weights} \label{sec:lp_reg}
In this section, we reconsider two-layer neural network training problems with an alternative $\ell_p^2$ regularization on the hidden neurons, which is a generalization of the setting in \eqref{eq:twolayer_objective_generic}. Hence, we have the following optimization problem
\begin{align}\label{eq:twolayer_objective_generic_lp}
     p^*=\min_{\theta \in \Theta} \mathcal{L}(f_{\theta}(\data),\labelvec)+\frac{\beta}{2}\sum_{j=1}^m (\|\firstw_j\|_p^2 + |{\secondw_j}|^2)\,.
\end{align}
After applying the scaling in Lemma \ref{lemma:scaling}, we equivalently write \eqref{eq:twolayer_objective_generic_lp} as
\begin{align}\label{eq:twolayer_objective_generic_lp_l1}
     p^*=\min_{\theta \in \Theta_s} \mathcal{L}(f_{\theta}(\data),\labelvec)+\beta \|\secondwvec\|_1 \,, 
\end{align}
where $\Theta_s:=\{\theta \in \Theta: \|\firstw_j\|_p\leq 1,\, \forall j \in [m]\}$. Therefore, we have the following equivalent convex program for $\ell_p^2$ regularized networks.
\begin{cor} \label{cor:main_lp}
As a result of Theorem \ref{theo:mainconvex}, the non-convex training problem in \eqref{eq:twolayer_objective_generic_lp} can be cast as a finite dimensional convex program as follows
\begin{align}
&p^*=\min_{ \weight,\weight^\prime  \in \mathcal{C}(\data)}\, \mathcal{L}(f_{\theta_c}(\dataf),\labelvec)  +\beta  \sum_{i=1}^{2P} \|\weight_i\|_p, \label{eq:twolayerconvexprogram_lp}   
\end{align}
where $f_{\theta_c}(\dataf)=\dataf\weight$ and the rest of definitions directly follow from Theorem \ref{theo:mainconvex}.
\end{cor}
We note that the case where $p=1$ is regularized via $\sum_{i=1}^{2P} \|\firstw_i\|_1$ with the squared loss is equivalent to the LASSO feature selection method with additional linear constraints \cite{Tibshirani96,Chen98}.

\section{Interpolation regime (weak regularization)} \label{sec:weakly_reg}
We now consider the minimum-norm variant of \eqref{eq:twolayer_objective_generic_l1}, which corresponds to interpolation or weak regularization, i.e., $\beta\rightarrow 0$.
Suppose that the minimum value of the loss $\mathcal{L}(f_{\theta}(\data),\labelvec)$ is zero, which is satisfied by many popular choices, e.g., squared loss and hinge loss. Taking the $\beta \rightarrow 0$ limit yields the following optimization problem
\begin{align}\label{eq:twolayer_objective_generic_weakreg_l1}
     p^*_{\beta\rightarrow 0}=\min_{\theta \in \Theta_s} \|\secondwvec\|_1 \,, \text{ s.t. }\mathcal{L}(f_{\theta}(\data),\labelvec)=0.
\end{align}
Then, by Theorem \ref{theo:mainconvex}, the equivalent convex program for \eqref{eq:twolayer_objective_generic_weakreg_l1} is
\begin{align}
&p^*_{\beta\rightarrow 0}=\min_{ \weight \in \mathcal{C}(\data)}\,    \sum_{i=1}^{2P} \|\weight_i\|_2 \text{ s.t. }\mathcal{L}(f_{\theta_c}(\dataf),\labelvec)=0, \label{eq:twolayerconvexprogram_weakreg}   
\end{align}
given that the set $\mathcal{L}(f_{\theta}(\data),\labelvec)=0$ is convex.
\begin{rem}
Notice that \eqref{eq:twolayerconvexprogram_weakreg} represents a convex optimization problem that seeks to find a solution with minimum group norm and zero training error. Considering the squared loss, \eqref{eq:twolayerconvexprogram_weakreg} further simplifies to
\begin{align*}
&\min_{ \weight,\weight^\prime  \in \mathcal{C}(\data)}\,   \sum_{i=1}^{2P} \|\weight_i\|_2 \text{ s.t. } \dataf\weight=\labelvec. 
\end{align*}
{The above form illustrates an interesting contrast between our exact formulation and various kernel characterizations such as infinitely wide networks and the Neural Tangent Kernel \cite{ntk_jacot}  \cite{lazy_training_bach, bach_kernel2,srebro_kernel,srebro_kernel2}. These kernel formulations are related to approximating the training problem \eqref{eq:twolayer_objective_generic} as a minimum $\ell_2$-norm kernel interpolation using a fixed kernel matrix constructed from $\data$. In contrast, our characterization minimizes $\ell_{2,1}$-norm encouraging feature selection after a fixed high-dimensional feature map. More importantly, unlike approximations of neural networks via kernel methods, our approach provides an exact characterization of the training problem \eqref{eq:twolayer_objective_generic}.}
\end{rem}

\section{Hyperplane arrangements} \label{sec:hyperplane_arrangements}
In this section, we will explore the diagonal matrices that appear in our convex optimization problem \eqref{eq:twolayerconvexprogram}. These matrices are determined by the hyperplane arrangements of the data matrix $\data$. We will also discuss how to exactly construct these arrangements in order to solve the convex program to global optimality. Additionally, we will introduce a class of data matrices for which the arrangements corresponding to non-zero variables at the optimum can be simplified. This provides a significant computational complexity reduction in finding the optimal solution. 

\subsection{Constructing and approximating arrangement patterns}
Constructing hyperplane arrangements has long been an important area of study in discrete mathematics and computational geometry. There are several analytic approaches to construct all possible hyperplane arrangements for a given data matrix $\data$. We refer the reader to \cite{edelsbrunner1986constructing,avis1996reverse,halperin2017arrangements,rada2018new}. In \cite{edelsbrunner1986constructing}, the authors present an algorithm that enumerates all possible hyperplane arrangements in $\mathcal{O}(n^r)$ time for a rank-$r$ data matrix. 

An alternative approach to reduce computational cost is to randomly sample a subset of hyperplane arrangements, as described in Section \ref{sec:efficient_arrangement}. This approximate solution to the convex neural network problem \eqref{eq:twolayerconvexprogram} involves solving a subsampled convex program and is backed by approximation guarantees, as shown in Theorem \ref{theo:effcient_sampling}. Our numerical experiments in Section \ref{sec:numerical} demonstrate that this approximation scheme performs exceptionally well in practice.

\subsection{Spike-free data matrices}
Now we show that the convex program \eqref{eq:twolayerconvexprogram} simplifies significantly for a certain class of data matrices. We first define the minimal set of hyperplane arrangements that globally optimizes \eqref{eq:twolayer_objective_generic} as
\begin{align*}
    \mathcal{D}^*:=\argmin_{\substack{\mathcal{D}  }} \big \vert \left\lbrace \mathcal{D} \subseteq \mathcal{D}_{opt}: p^*=d^*\right \rbrace\big \vert,
\end{align*}
where $\mathcal{D}_{opt} $ is defined as
\begin{align*}
    \mathcal{D}_{opt}:=\left\{ \diag_i\,:\, \max_{\substack{\firstw \in \ball_2 \cap  \mathcal{C}(\data)}}\left \vert \dual^{*^T} \diag_i \data \firstw \right \vert=\beta  \right\}
\end{align*}
based on the dual characterization in \eqref{eq:twolayer_dual_generic}, $\dual^*$ denotes the optimal dual parameter, and $\mathcal{C}(\data)$ is defined in Theorem \ref{theo:mainconvex}. We define $P^*:=\vert \mathcal{D}^*\vert$ as the minimum number of hyperplane arrangements required to solve the convex program \eqref{eq:twolayerconvexprogram} for a given data matrix $\data \in \mathcal{X}$.

Next, we introduce a set of data matrices $\mathcal{X}$, called spike-free\footnote{The definition and further properties of spike-free matrices can be found in \cite{ergen2020convex,ergen2020convex2}.}, for which one hyperplane arrangement is sufficient to solve \eqref{eq:twolayerconvexprogram} exactly.

As an example, whitened high-dimensional ($n\le d$) data matrices that satisfy $\data\data^T=\vec{I}_n$ are spike-free data matrices as shown in \cite{ergen2020convex2}. We first define the set $\rectset:=\{\act(\data \firstw)\,:\, \firstw \in \ball_2\}$. Then, we say that a data matrix $\data$ is spike-free if 
$\rectset$ can be equivalently represented as $\data \ball_2 \cap \mathbb{R}_+^n$, where $\data \ball_2=\{\data \firstw\,:\, \firstw \in \ball_2\}$. More precisely, we define the set of spike-free data matrices as $\mathcal{X}=\{\data \in \mathbb{R}^{n \times d}:\rectset=\data \ball_2 \cap \mathbb{R}_+^n\}$. Assuming $\data$ is spike-free, the output of the ReLU activation in \eqref{eq:twolayer_dual_generic}, i.e., $\act(\data \firstw)$, can be replaced with $\{\data \firstw:\data \firstw\geq 0\}$, which corresponds to a single hyperplane arrangement $\mathcal{D}^*=\vec{I}_n$. Consequently, the number of hyperplane arrangements in Theorem \ref{theo:mainconvex} reduces to one, i.e., $P^*=1$ and $\mathcal{D}^*=\vec{I}_n$. Based on this observation, the equivalent convex program for spike-free data matrices is as follows.

\begin{theo}\label{theo:main_spikefree}
Given a spike-free data matrix $\data \in \mathcal{X}$, the equivalent convex program for the non-convex problem in \eqref{eq:twolayer_objective_generic} is given by
\begin{align}
&\min_{ \substack{\weight,\weight^\prime \in \mathcal{C}_s(\data)} }\,  \mathcal{L}(\data(\weight^\prime-\weight),\labelvec)  +\beta \left(\|\weight\|_{2}+\|\weight^\prime\|_{2}\right), \label{eq:twolayerconvexprogram_spikefree}   
\end{align}
where
\begin{align*}
    \mathcal{C}_s(\data):=\{\weight,\weight^\prime \in \mathbb{R}^d\;:\;\data\weight\geq 0,\; \data \weight^\prime \geq 0\}.
\end{align*}
\end{theo}
The above result shows that the convex program in \eqref{eq:twolayerconvexprogram} reduces to a simple mixture of two linear models for spike-free data matrices.

\section{Vector output networks}\label{sec:vector_output}
In this section, we consider a neural network with $C$ outputs, which are commonly used for vector valued prediction, e.g., multi-class classification or vector regression. Here, we have matrix valued targets $\labelmat \in \mathbb{R}^{n \times C}$, and the non-convex regularized training problem is as follows
\begin{align}\label{eq:twolayer_objective_generic_vector}
     p_v^*:=\min_{\theta \in \Theta} \mathcal{L}(f_{\theta}(\data),\labelmat)+\frac{\beta}{2} (\|\firstwmat\|_F^2 + \|\secondwmat\|_F^2)\,,
\end{align}
where $f_{\theta}(\data)=\act(\data \firstwmat)\secondwmat$. By applying the scaling argument in Lemma \ref{lemma:scaling}, \eqref{eq:twolayer_objective_generic_vector} can be written as
\begin{align}\label{eq:twolayer_objective_generic_vector_scaled}
     p_v^*:=\min_{\theta \in \Theta_s} \mathcal{L}(f_{\theta}(\data),\labelvec)+\beta \sum_{j=1}^m\|\secondwvec_j\|_2 \,.
\end{align}
Then, applying similar steps as in Theorem \ref{theo:mainconvex}, we obtain the following result.
\begin{theo}\label{theo:mainconvex_vector}
The non-convex training problem in \eqref{eq:twolayer_objective_generic_vector} can be cast as a finite dimensional convex program as follows
\begin{align}\label{eq:twolayerconvexprogram_vector}  
&p_v^*=\min_{ \weightmat_i  \in \mathbb{R}^{d \times C}}\, \mathcal{L}(f_{\theta_c}(\dataf),\labelmat)  +\beta \sum_{i=1}^P \|\weightmat_i\|_{\mathcal{C}_i},  
\end{align}
where $\theta_c:=\{\{\weightmat_i\}_{i=1}^P\}$, $f_{\theta_c}(\data)$, and the constrained nuclear norm $\|\cdot\|_{\mathcal{C}_i}$ are defined as 
\begin{align*}
f_{\theta_c}(\data):&= \sum_{i=1}^P \diag_i\data\weightmat_i\\
\|\weightmat\|_{\mathcal{C}_i}:&=\min_{t\ge 0\,} ~t \quad \mbox{s.t.} \quad \weightmat \in t\, \mathrm{Conv}\left\{\vec{Z}=\vec{u} \vec{g}^T\,: \, \vec{u} \in \ball_2 \cap \mathcal{P}_i,\, \|\vec{Z}\|_*\leq 1\right\}
\end{align*}
where $\mathrm{Conv}$ denotes the convex hull of its argument and $\mathcal{P}_i:=\{\vec{u} \in \mathbb{R}^d \, :\, (2\diag_i -\vec{I}_n)\data \vec{u} \geq 0\}$ are linear constraints.
\end{theo}
{We note that the norm $\|\cdot\|_{\mathcal{C}_i}$ is a constrained version of the nuclear norm, and therefore induces low-rank structure in the variables $\weightmat_1,...,\weightmat_P$. Therefore, in contrast to scalar output networks, Theorem \ref{theo:mainconvex_vector} shows that weight decay regularized neural networks with piecewise linear activations can be equivalently characterized as piecewise low-rank convex models. We further observe that dropping the linear constraints $\mathcal{P}_i$ from the definition of $\|\weightmat\|_{\mathcal{C}_i}$  reduces the constrained nuclear norm to the ordinary nuclear norm. In this case, the regularization term in the convex objective \eqref{eq:twolayerconvexprogram_vector} simplifies to the sum of nuclear norms, which is a natural generalization of the group $\ell_1$ regularizer in \eqref{eq:twolayerconvexprogram}. A numerical algorithm to solve \eqref{eq:twolayerconvexprogram_vector} to global optimality was proposed in \cite{sahiner2021vectoroutput}. Dropping the linear constraints was investigated in \cite{vikul2021generative}. Additionally, recent literature showed that nuclear norm also plays a role in the implicit regularization of linear networks trained via gradient descent \cite{arora2019implicit,gunasekar2017implicit}. }

\subsection{$\ell_1^2$ regularization for the second-layer}
Although the problem in \eqref{eq:twolayerconvexprogram_vector} is convex, handling the constrained nuclear norm $\|\cdot\|_{\mathcal{C}_i}$ can be challenging for high-dimensional problems. To alleviate this, we consider a modification of the weight decay regularization as follows 
\begin{align}\label{eq:twolayer_objective_generic_vector_l1}
     p_{v1}^*:=\min_{\theta \in \Theta} \mathcal{L}(f_{\theta}(\data),\labelvec)+\frac{\beta}{2}\sum_{j=1}^m (\|\firstw_j\|_2^2 + \|\secondwvec_j\|_1^2) \,.
\end{align}
Next, we show that the above problem can be cast as a polynomial-time solvable convex program.
\begin{theo}\label{theo:mainconvex_vector_l1}
 The non-convex problem in \eqref{eq:twolayer_objective_generic_vector_l1} can be equivalently formulated as the following convex program
 \begin{align}\label{eq:twolayerconvexprogram_vector_l1}  
&p_{v1}^*=\min_{ \weight_{l}  \in \mathcal{C}(\data)}\, \sum_{l=1}^C\mathcal{L}(\dataf \weight_l,\labelvec_l)  +\beta \sum_{l=1}^C\sum_{i=1}^{2P} \|\weight_{l,i}\|_2,
\end{align}
where the set $\mathcal{C}(\data)$ and $P$ are defined as in Theorem \ref{theo:mainconvex}.
\end{theo}
We remark that \eqref{eq:twolayerconvexprogram_vector_l1} can be decomposed into $C$ independent convex programs, each of which is the same as \eqref{eq:twolayerconvexprogram}. Therefore, unlike \eqref{eq:twolayerconvexprogram_vector}, the problem in \eqref{eq:twolayerconvexprogram_vector_l1} can be efficiently solved via standard convex optimization solvers.

\begin{figure*}[ht]
\centering
\captionsetup[subfigure]{oneside,margin={1cm,0cm}}
	\begin{subfigure}[t]{0.32\textwidth}
	\centering
	\includegraphics[width=1\textwidth, height=0.8\textwidth]{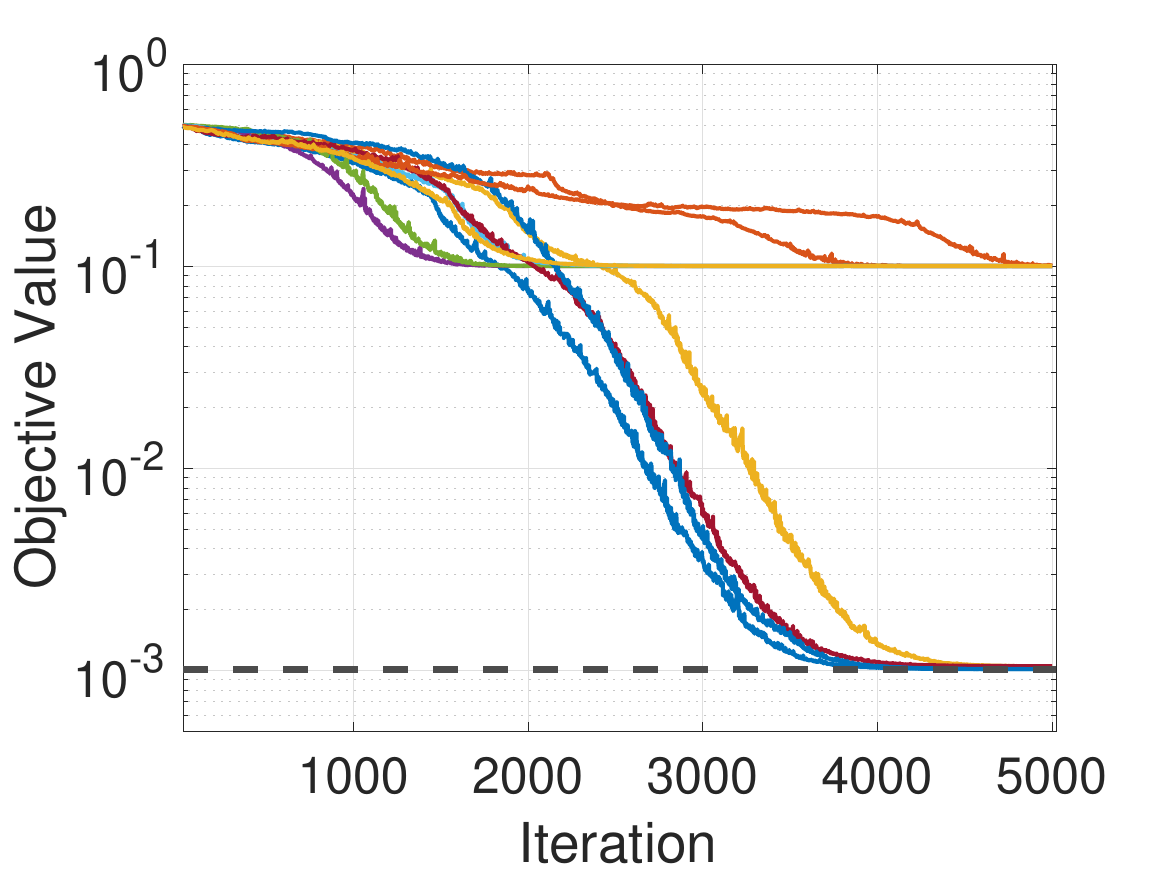}
	\caption{$m=8$\centering} \label{fig:sgd_m=8}
\end{subfigure} \hspace*{\fill}
	\begin{subfigure}[t]{0.32\textwidth}
	\centering
	\includegraphics[width=1\textwidth, height=0.8\textwidth]{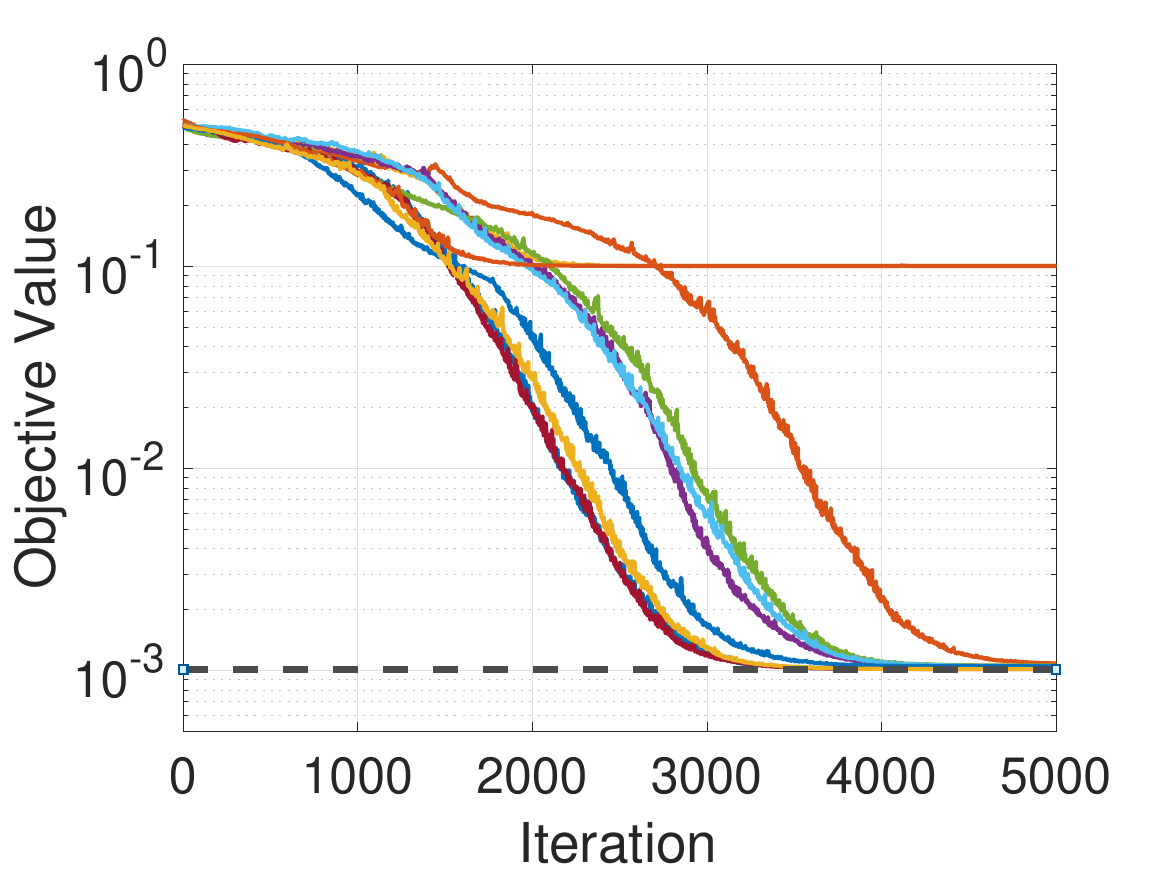}
	\caption{$m=15$\centering} \label{fig:sgd_m=15}
\end{subfigure} \hspace*{\fill}
	\begin{subfigure}[t]{0.32\textwidth}
	\centering
	\includegraphics[width=1\textwidth, height=0.8\textwidth]{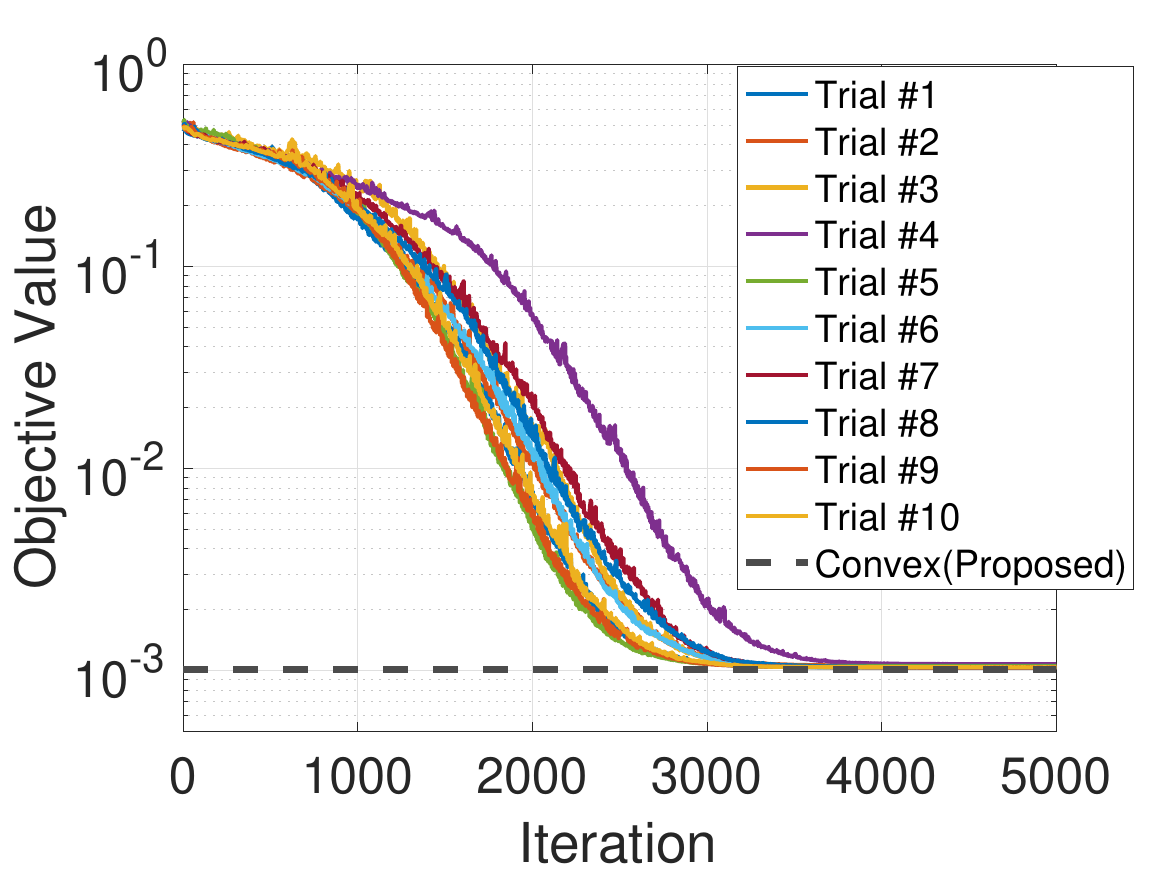}
	\caption{$m=50$\centering} \label{fig:sgd_m=50}
\end{subfigure} \hspace*{\fill}
\caption{Training cost of a two-layer ReLU network trained with SGD (10 initialization trials) on a one dimensional dataset with $(n,d,\beta)=(5,1,10^{-3})$, where Convex denotes proposed convex programming approach in \eqref{eq:twolayerconvexprogram}. SGD can be stuck at local minima for small $m$, while the proposed approach is optimal as guaranteed by Theorem \ref{theo:mainconvex}. }\label{fig:sgd_1d}

\centering
\captionsetup[subfigure]{oneside,margin={1cm,0cm}}
	\begin{subfigure}[t]{0.45\textwidth}
	\centering
	\includegraphics[width=.9\textwidth, height=0.7\textwidth]{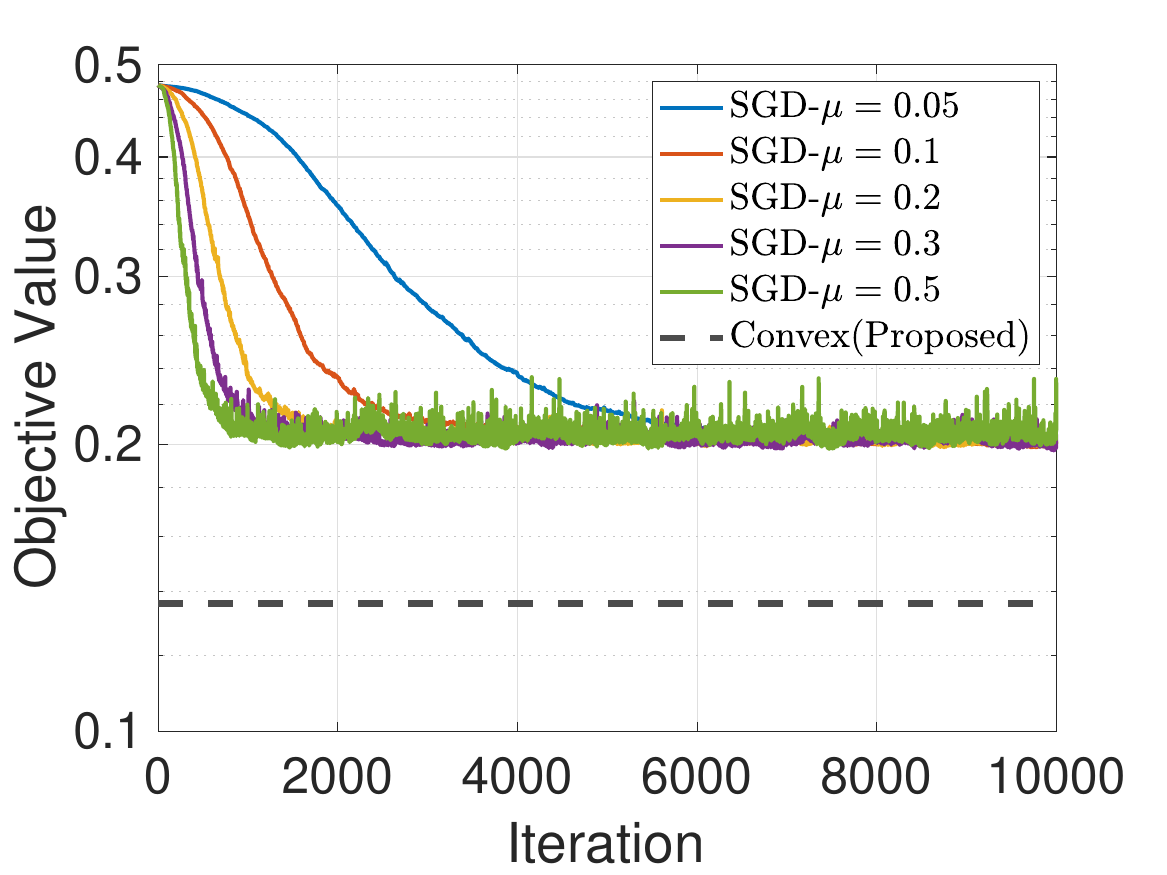}
	\caption{Objective value\centering} \label{fig:ecg_obj}
\end{subfigure} \hspace*{\fill}
	\begin{subfigure}[t]{0.45\textwidth}
	\centering
	\includegraphics[width=.9\textwidth, height=0.7\textwidth]{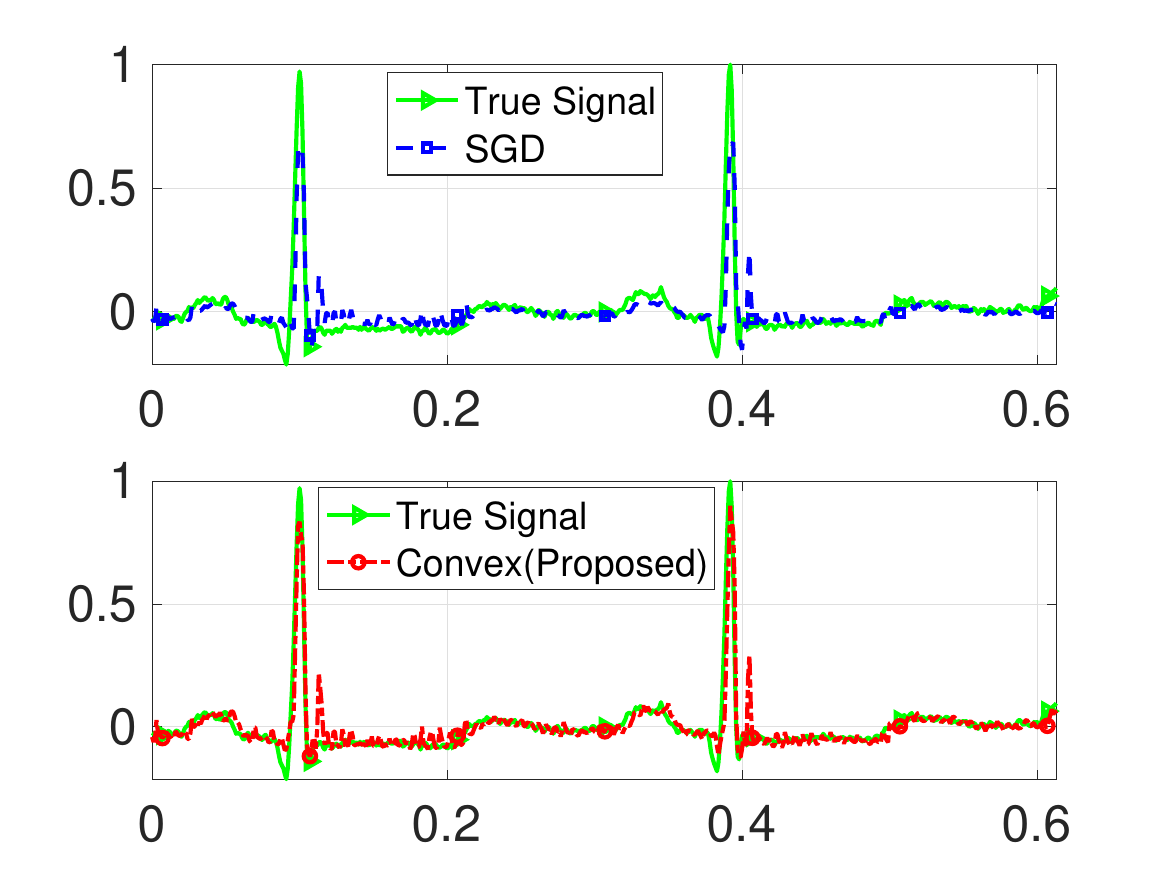}
	\caption{Test prediction\centering} \label{fig:ecg_pred}
\end{subfigure} \hspace*{\fill}
\caption{Prediction performance comparison of a two-layer ReLU network trained with SGD and the convex program \eqref{eq:twolayerconvexprogram} on the ECG dataset, where $(n,d,m,\beta)=(2393,3,50,0.005)$ and $\mu$ denotes the learning rate for SGD. As predicted by our theory, SGD provides poor training and test performance compared to the convex program \eqref{eq:twolayerconvexprogram}.}\label{fig:ecg}
\end{figure*}

\begin{figure*}[ht]
\centering
\captionsetup[subfigure]{oneside,margin={1cm,0cm}}
	\begin{subfigure}[t]{0.32\textwidth}
	\centering
	\includegraphics[width=1\textwidth, height=0.8\textwidth]{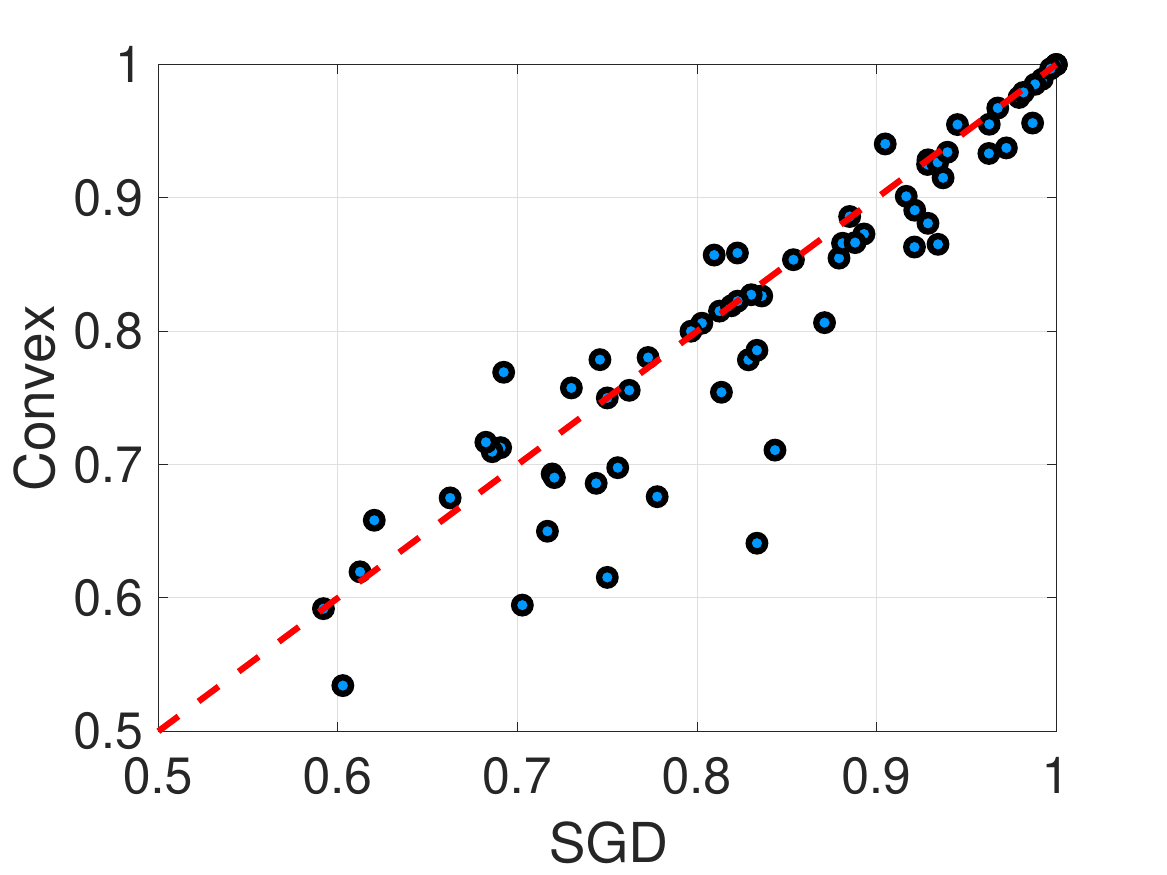}
	\caption{Test accuracy-$m=10$\centering} \label{fig:uci_acc10}
\end{subfigure} \hspace*{\fill}
	\begin{subfigure}[t]{0.32\textwidth}
	\centering
	\includegraphics[width=1\textwidth, height=0.8\textwidth]{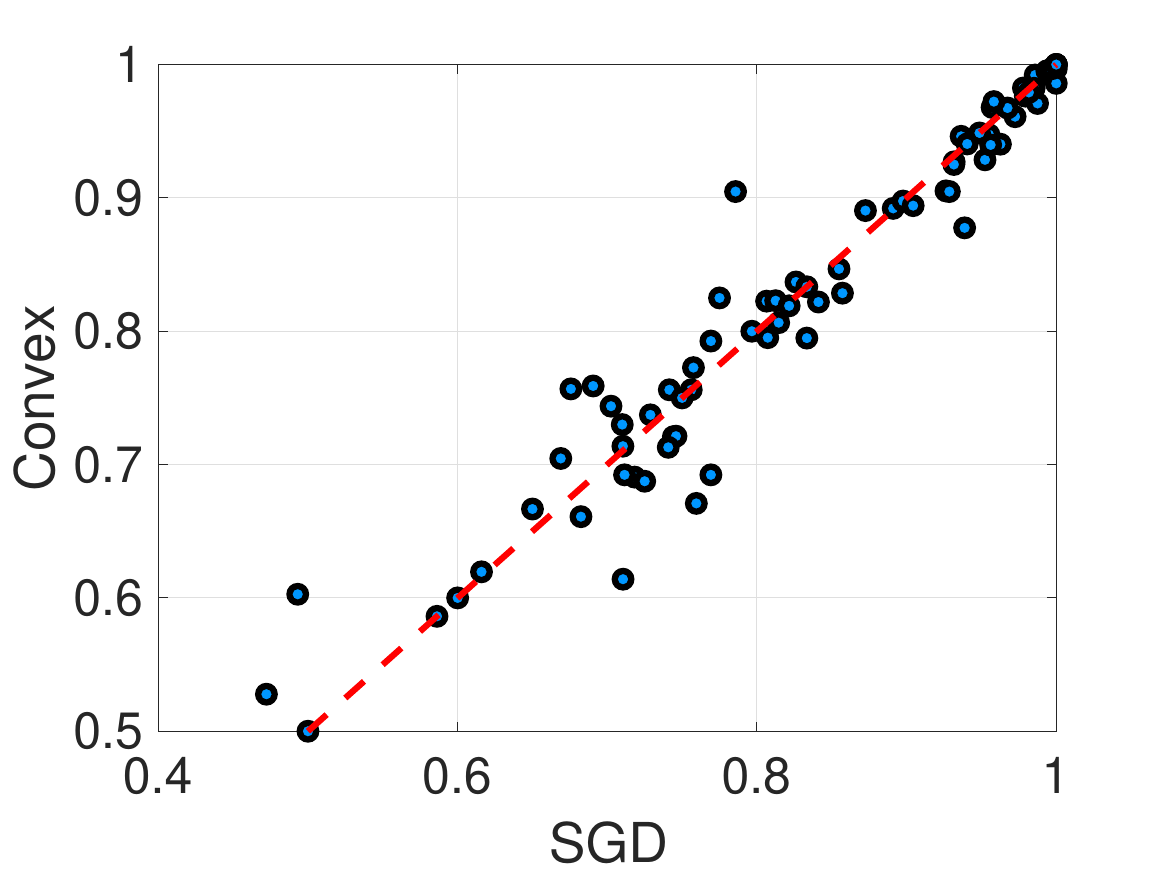}
	\caption{Test accuracy-$m=50$\centering} \label{fig:uci_acc50}
\end{subfigure} \hspace*{\fill}
	\begin{subfigure}[t]{0.32\textwidth}
	\centering
	\includegraphics[width=1\textwidth, height=0.8\textwidth]{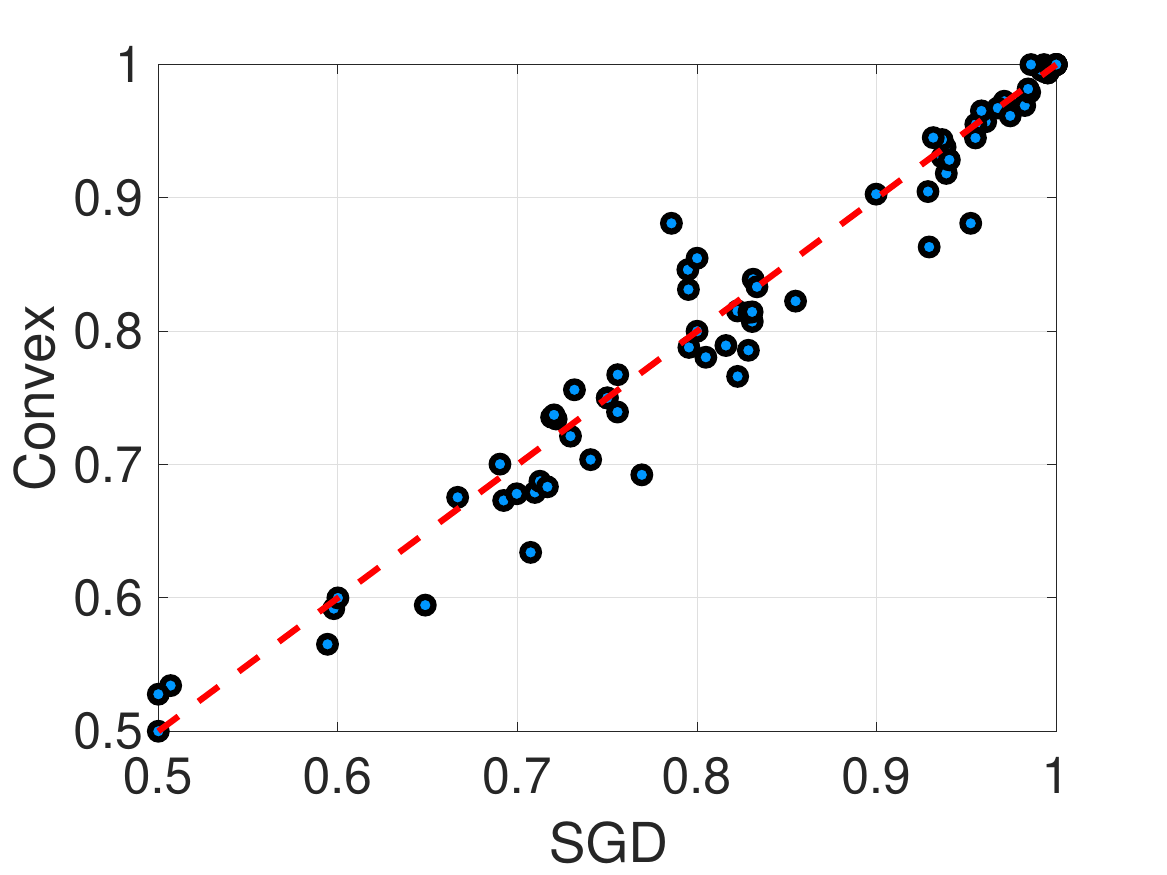}
	\caption{Test accuracy-$m=200$\centering} \label{fig:uci_acc200}
\end{subfigure}
	\begin{subfigure}[t]{0.32\textwidth}
	\centering
	\includegraphics[width=1\textwidth, height=0.8\textwidth]{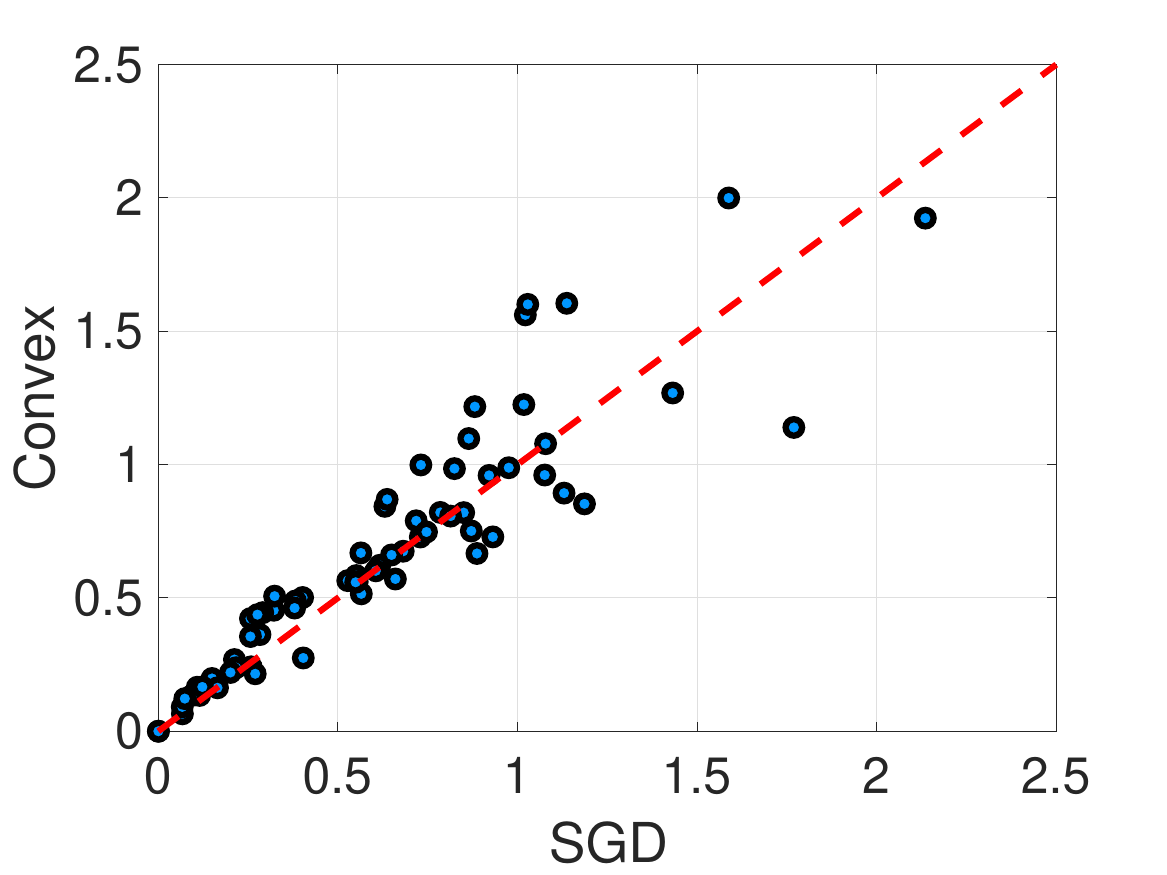}
	\caption{Test error-$m=10$\centering} \label{fig:uci_err10}
\end{subfigure} \hspace*{\fill}
	\begin{subfigure}[t]{0.32\textwidth}
	\centering
	\includegraphics[width=1\textwidth, height=0.8\textwidth]{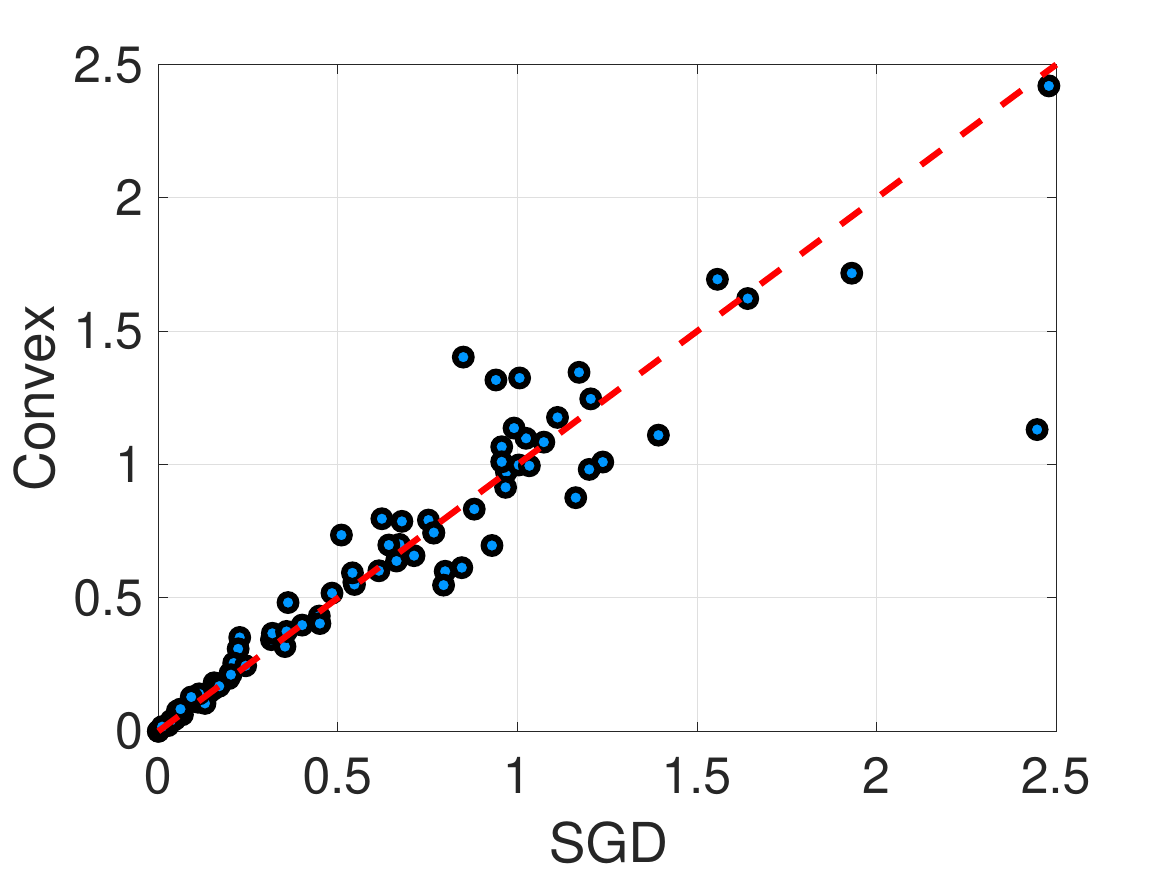}
	\caption{Test error-$m=50$\centering} \label{fig:uci_err50}
\end{subfigure} \hspace*{\fill}
	\begin{subfigure}[t]{0.32\textwidth}
	\centering
	\includegraphics[width=1\textwidth, height=0.8\textwidth]{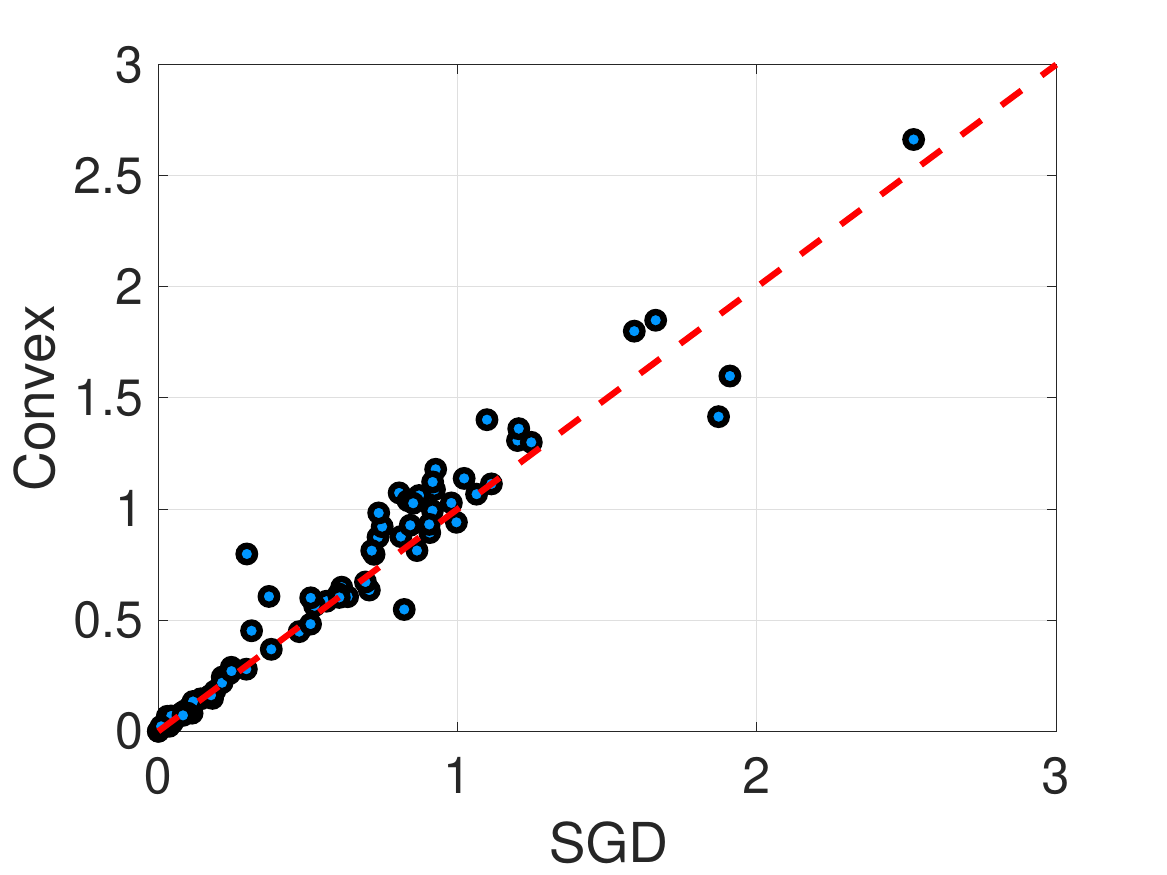}
	\caption{Test error-$m=200$\centering} \label{fig:uci_err200}
\end{subfigure} \hspace*{\fill}

\caption{Test accuracy and error values of a two-layer ReLU network trained with SGD and the convex program in \eqref{eq:twolayerconvexprogram} on the UCI datasets with $\beta=10^{-3}$, where each blue dot denotes a certain dataset and the corresponding axis values represent the performance of training algorithms on the dataset. }\label{fig:uci}
\end{figure*}

\begin{figure*}[ht]
\centering
\captionsetup[subfigure]{oneside,margin={1cm,0cm}}
	\begin{subfigure}[t]{0.32\textwidth}
	\centering
	\includegraphics[width=1\textwidth, height=0.8\textwidth]{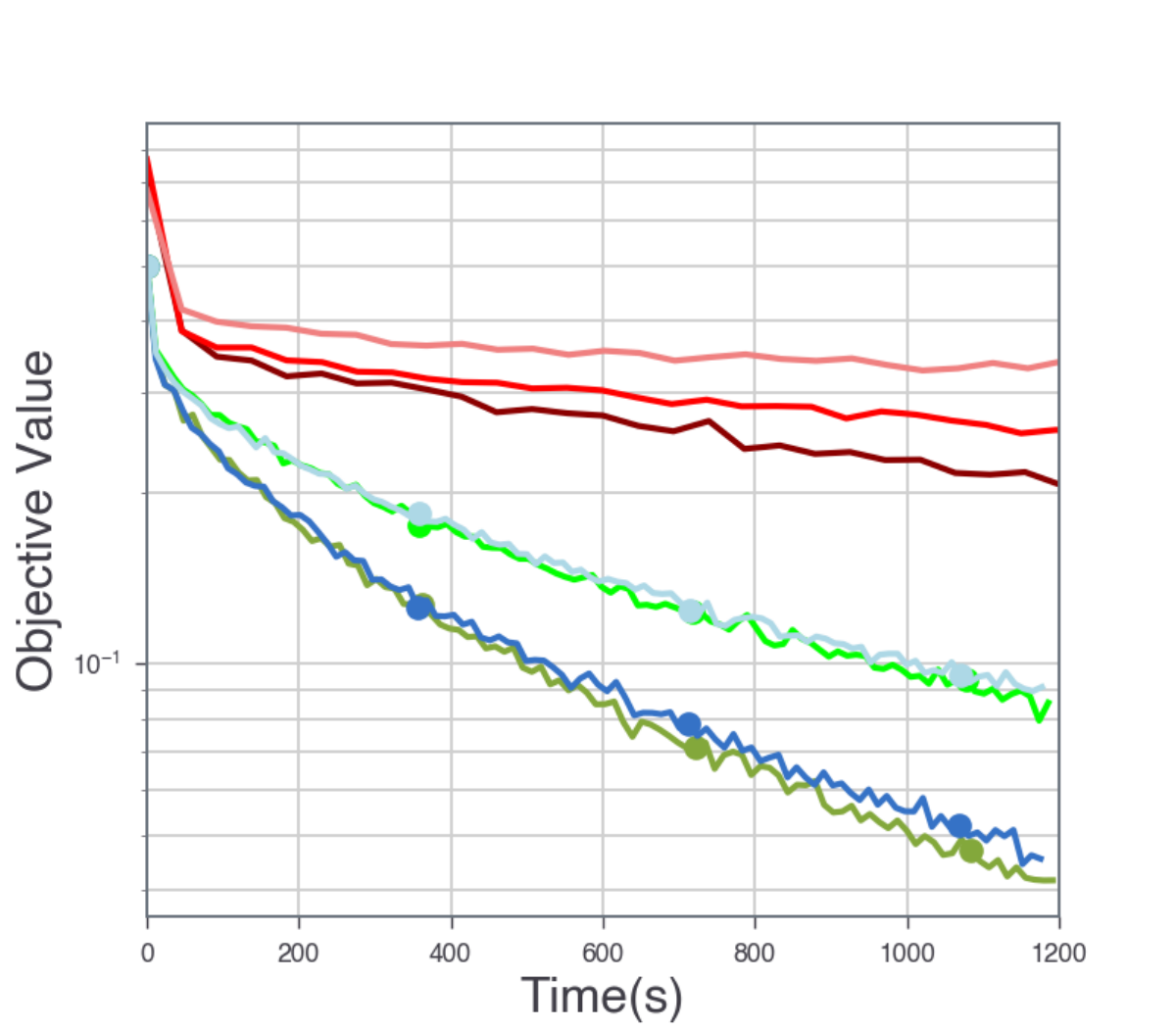}
	\caption{Objective value for different learning rates ($\mu$)\centering} \label{fig:stepsize_comp_obj_sgd}
\end{subfigure} \hspace*{\fill}
	\begin{subfigure}[t]{0.32\textwidth}
	\centering
	\includegraphics[width=1\textwidth, height=0.8\textwidth]{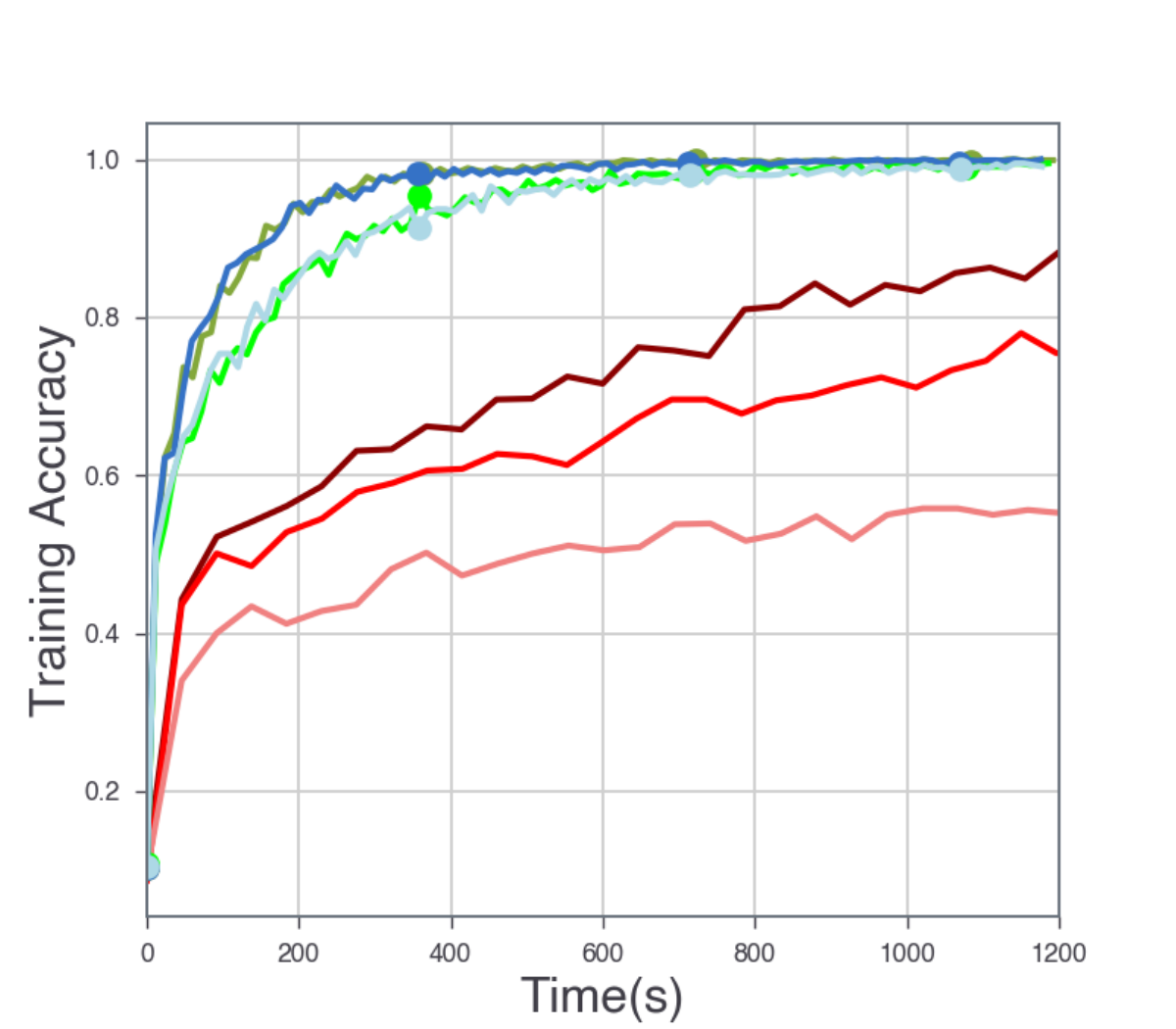}
	\caption{Training accuracy (10-class) for different learning rates ($\mu$)\centering} \label{fig:stepsize_comp_trainingacc_sgd}
\end{subfigure} \hspace*{\fill}
	\begin{subfigure}[t]{0.32\textwidth}
	\centering
	\includegraphics[width=1\textwidth, height=0.8\textwidth]{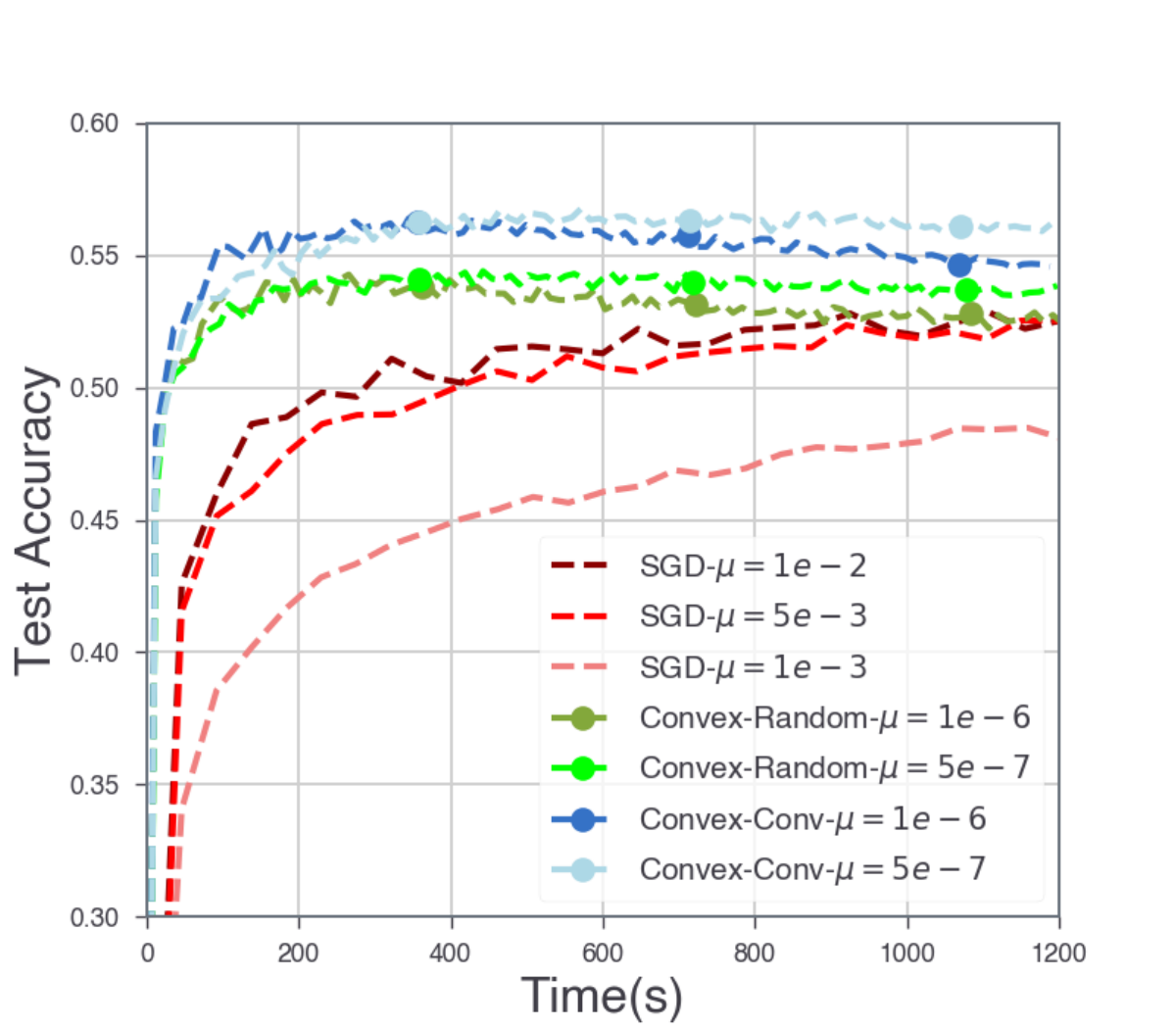}
	\caption{Test accuracy (10-class) for different learning rates ($\mu$)\centering} \label{fig:stepsize_comp_testacc_sgd}
\end{subfigure} \hspace*{\fill}
\caption{Comparison of the methods on the CIFAR-10 dataset, where $(n,d,m,C,\beta)=(50000,3072,4096,10,10^{-3})$, batch size is $1000$, $P=m$, and the activation function is ReLU. The proposed convex optimization problem is solved using SGD. Here, we use solid and dashed lines for training and test results, respectively. }\label{fig:cifar_multi_comp_sgd}
\end{figure*}

\begin{figure*}[ht]
\centering
\captionsetup[subfigure]{oneside,margin={1cm,0cm}}
	\begin{subfigure}[t]{0.32\textwidth}
	\centering
	\includegraphics[width=1\textwidth, height=0.8\textwidth]{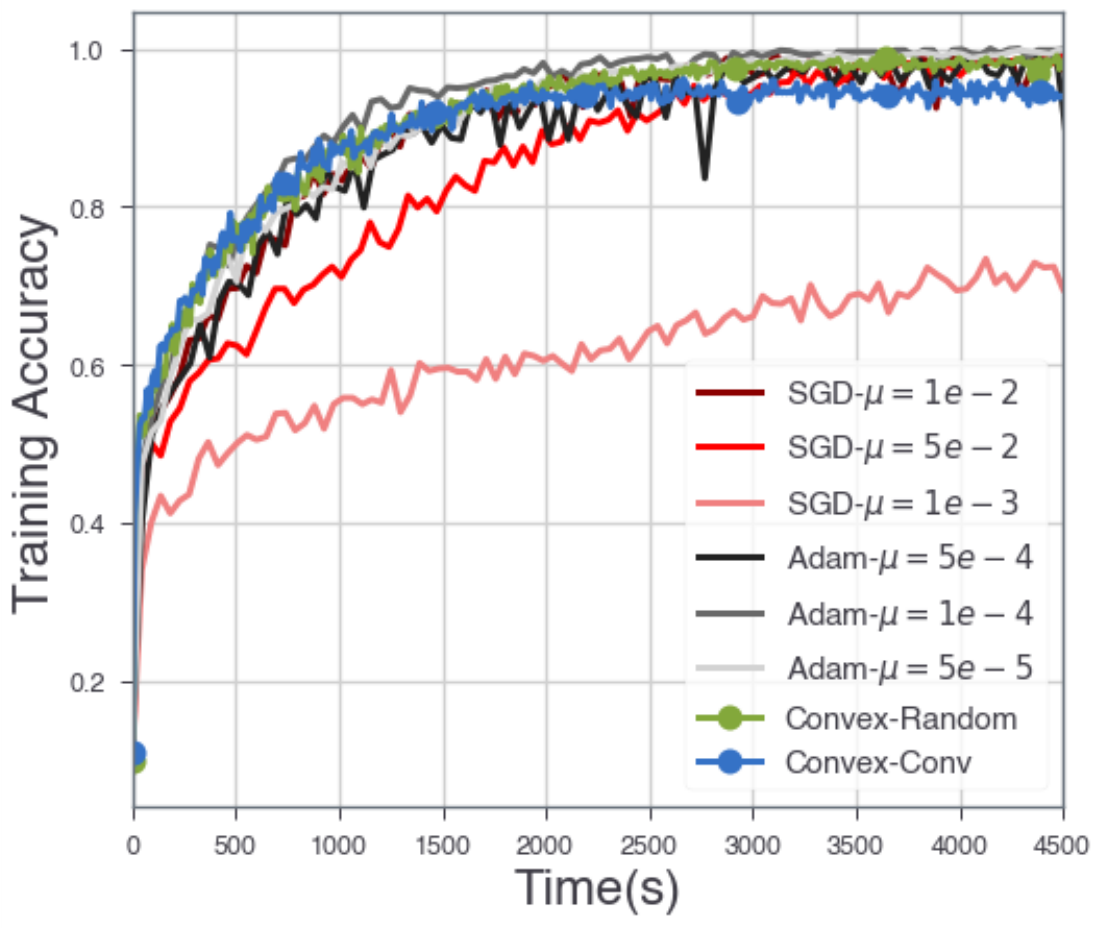}
	\caption{Training accuracy (10-class) for different learning rates ($\mu$)\centering} \label{fig:stepsize_comp_training}
\end{subfigure} \hspace*{\fill}
	\begin{subfigure}[t]{0.32\textwidth}
	\centering
	\includegraphics[width=1\textwidth, height=0.8\textwidth]{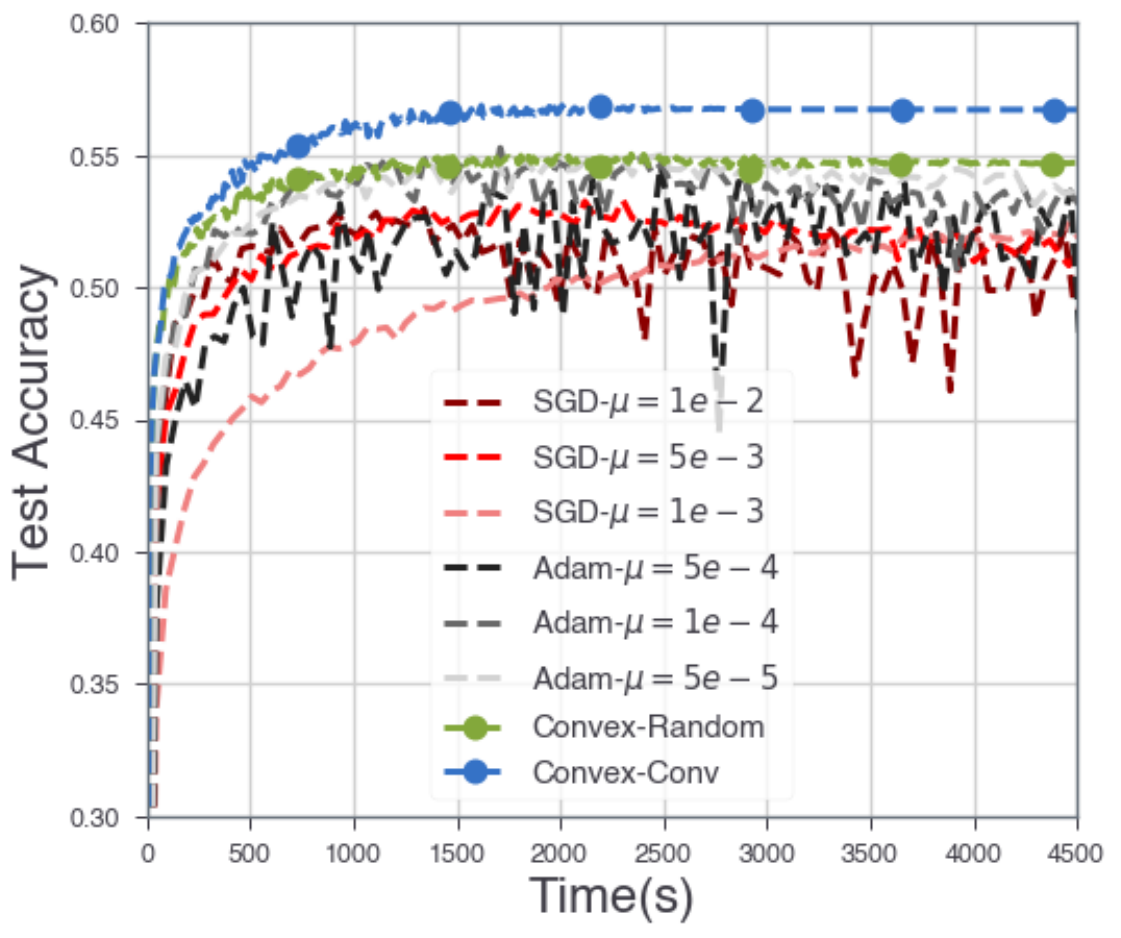}
	\caption{Test accuracy (10-class) for different learning rates ($\mu$)\centering} \label{fig:stepsize_comp_test}
\end{subfigure} \hspace*{\fill}
	\begin{subfigure}[t]{0.32\textwidth}
	\centering
	\includegraphics[width=1\textwidth, height=0.8\textwidth]{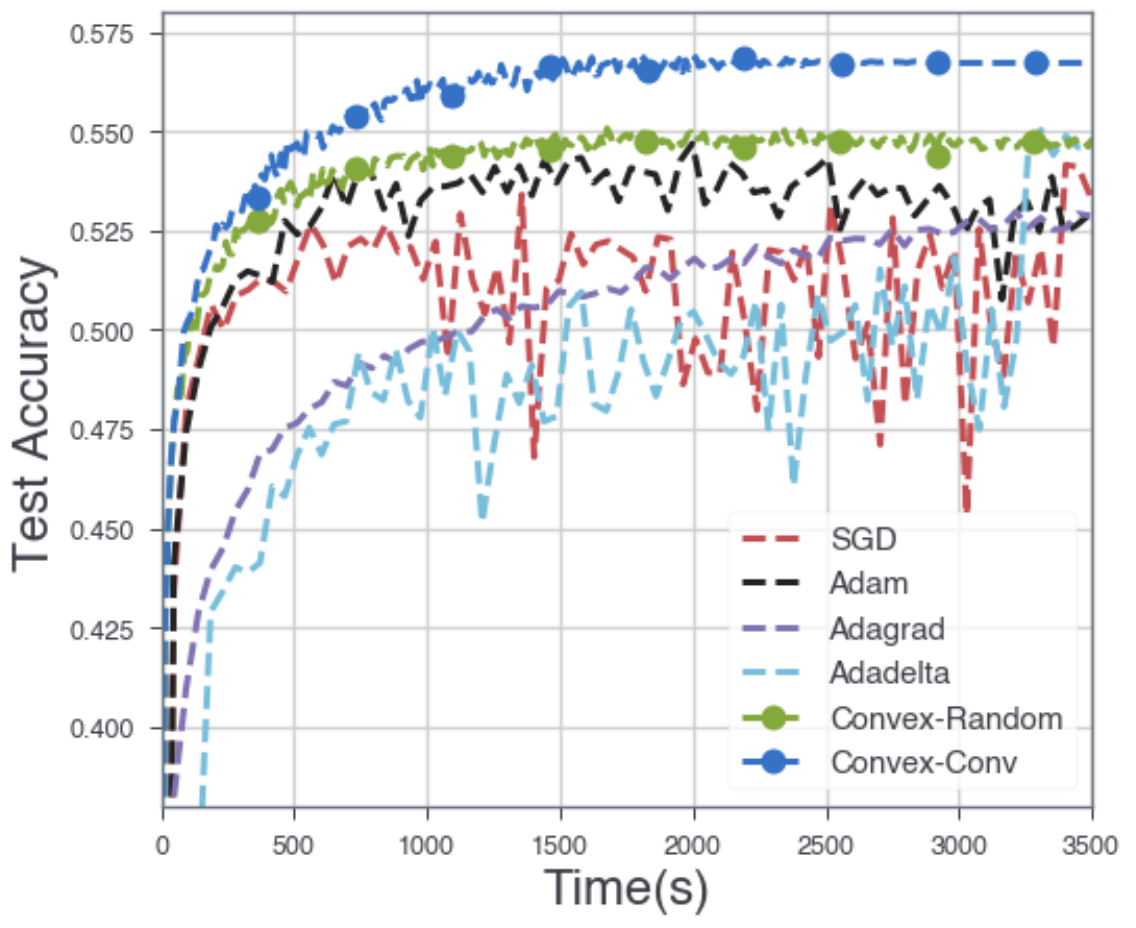}
	\caption{Test accuracy (10-class) for different optimizers\centering} \label{fig:method_comp}
\end{subfigure} \hspace*{\fill}
\caption{Comparison of the methods on the CIFAR-10 dataset, where $(n,d,m,C,\beta)=(50000,3072,4096,10,10^{-3})$, batch size is $1000$, $P=m$, and the activation function is ReLU. The proposed convex optimization problem is solved using Adagrad. Here, we use solid and dashed lines for training and test results, respectively. }\label{fig:cifar_multi_comp2}
\end{figure*}

\begin{figure*}[!htb]
\centering
\captionsetup[subfigure]{oneside,margin={1cm,0cm}}
	\begin{subfigure}[t]{0.4\textwidth}
	\centering
	\includegraphics[width=1\textwidth, height=0.8\textwidth]{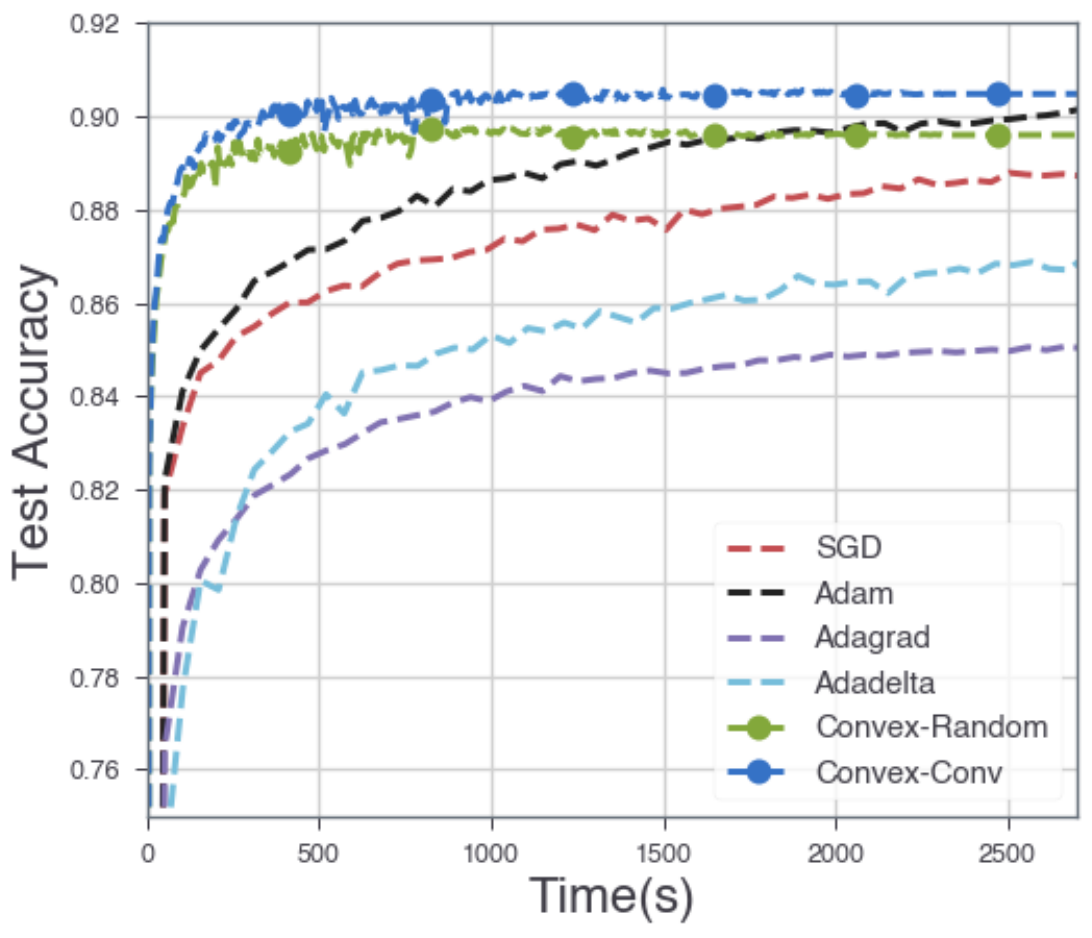}
	\caption{Fashion MNIST (10-class)\centering} \label{fig:fmnist_multi_acc_comp}
\end{subfigure} \hspace*{\fill}
	\begin{subfigure}[t]{0.4\textwidth}
	\centering
	\includegraphics[width=1\textwidth, height=0.8\textwidth]{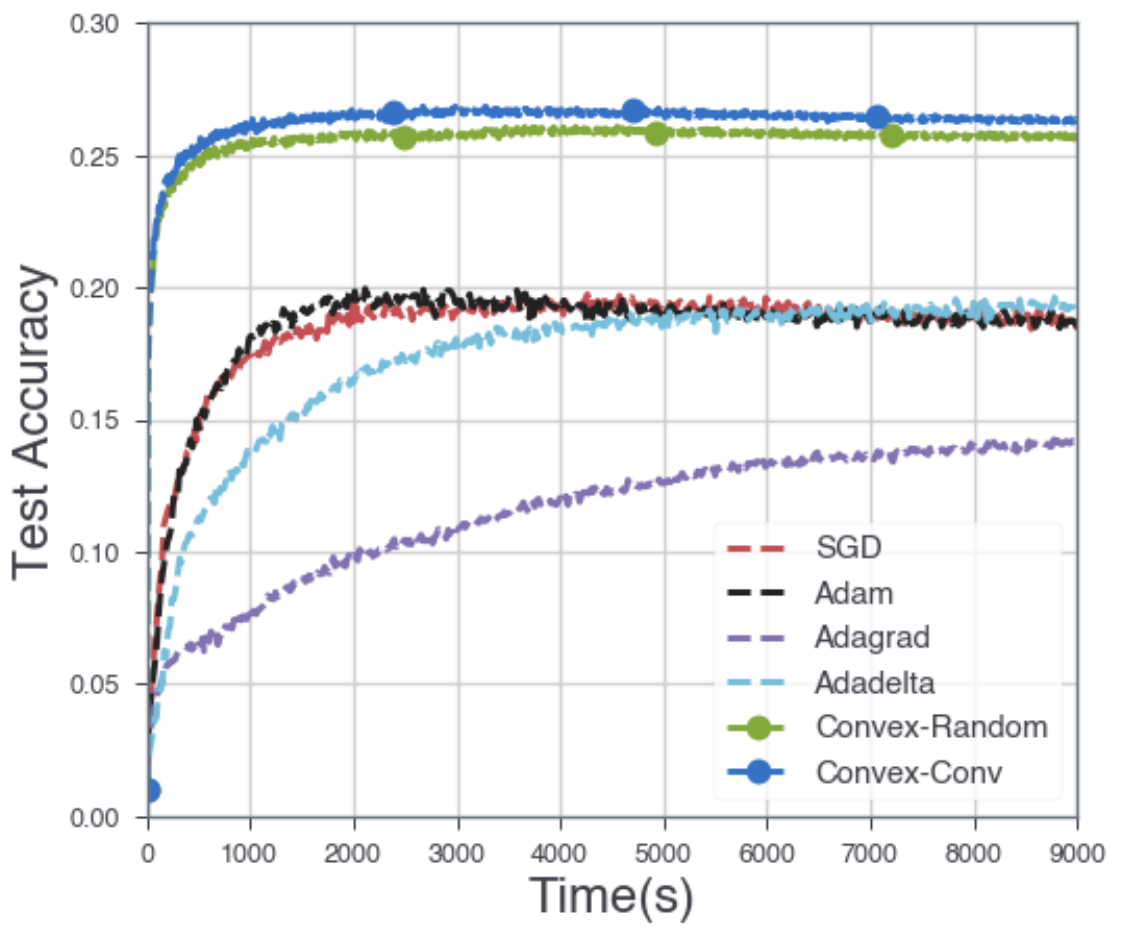}
	\caption{CIFAR-100 (100-class)\centering} \label{fig:cifar100_multi_acc_comp}
\end{subfigure} \hspace*{\fill}
\caption{Comparison of the methods on Fashion MNIST with $(n,d,m,C,\beta)=(60000,784,4096,10,10^{-3})$ and CIFAR-100 with $(n,d,m,C,\beta)=(50000,3072,512,100,10^{-3})$, where batch size is $1000$, $P=m$, and the activation function is ReLU for both datasets. We use Adam to solve the proposed convex optimization problem. }\label{fig:fmnistcifar100_multi_comp}
\end{figure*}

\begin{table*}
  \caption{Highest test accuracies achieved by 1-layer Neural Network (NN), which is the conventional logistic regression method, 2-layer NN trained via the standard non-convex approaches, 2-layer NN trained via the proposed convex approaches, and 2-layer NN trained on a data matrix preprocessed via K-means clustering algorithm (see Algorithm \ref{alg:convex_kmeans} for the pseudocode)}
  \label{tab:acc}
  \centering
   \resizebox{1\columnwidth}{!}{%
   \begin{tabular}{l|c|c|c}
    \toprule
    & \textbf{CIFAR-10}  & \textbf{Fashion MNIST} & \textbf{CIFAR-100}     \\
    \cmidrule(r){1-4}
    \textbf{1-layer NN (Logistic regression)} & 0.4076 &  0.8392  &  0.0939  \\
   \textbf{2-layer NN (non-convex)} & 0.5416 & 0.9002 & 0.1995  \\
    \textbf{2-layer NN (convex)} &0.5688& 0.9057 &  0.2684 \\
    \textbf{2-layer NN (preprocessing+non-convex)} & 0.7770 &  0.9260 & 0.4771 \\
   \textbf{2-layer NN (preprocessing+convex)}   & \textbf{0.8163} & \textbf{0.9327} & \textbf{0.5393} \\
    \bottomrule
  \end{tabular}}
\end{table*}

\section{Numerical experiments} \label{sec:numerical}
In this section\footnote{We provide the details about our experimental setup and additional experiments in \ref{sec:supp_exps}.}, we present numerical experiments to verify our theoretical results. We start with a one-dimensional toy dataset with $n=5$ given by $\data=[-2 \; -1 \; 0 \; 1 \; 2]^T$ and $y=[1 \;-1 \; 1 \; 1 \; -1]^T$, where we include a bias term by concatenating a column of ones to the data matrix $\data$. We then train a two-layer ReLU network with SGD and the proposed convex program using squared loss. In Figure \ref{fig:sgd_1d}, we plot the value of the regularized objective function with respect to the iteration index for SGD in $10$ independent trials for initial parameters. We solve the convex program in \eqref{eq:twolayerconvexprogram} via CVX \cite{cvx} and plot the objective value as a horizontal dashed line denoted as ``Convex''. Additionally, we repeat the same experiment for the different number of neurons: $m=8,15$, and $50$. As demonstrated in the figure, SGD is likely to get stuck at local minima when the number of neurons is small. As we increase $m$, the number of trials that successfully converge to global minima gradually increases. We also note that Convex achieves the optimal objective value as claimed in the previous sections.

We also compare the prediction performance of neural network training algorithms on a time series prediction problem, where we use the ECG data in \cite{ecg_data}. For each sample $\labelscalar_i$, we consider the previously observed three samples as our features, i.e., $\datavec_i =[\labelscalar_{i-1},\, \labelscalar_{i-2},\,\labelscalar_{i-3}]^T$ and consider predicting the value $\labelscalar_i$. Therefore, we obtained a time series dataset with $n=2393$ and $d=3$. In Figure \ref{fig:ecg}, we plot the training objective in \eqref{eq:twolayer_objective_generic} and test predictions, where we use a batch size of $100$ for SGD. In addition, we also experiment with different learning rates $\mu$ as demonstrated in \ref{fig:ecg_obj}. Here, we observe that the SGD trials fail to achieve the optimal training objective value obtained by our convex optimization method. Consequently, SGD also exhibits poor predictive performance in the test set as shown in Figure \ref{fig:ecg_pred}.

Next, we present numerical experiments performed on several datasets taken from UCI machine learning repository \cite{uci_repository}. In particular, we consider small/medium scale datasets used in \cite{Arora2020Harnessing} and then follow the same preprocessing steps. Specifically, we use $90$ UCI datasets with the number of samples less than 5000. For each of these datasets, we use a conventional regression framework with squared loss and then plot the test accuracy and error obtained by SGD and Convex in Figure \ref{fig:uci}. Similar to Figure \ref{fig:sgd_1d}, as the number of neurons $m$ increases, the performance gap between SGD and Convex closes, and the distribution of data points approaches a line with slope one.

We also perform experiments on some well-known image classification datasets, namely CIFAR-10, CIFAR-100, and Fashion-MNIST \cite{cifar10,fashionmnist}. For all of these experiments, we use the convex program in Theorem \ref{theo:mainconvex_vector_l1}, where the problem decomposes into $C$ independent problems for a network with $C$ outputs. Moreover, we use the approximate version of the convex program, where the hyperplane arrangements are sampled randomly as discussed in Section \ref{sec:efficient_arrangement}. We sample hyperplane arrangements using a normal distribution and denote this approach as ``Convex-Random''. We also randomly generate convolutional filters and use their sign patterns as hyperplane arrangements for the convex program, which is denoted as ``Convex-Conv''. In addition, we apply K-Means based preprocessing as proposed in \cite{coates2012learning,coates2011analysis} to the raw data matrix to obtain a richer set of features, which are presented as preprocessing+convex and preprocessing+non-convex in Table \ref{tab:acc} (see Algorithm \ref{alg:convex_kmeans} in \ref{sec:supp_exps} for the full description of the algorithm). We first consider a ten class classification problem on CIFAR-10 with the parameters $(n,d,m,C,\beta)=(50000,3072,4096,10,10^{-3})$, batch size of $1000$, and the ReLU activation. In Figure \ref{fig:cifar_multi_comp_sgd}, we compare these two approaches against SGD with different learning rates ($\mu$) and demonstrate the superior performance of our convex models in terms of objective value, training, and test accuracies. Among the convex models, we observe that Convex-Conv substantially improves upon Convex-Random. In addition, preprocessing+convex yields $\sim 25\%$ accuracy improvement compared to other convex models (see Table \ref{tab:acc}). Furthermore, we compare our convex models against the non-convex formulation trained with different optimizers in Figure \ref{fig:cifar_multi_comp2}. Here, our convex models achieve better training and test performance compared to the non-convex methods. Similarly, we also validate the performance of the proposed convex model on Fashion MNIST with $(n,d,m,C,\beta)=(60000,784,4096,10,10^{-3})$ and CIFAR-100 with $(n,d,m,C,\beta)=(50000,3072,512,100,10^{-3})$, where the batch size is $1000$ and the activation is ReLU. For Fashion-MNIST, even though the convex models again achieve higher test accuracies compared to the non-convex ones in Figure \ref{fig:fmnistcifar100_multi_comp}, Adam also provides comparable performance. However, for CIFAR-100 (with $C=100$), we observe a notable accuracy improvement with respect to the non-convex approaches. 

\section{Conclusion}
We studied two-layer neural network architectures with piecewise linear activations and introduced a convex optimization framework to analyze the regularized training problem. We derived exact convex optimization formulations for the original non-convex training problem, which can be globally optimized by convex solvers with polynomial-time complexity. These convex representations reside in a higher dimensional space, where the data matrix is partitioned over all possible hyperplane arrangements and group sparsity or low-rankness is enforced via group $\ell_p$, $\ell_1$ or nuclear norm regularizers. In addition, our results show that the form of the structural regularization induced on the weights of the convex model is a function of the architecture, the number of outputs, and the regularization in the non-convex problem. We believe that this result sheds light into the generalization of neural network models and their architectural bias, which are extensively studied in the recent literature. {Our results show that neural networks with piecewise linear activations can be seen as parsimonious piecewise linear models. We believe that this perspective offers a clearer interpretation of these non-convex models, as their convex counterparts are more transparent and easier to understand.  Moreover, due to convexity, the equivalent training problems do not require non-convex optimization heuristics or extensive hyperparameter searches such as choosing a proper learning rate schedule and initialization scheme.}. We showed that randomly sampling hyperplane arrangements and solving the subsampled convex problem works extremely well in practice. Furthermore, we proposed an approximation algorithm that leverages low-rank approximation of the data matrix such that the equivalent convex program can be globally optimized with polynomial-time complexity in terms of all the problem parameters, i.e., the number of samples $n$, the feature dimension $d$, and the number of neurons $m$. We also proved strong approximation bounds for this algorithm.

Our work poses multiple promising open problems to explore. First, one can obtain a better understanding of neural networks, their optimization landscapes, and their generalization properties by leveraging our equivalent convex formulations. In the light of our results, backpropagation can be viewed as a heuristic method to solve the convex program. Moreover, the loss landscape of the non-convex objective and the dynamics of gradient based optimizers can be further investigated by utilizing the optimal set of the convex program. After our work, this was explored in \cite{lacotte2020local}, where the authors reported interesting results regarding the hidden convex landscape of the non-convex objective. Furthermore, one can extend our convex optimization framework to various other architectures, e.g., CNNs, recurrent networks, transformers, and autoencoders. Here, we extended our approach to certain simple CNNs. Recently, \cite{ergen2020cnn} further extended our approach to CNNs with ReLU activations and various pooling strategies. Similarly, based on this work, a series of follow-up papers analyzed deep linear networks \cite{ergen2021revealing}, generative networks \cite{vikul2021generative,sahiner2021gan}, deep ReLU networks \cite{ergen2021deep}, and transformer networks \cite{sahiner2022transformer, ergen2022transformer} via convex duality. In addition, \cite{ergen2021batchnorm} analyzed Batch Normalization, which is a popular heuristic to stabilize the training of deep neural networks via our convex methodology. Finally, to the best of our knowledge, this work provides the first polynomial-time training algorithm to \emph{globally} train two-layer neural network architectures for any data matrix with fixed rank. We conjecture that more efficient solvers for the convex program can be developed for larger scale experiments by utilizing the connection to sparse models \cite{Donoho06,CandesTao05}.

\section*{Acknowledgements}
This work was partially supported by the National Science Foundation under grants ECCS-2037304 and DMS-2134248, an Army Research Office Early Career Award W911NF-21-1-0242, and the ACCESS – AI Chip Center for Emerging Smart Systems, sponsored by InnoHK funding, Hong Kong SAR.


\bibliography{references}
\bibliographystyle{plainnat}

\newpage
\appendix

\addcontentsline{toc}{section}{Appendix} 

\part{Appendix} 
\parttoc 

\section{Details about our experimental setup and additional numerical results}\label{sec:supp_exps}
\begin{algorithm}[!h]
		\caption{\texttt{Convex neural network training via K-means feature embeddings}}
		\begin{algorithmic}[1]
        \STATE{Set $P_c,\epsilon,h,k$ (in our experiments $(P_c,\epsilon,h,k)=(4\text{x} 10^{5},0.1,6,9)$)}
        \STATE{Randomly extract $P_c$ patches of size $h \times h$ from the dataset: $\{\vec{p}_i\}_{i=1}^{P_c}$}
        \FOR{$i=1:P_c$}
        \STATE{Normalize the patch:} $\vec{\bar{p}}_i=\frac{\vec{p}_i-\text{mean}(\vec{p}_i)}{\sqrt{\text{var}(\vec{p}_i)+\epsilon}}$
        \ENDFOR
        \STATE{Form a patch matrix: $\vec{P}=[\vec{\bar{p}}_1 \, \ldots\, \vec{\bar{p}}_{P_c}]$}
        \STATE{Apply ZCA whitening to the patch matrix: \begin{align*}
        &[\vec{V},\vec{D}]=\text{eig}(\text{cov}(\vec{P})) \\
        &\vec{\tilde{P}}= \vec{V}(\vec{D}+\epsilon \vec{I})^{-\frac{1}{2}}\vec{V}^T \vec{P}\end{align*}}
        \STATE{Cluster patches using K-means as in \cite{coates2012learning, coates2011analysis} to obtain $m$ cluster means: $\{\vec{c}_j\}_{j=1}^m $ }
        \FOR{$i=1:n$}
        \STATE{Extract all the patches of size $h \times h$ in the image $\data_i \in \mathbb{R}^{d \times d}$ : $\data_{ip} \in \mathbb{R}^{h^2 \times (d-h+1)^2}$}
        \STATE{Compute pairwise distances between patches and cluster means and then threshold the distances: $\vec{K}_{dist} \in \mathbb{R}^{(d-h+1)^2 \times m}$}
        \STATE{Threshold the distances as: $\bar{\vec{K}}_{dist}=\max\{\vec{K}_{dist}-\vec{1}\bm{m}^T,0\}$, where $\bm{m}$ is a vector of means for each row of $\vec{K}_{dist}$}
        \STATE{Apply $k \times k$ pooling (with stride $k$) on the reshaped data of size $(d-h+1) \times (d-h+1) \times m$: $\vec{Q}=\text{pooling}(\bar{\vec{K}}_{dist})$}
        \STATE{Flatten the resulting vector: $\bar{\vec{x}}_i=\text{flatten}(\vec{Q}) \in \mathbb{R}^{d_{new}}$}
        \ENDFOR
        \STATE{Form a new data matrix consisting of $\{\bar{\vec{x}}_i\}_{i=1}^n$: $\bar{\vec{X}} \in \mathbb{R}^{n \times d_{new}}$}
        \STATE{Solve the convex training problem in \eqref{eq:twolayerconvexprogram_uncons} using $\bar{\vec{X}}$ }
		\end{algorithmic}\label{alg:convex_kmeans}	
	\end{algorithm}
In this section, we provide detailed information about our experimental setup.

We note that for the synthetic experiment in Figure \ref{fig:low_rank}, we obtain the data labels $\vec{y} \in \mathbb{R}^n$ by first forward propagating the input data matrix through a randomly initialized two-layer ReLU network with five neurons and then adding a noise term. Particularly, we first randomly generate the layer weights as $\firstw_j \sim N(\vec{0},\vec{I}_d)$ and $\secondw_j \sim N(0,1), \; \forall j \in [5]$ and then obtain the labels as $\vec{y}=\relu{\data \firstwmat}\secondwvec+0.1\bm{\epsilon}$, where $\bm{\epsilon} \sim N(\vec{0},\vec{I}_n)$.

For small scale experiments in Figure \ref{fig:ecg} and \ref{fig:uci}, we use CVX \cite{cvx} and CVXPY \cite{cvxpy,cvxpy_rewriting} with the SDPT3 solver \cite{tutuncu2001sdpt3} to solve convex optimization problems in \eqref{eq:twolayerconvexprogram} and \eqref{eq:twolayerconvexprogram_vector_l1}. Moreover, the training is performed on a CPU with 50GB of RAM. For ECG and UCI experiments, we use the $66\%-34\%$, $60\%-40\%$ splitting ratio for the training and test sets. Moreover, the learning rate of SGD is tuned via a grid-search on the training split. Specifically, we try different values and choose the best performing learning rate on the validation datasets. 

For the image classification experiments in Figure \ref{fig:cifar_multi_comp_sgd}, \ref{fig:cifar_multi_comp2}, and \ref{fig:fmnistcifar100_multi_comp}, we use a GPU with 50GB of memory. In particular, to solve the convex optimization problems in \eqref{eq:twolayerconvexprogram}, we first introduce an equivalent unconstrained convex problem as follows
\begin{align}\label{eq:twolayerconvexprogram_uncons} 
&\min_{ \weight  \in \mathcal{C}(\data)}\,  \mathcal{L}(\dataf\weight ,\labelvec)  +\beta \sum_{i=1}^{2P}\|\weight_i\|_{2} +\rho \vec{1}^T \sum_{i=1}^{P} \left(\relu{-(2\diag_i -\vec{I}_n)\weight_i}+\relu{-(2\diag_i -\vec{I}_n)\weight_{i+P}}\right)  
\end{align}
where $\rho >0$ is a trade-off parameter. Now, since the equivalent problem in \eqref{eq:twolayerconvexprogram_uncons} is an unconstrained convex optimization problem, we can directly optimize its parameters using standard first order optimizers such as SGD and Adam. Therefore, we can use PyTorch to optimize both the non-convex objective in \eqref{eq:twolayer_objective_generic} and the convex objective in \eqref{eq:twolayerconvexprogram_uncons} on the larger scale datasets, e.g., CIFAR-10, CIFAR-100, and Fashion-MNIST. For the learning rates, we again follow the same grid-search technique. In addition, for all the experiments, we set the trade-off parameter to $\rho=0.01$.

\begin{figure*}[ht]
\centering
\captionsetup[subfigure]{oneside,margin={1cm,0cm}}
	\begin{subfigure}[t]{0.45\textwidth}
	\centering
	\includegraphics[width=1\textwidth, height=0.8\textwidth]{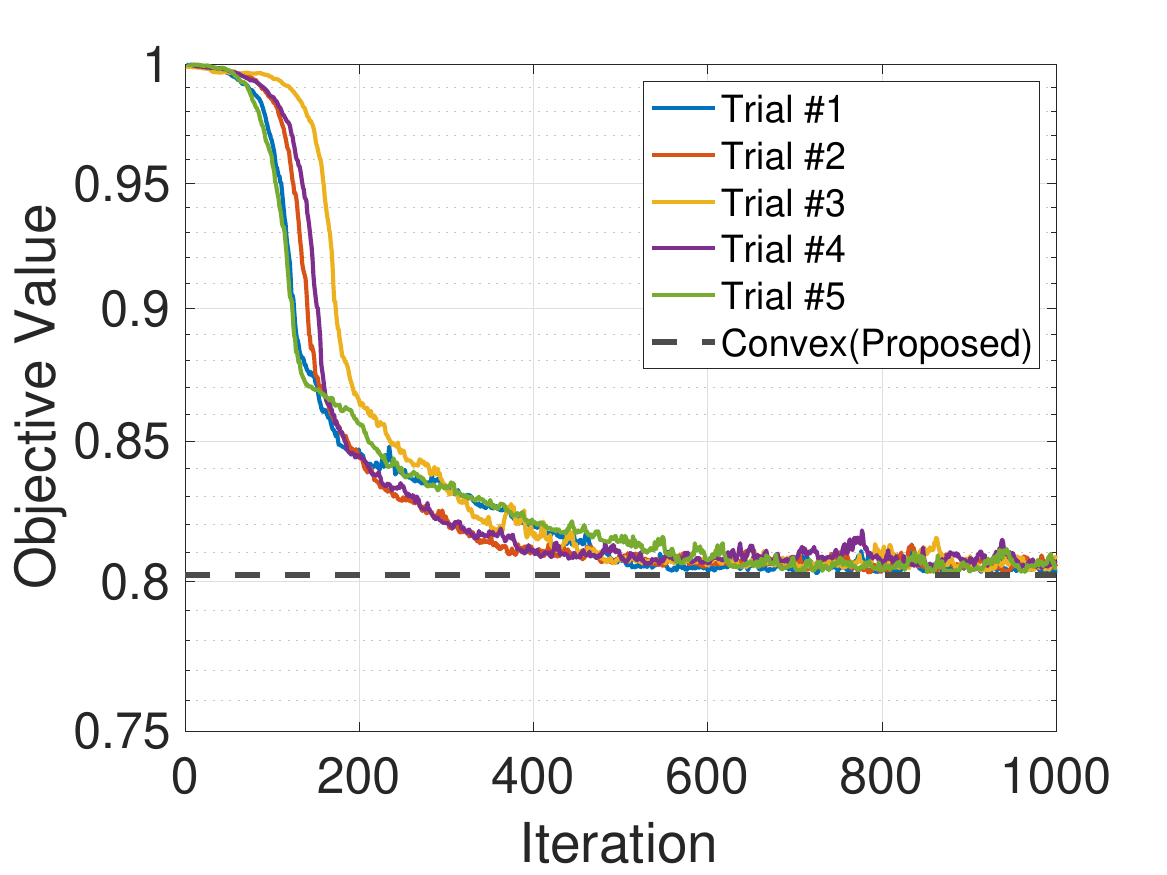}
	\caption{Objective value\centering} \label{fig:linear_cnn_obj}
\end{subfigure} \hspace*{\fill}
	\begin{subfigure}[t]{0.45\textwidth}
	\centering
	\includegraphics[width=1\textwidth, height=0.8\textwidth]{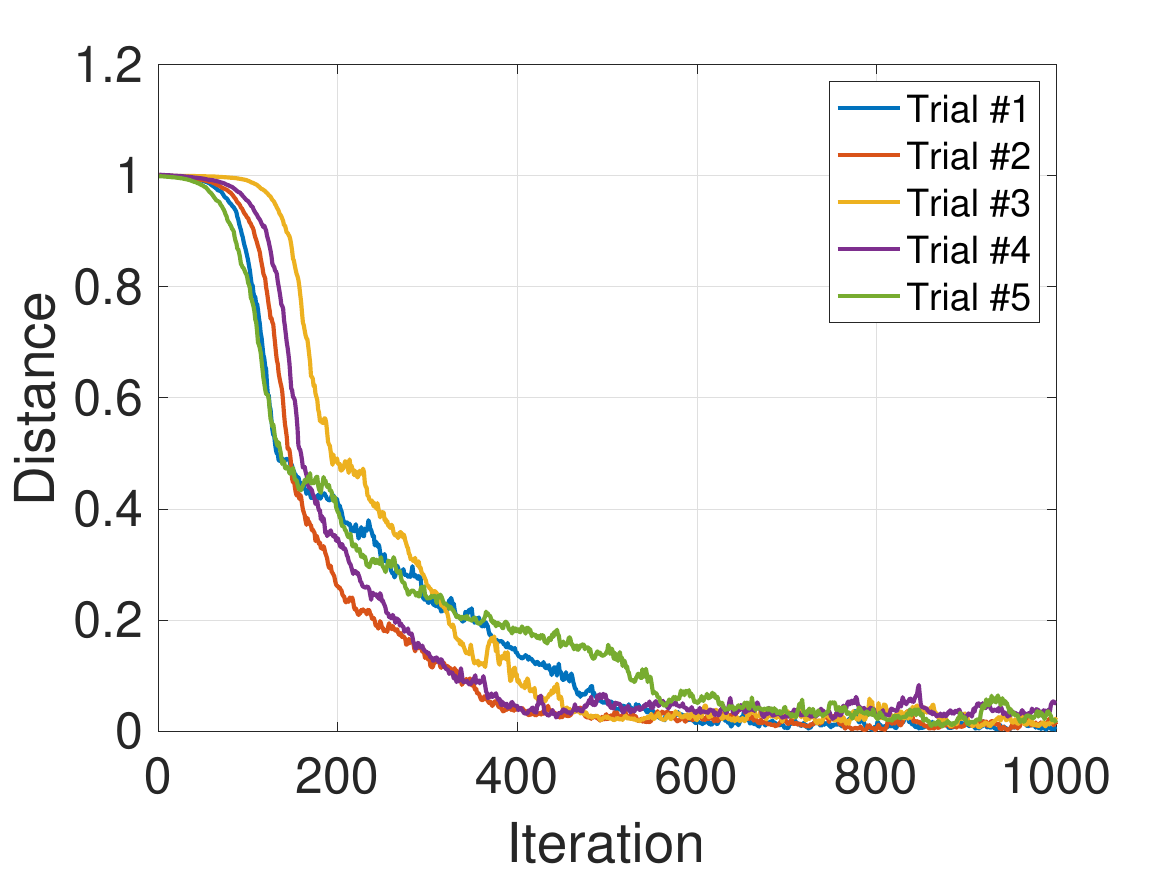}
	\caption{Distance to the solution of the convex program \centering} \label{fig:linear_cnn_distance}
\end{subfigure} \hspace*{\fill}
\caption{Training accuracy of a two-layer linear CNN trained with SGD (5 initialization trials) on a subset of CIFAR-10, where Convex denotes the proposed convex program in \eqref{eq:linear_cnn_l1}. Filters found via SGD converge to the solution of \eqref{eq:linear_cnn_l1}.}\label{fig:linear_cnn}
\vskip -0.2in
\end{figure*}

Finally, we train a two-layer linear CNN architecture on a subset of CIFAR-10, where we denote the proposed convex program in \eqref{eq:linear_cnn_l1} as Convex. In Figure \ref{fig:linear_cnn}, we plot both the objective value and the Euclidean distance between the filters found by GD and Convex for $5$ independent realizations with $n=387$, $m=30$, $h=10$, and batch size of $60$. In this experiment, all the independent realizations converge to the objective value obtained by Convex and find almost the same filters with Convex.

\section{Proof of Lemma \ref{lemma:scaling} }
{We first note that similar observations are also made in the previous studies \cite{pilanci2020neural,ergen2021revealing,parhi_minimum,infinite_width,tibshirani2021equivalences,neyshabur_reg}.}

For any $\theta \in \Theta$, we can rescale the parameters as ${\firstwb}_{j}=\gamma_j\firstw_{j}$ and ${\secondwb}_{j}= \secondw_{j}/\gamma_j$, for any $\gamma_j>0$. Then, the network output becomes
\begin{align*}
    f_{\bar{\theta}}(\data)= \sum_{j=1}^m\actt{\data {\firstwb}_{j}} {\secondwb}_{j}= \sum_{j=1}^m\actt{\data \firstw_{j}\gamma_j} \frac{\secondw_{j}}{\gamma_j}=\sum_{j=1}^m\actt{\data \firstw_{j}} \secondw_{j}= f_{\theta}(\data),
\end{align*}
which proves $ f_{\bar{\theta}}(\data)=f_{\theta}(\data)$. In addition to this, given $p\geq 1$, we have the following basic inequality
\begin{align*}
   \frac{1}{2} \sum_{j=1}^m (\|\firstw_j\|_p^2+{|\secondw_j}|^2) \geq \sum_{j=1}^m (\| \firstw_{j}\|_p|\secondw_j| \text{ }),
\end{align*}
where the equality is achieved with the scaling choice $\gamma_j=\big(\frac{|\secondw_{j}|}{\| \firstw_{j}\|_p}\big)^{\frac{1}{2}}$ is used. Since the scaling operation does not change the right-hand side of the inequality, we can set $\|\firstw_{j} \|_p=1, \forall j \in [m]$. Therefore, the right-hand side becomes $\| \secondwvec\|_1$.

Now, let us consider a modified version of the problem, where the unit norm equality constraint is relaxed as $\| \firstw_{j} \|_p \leq 1$. Let us also assume that for a certain index $j$, we obtain  $\| \firstw_{j} \|_p < 1$ with $\secondw_{j}\neq 0$ as an optimal solution. This shows that the unit norm inequality constraint is not active for $\firstw_{j}$, and hence removing the constraint for $\firstw_{j}$ will not change the optimal solution. However, when we remove the constraint, $\| \firstw_{j}\|_p \rightarrow \infty$ reduces the objective value since it yields $\secondw_j=0$. Therefore, we have a contradiction, which proves that all the constraints that correspond to a nonzero $\secondw_j$ must be active for an optimal solution. This also shows that replacing $\|\firstw_{j}\|_p=1$ with $\| \firstw_{j}\|_p \leq 1$ does not change the solution to the problem.

\section{Convex duality for two-layer networks}\label{sec:appendix_convexduality}
Now we introduce our main technical tool for deriving convex representations of the non-convex objective function \eqref{eq:twolayer_objective_generic}. We start with the $\ell_1$ penalized representation, which is equivalent to \eqref{eq:twolayer_objective_generic}
\begin{align}
     p^*=\min_{\theta \in \Theta_s} \mathcal{L}(f_{\theta}(\data),\labelvec) +\beta \|\secondwvec\|_1 \,.
    \label{eq:2layer_regularized_cost_l1}
\end{align}
Replacing the minimization problem for the output layer weights $\secondwvec$ with its convex dual, we obtain (see Appendix \ref{sec:twolayer_dualform_appendix})
\begin{align*}
     p^*=\min_{\firstw_j \in \ball_2}\max_{\vec{\dual}} - \mathcal{L}^*(\dual)   \text{ s.t. } \left\vert \dual^T \actt{\data \firstw_j} \right\vert\leq \beta,\, \forall j \in [m],\,.
\end{align*}
Interchanging the order of $\min$ and $\max$, we obtain the lower-bound $d^*$ via weak duality
\begin{align}
     p^*\ge d^*:= \max_{\vec{\dual}}\min_{\substack{ \theta \in \Theta_s}} - \mathcal{L}^*(\dual)  \text{ s.t. } \max_{\firstw \in \ball_2}\left\vert \dual^T \actt{\data \firstw_j} \right\vert\leq \beta.
    \label{eq:2layer_regularized_cost_innerdual}
\end{align}
The above problem is a convex \emph{semi-infinite} optimization problem with $n$ variables and infinitely many constraints. We will show that strong duality holds, i.e., $p^*=d^*$, as long as the number of hidden neurons $m$ satisfies $m\ge m^*$ for some $m^*\in \mathbb{N}$, $ m^* \le n+1$, which will be defined in the sequel. As it is shown, $m^*$ can be smaller than $n+1$.  
 The dual of the dual program \eqref{eq:2layer_regularized_cost_innerdual} can be derived using standard semi-infinite programming theory \cite{semiinfinite_goberna}, and corresponds to the bi-dual of the non-convex problem \eqref{eq:twolayer_objective_generic}.

 Now we briefly introduce basic properties of signed measures that are necessary to state the dual of \eqref{eq:2layer_regularized_cost_innerdual} and refer the reader to \cite{rosset2007,bach2017breaking} for further details. Consider an arbitrary measurable input space $\mathcal{X}$ with a set of continuous basis functions $\actinf_{\firstw}:\mathcal{X}\rightarrow \reals$ parameterized by $\firstw\in \ball_2$. We then consider real-valued Radon measures equipped with the uniform norm \cite{Rudin}. For a signed Radon measure $\bm{\mu}$, we can define an infinite width neural network output for the input $\datavec\in\mathcal{X}$ as $f(\datavec) = \int_{\firstw\in \ball_2} \actinf_\firstw(\datavec) d\bm{\mu}(\firstw)$\,. The total variation norm of the signed measure $\bm{\mu}$ is defined as the supremum of $\int_{\firstw \in\ball_2}q(\firstw)d\bm{\mu}(\firstw)$ over all continuous functions $q(\firstw)$ that satisfy $|q(\firstw)|\le 1$. Consider the basis functions $\actinf_{\firstw}(\datavec)=\act(\datavec^T\firstw)$. We may express networks with finitely many neurons as in \eqref{eq:twolayer_network} by
\begin{align*}
    f(\datavec) = \sum_{j=1}^m \actinf_{\firstw_j}(\datavec) \secondw_j\,,
\end{align*}
which corresponds to $\bm{\mu} = \sum_{j=1}^m \secondw_j \delta(\firstw-\firstw_j)$ where $\delta$ is the Dirac delta measure. And the total variation norm $\|\bm{\mu}\|_{TV}$ of $\bm{\mu}$ reduces to the $\ell_1$-norm $\|\secondwvec\|_1$.

We state the dual of \eqref{eq:2layer_regularized_cost_innerdual} (see Section 2 of \cite{shapiro2009semi} and Section 8.6 of \cite{semiinfinite_goberna}) as follows
\begin{align}
    d^*\le p_{\infty}^*=\min_{\mu} \mathcal{L}\left(\int_{\firstw\in \ball_2} \act(\data\firstw)d\bm{\mu}(\firstw) ,\labelvec \right) + \beta\, \|\bm{\mu}\|_{TV}. \label{eq:infdim}
\end{align}
%
Furthermore, an application of Caratheodory's theorem shows that the infinite dimensional bi-dual \eqref{eq:infdim} always has a solution that is supported with $m^*$ Dirac deltas, where $m^*\le n+1$ \cite{rosset2007}. Therefore, we have
\begin{align*}
     p_{\infty}*&=\min_{\substack{ \firstw_j \in \ball_2\\ \{\secondw_j,\firstw_j\}_{j=1}^{m^*}}} \mathcal{L}\left(\sum_{j=1}^{m^*} \act(\data \firstw_j ) \secondw_j ,\labelvec \right) +\beta \|\secondwvec\|_1\,, 
    \\
    &=p^*\,,
\end{align*}
as long as $m\ge m^*$. We show that strong duality holds, i.e., $d^*=p^*$ in Appendix \ref{appendix_semi_infinite_duality} and \ref{appendix_semi_infinite_duality2}. In the sequel, we illustrate how $m^*$ can be determined via a finite dimensional parameterization of \eqref{eq:2layer_regularized_cost_innerdual} and its dual.
%

\subsubsection{A geometric insight: neural gauge function}
\begin{figure*}[t]
	\centering
	\captionsetup[subfigure]{oneside}
		\begin{subfigure}[t]{0.32\textwidth}
		\centering
		\includegraphics[width=0.95\textwidth, height=0.9\textwidth]{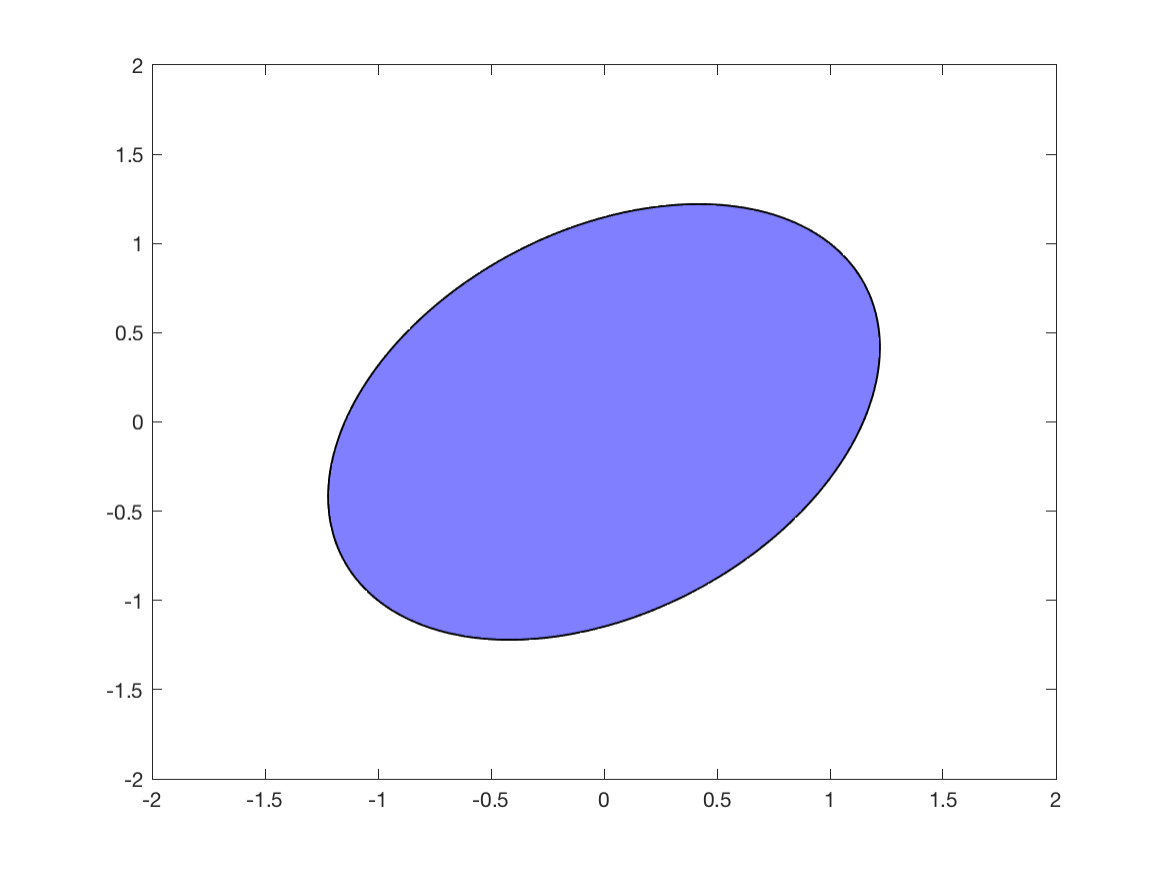}
		\caption{Ellipsoidal set: $\{\data \firstw : \firstw \in \mathbb{R}^d, \|  \firstw\|_2 \leq 1\}$\centering} \label{fig:spike2_ellipsoid}
	\end{subfigure} \hspace*{\fill}
		\begin{subfigure}[t]{0.32\textwidth}
		\centering
		\includegraphics[width=0.95\textwidth, height=0.9\textwidth]{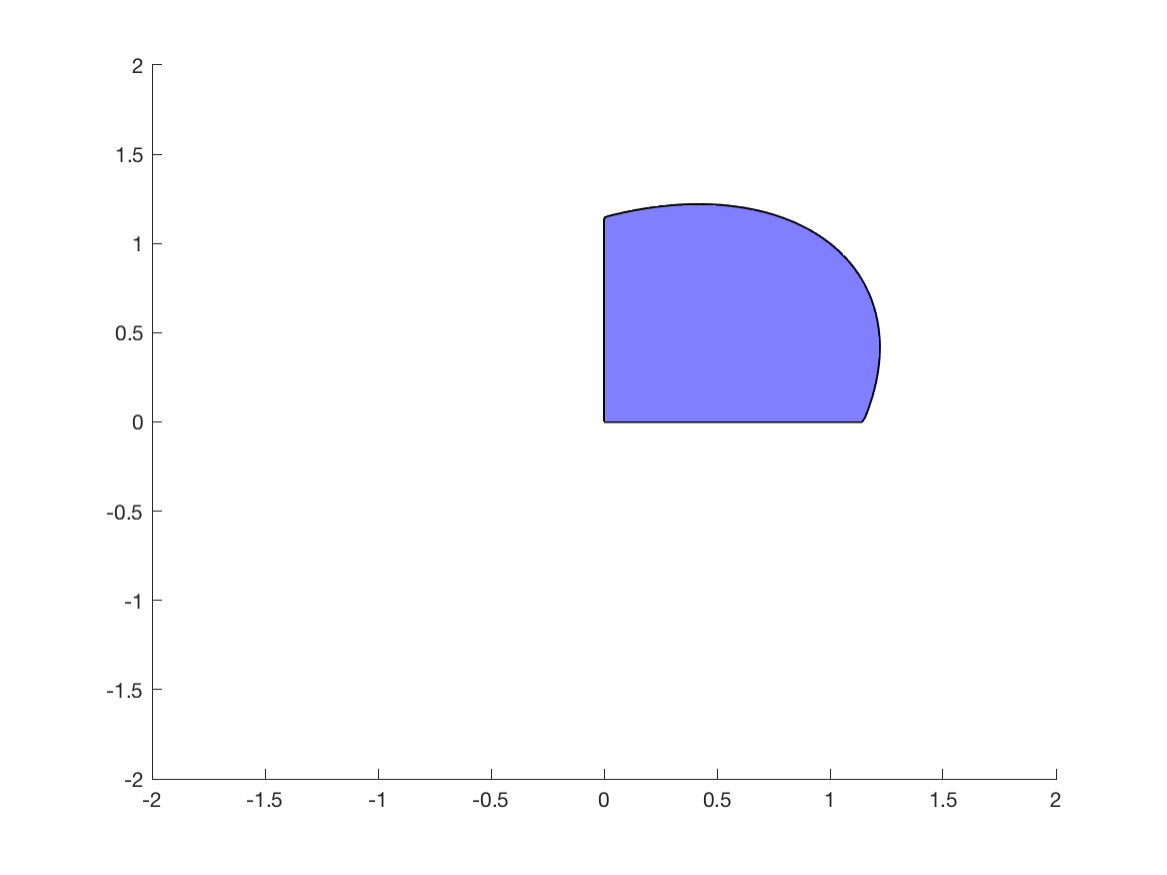}
		\caption{Rectified ellipsoidal set $\rectset$: $\big \{ \act( \data  \firstw ) : \firstw \in \mathbb{R}^d, \|  \firstw\|_2 \leq 1   \big\}$\centering} \label{fig:spike2_recellipsoid}
	\end{subfigure} \hspace*{\fill}
			\begin{subfigure}[t]{0.32\textwidth}
		\centering
		\includegraphics[width=0.95\textwidth, height=0.9\textwidth]{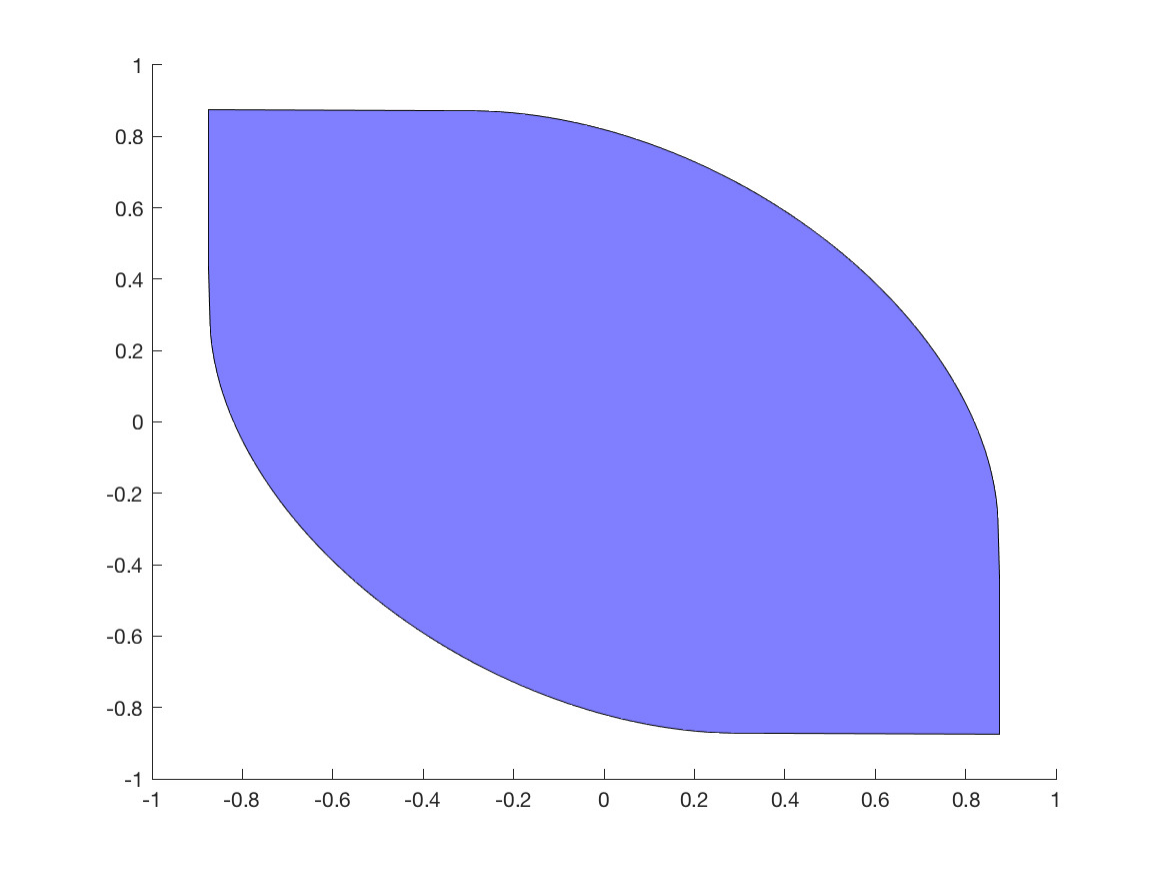}
		\caption{Polar set $(\rectset\cup-\rectset)^\circ$: $\{\dual : \vert\, \dual^T \weight \vert \le 1\,, \forall \weight\in \rectset\}$\centering} \label{fig:spike2_polar}
	\end{subfigure} \hspace*{\fill}
	\medskip
	\caption{Sets involved in the construction of the Neural Gauge. Ellipsoidal set, rectified ellipsoid $\rectset$ and the polar of $\rectset\cup-\rectset$.}
\label{fig:spike2}
	\end{figure*}

An interesting geometric insight can be provided in the weakly regularized case where $\beta \rightarrow 0$. In this case, minimizers of \eqref{eq:2layer_regularized_cost_l1} and hence \eqref{eq:twolayer_objective_generic} approach minimum-norm interpolation $p_{\beta\rightarrow 0}^*:=\lim_{\beta\rightarrow 0} \beta^{-1}p^*$ given by
\begin{align}
    p^*_{\beta\rightarrow 0}=&\min_{\{\firstw_j,\secondw_j\}_{j=1}^m} \sum_{j=1}^m \vert \secondw_j \vert \mbox{ s.t. } \sum_{j=1}^m \act(\data \firstw_j)\secondw_j=\labelvec,\, \firstw_j \in \ball_2 \,\forall j.\textbf{} \nonumber
\end{align}
We show that $p^*_{\beta\rightarrow 0}$ is the gauge function of the convex hull of $\rectset \cup -\rectset$ where $\rectset:=\{ \act(\data \firstw)\,:\,\firstw\in \ball_2 \}$ (see Appendix \ref{sec:gauge_appendix}), i.e.,
\begin{align*}
    p^*_{\beta\rightarrow 0} = &\inf_{t:t\ge 0} \,\, t \mbox{ s.t. }
     \labelvec \in t \,\mbox{Conv}\{{\rectset \cup -\rectset\}}\,,
\end{align*}
which we call \emph{Neural gauge} due to the connection to the minimum-norm interpolation problem. Using classical polar gauge duality (see e.g. \cite{Rockafellar}, it holds that
\begin{align}\label{eq:supportfun}
    p^*_{\beta\rightarrow 0}= &\max \labelvec^T \vec{z}  \mbox{ s.t. } \vec{z} \in (\rectset \cup -\rectset)^\circ \,,
\end{align}
where $(\rectset \cup -\rectset)^\circ$ is the polar of the set $\rectset \cup -\rectset$. Therefore, evaluating the support function of this polar set is equivalent to solving the neural gauge problem, i.e., minimum-norm interpolation  $p^*_{\beta\rightarrow 0}$. These sets are illustrated in Figure \ref{fig:spike2}. Note that the polar set $(\rectset \cup -\rectset)^\circ$ is always convex (see Figure \ref{fig:spike2}c),  which also appears in the dual problem $\eqref{eq:2layer_regularized_cost_innerdual}$ as a constraint. In particular, $\lim_{\beta\rightarrow 0} \beta^{-1}d^*$ is equal to the support function. Our finite dimensional convex program leverages the convexity and an efficient description of this set as we discuss next.

\section{Proof of Theorem \ref{theo:mainconvex}}
\label{sec:proofthm1}
We now prove the main result two-layer networks. We start with the dual representation derived in Section \ref{sec:convex_duality}
\begin{align}
\label{eq:dual}
&\max_{\dual} \, -\mathcal{L}^*(\dual) \mbox{ s.t.}\, \max_{\firstw \in \ball_2} \, \vert \dual^T \act(\data \firstw) \vert \le \beta\,. 
\end{align}
Note that the constraint \eqref{eq:dual} can be represented as
\begin{align*}
\bigg\{\dual: \max_{\firstw \in \ball_2}   \dual^T \act(\data \firstw)  \le \beta \bigg\}  \bigcap \bigg\{\dual:\max_{\firstw \in \ball_2}   -\dual^T \act(\data \firstw)  \le \beta  \bigg\}.
\end{align*}

We now focus on a single-sided dual constraint
\begin{align}
\max_{\firstw \in \ball_2} \, \dual^T \act(\data \firstw) \le \beta, \label{eq:dualconst}
\end{align}
by considering hyperplane arrangements and a convex duality argument over each partition.
We first partition $\real^d$ into the following subsets
\begin{align*}
P_S := \{\firstw\,:\, \datavec_i^T \firstw &\ge 0, \forall i \in S,\,\,
\datavec_j^T \firstw  \le 0, \forall j \in S^c \}.
\end{align*}
Let $\mathcal{H}$ be the set of all hyperplane arrangement patterns for the matrix $\data$, defined as the following set
\begin{align*}
\mathcal{H} = \bigcup \big \{ \{\sign(\data \firstw)\} \,:\, \firstw \in \real^d \big \}\,.
\end{align*}
It is obvious that the set $\mathcal{H}$ is bounded, i.e., $\exists N_H \in \mathbb{N} <\infty$ such that $\vert \mathcal{H} \vert \le N_H$. We next define an alternative representation of the sign patterns in $\mathcal{H}$, which is the collection of sets that correspond to positive signs for each element in $\mathcal{H}$. More precisely, let
\begin{align*}
\mathcal{S} := \big\{ \{ \cup_{h_i=1} \{i\}  \} \,:\, \vec{h} \in \mathcal{H} \big\}.
\end{align*}
We also define a new diagonal matrix $\hat{\diag}(S)\in \real^{n\times n}$ as
\begin{align*}
\hat{\diag}(S)_{ii} := \begin{cases} 1 & \mbox{ if } i \in S\\
0 & \mbox{ otherwise}  \end{cases}.
\end{align*}
Note that $\hat{\diag}(S^c) = \vec{I}_n-\hat{\diag}(S)$, since $S^c$ is the complement of the set $S$.
With this notation, we can represent $P_S$ as
\begin{align*}
P_S &= \{ \firstw\,:\, \hat{\diag}(S)\data \firstw \ge 0,\,\, (\vec{I}_n-\hat{\diag}(S))\data\firstw\le 0 \}\\&= \{ \firstw\,:\,  (2\hat{\diag}(S)-\vec{I}_n)\data\firstw\ge 0 \}\, .
\end{align*}
We now express the maximization in the dual constraint in \eqref{eq:dualconst} over all possible hyperplane arrangement patterns as
\begin{align*}
\max_{\firstw \in \ball_2} \, \dual^T \act(\data \firstw) &= \max_{\firstw \in \ball_2} \dual^T \diag(S) \data \firstw\\
& = ~~\max_{\substack{S \subseteq [n]\\ S \in \mathcal{S} }}\, \max_{\substack{\firstw \in \ball_2 \\ \datavec_i^T \firstw \ge 0 \,\, \forall i \in S \\ \, \, \, \datavec_j^T \firstw \le 0 \,\, \forall j \in S^c}} \dual^T\diag(S) \data \firstw\\
& = ~~\max_{\substack{S \subseteq [n]\\ S \in \mathcal{S} }} \max_{\substack{\firstw \in \ball_2 \cap P_S}} \dual^T\diag(S) \data \firstw,
\end{align*}
where
\begin{align*}
\diag(S)_{ii} := \begin{cases} 1 & \mbox{ if } i \in S\\
\kappa & \mbox{ otherwise}  \end{cases}.
\end{align*}
We also note that since $\kappa<0.5$, $P_S$ can be equivalently represented as
\begin{align*}
P_S &= \{ \firstw\,:\,  (2{\diag}(S)-\vec{I}_n)\firstw\ge 0 \}\, .
\end{align*}
%
%
%
Enumerating all hyperplane arrangements $\mathcal{H}$, or equivalently $\mathcal{S}$, we index them in an arbitrary order via $i \in [\vert \mathcal{S} \vert]$. We denote $P=\vert \mathcal{S}\vert$. Hence, $S_1,\ldots,S_P \in \mathcal{S}$ is the list of all $P$ elements of $\mathcal{S}$. 
Next we use the strong duality result from Lemma \ref{lem:finitestrongdualconst} for the inner maximization problem.
The dual constraint \eqref{eq:dualconst} can be represented as 
\begin{align*}
\eqref{eq:dualconst} &\iff \forall i\in [P],\,  \min_{\substack{\alpha, \beta \in \real^{n} \\ \alpha, \beta \ge 0}}
\| \data^T \diag(S_i) \dual + \data^T\hat{\diag}(S_i)(\bm{\alpha} + \bm{\gamma} )-\data^T \bm{\gamma} \|_2 \le \beta\\
& \iff \forall i\in [P],\, \exists \bm{\alpha}_i,\bm{\gamma}_i \in {\real^n} \mbox{ s.t.}  \quad \bm{\alpha}_i,\bm{\gamma}_i \ge 0,\;  \| \data^T \diag(S_i) \dual + \data^T\hat{\diag}(S_i)(\bm{\alpha}_i + \bm{\gamma}_i )-\data^T \bm{\gamma}_i \|_2 \le \beta\,.
\end{align*}
Therefore, recalling the two-sided constraint in \eqref{eq:dual}, we can represent the dual optimization problem in \eqref{eq:dual} as a finite dimensional convex optimization problem with variables $\dual \in \real^n, \bm{\alpha}_i,\bm{\gamma}_i,\bm{\alpha}_i^\prime,\bm{\gamma}_i^\prime \in  {\real^n}, \forall i \in [P]$, and $2P$ second order cone constraints as follows
\begin{align*}
\max_{\substack{\dual\in \real^n \\ \bm{\alpha}_i,\bm{\gamma}_i \in {\real^n}\\\bm{\alpha}_i, \bm{\gamma}_i\ge 0, \, \forall i \in [P] \\ \bm{\alpha}_i^\prime,\bm{\gamma}_i^\prime \in {\real^n}\\\bm{\alpha}_i^\prime, \bm{\gamma}_i^\prime\ge 0, \, \forall i \in [P] }}\, -\mathcal{L}^*(\dual) \mbox{ s.t.}\, &\| \data^T \diag(S_1) \dual + \data^T\hat{\diag}(S_1)(\bm{\alpha}_1 + \bm{\gamma}_1 )-\data^T \bm{\gamma}_1 \|_2 \le \beta \nonumber\\ 
& \quad \vdots \nonumber\\
& \| \data^T \diag(S_P) \dual + \data^T\hat{\diag}(S_P)(\bm{\alpha}_P + \bm{\gamma}_P )-\data^T \bm{\gamma}_P \|_2 \le \beta \nonumber\\
&  \| -\data^T \diag(S_1) \dual + \data^T \hat{\diag}(S_1)(\bm{\alpha}_1^\prime + \bm{\gamma}_1^\prime )-\data^T \bm{\gamma}_1^\prime \|_2 \le \beta \nonumber\\
& \quad \vdots \nonumber\\
&  \| -\data^T \diag(S_P) \dual + \data^T \hat{\diag}(S_P)(\bm{\alpha}_P^\prime + \bm{\gamma}_P^\prime )-\data^T \bm{\gamma}_P^\prime \|_2 \le \beta. \nonumber\\
\end{align*}
The above problem can be represented as a standard finite dimensional second order cone program. Note that the particular choice of parameters $\dual=\bm{\alpha}_i=\bm{\gamma}_i=\bm{\alpha}_i^\prime=\bm{\gamma}_i^\prime=0$, $\forall i \in [P]$, are strictly feasible in the above constraints as long as $\beta>0$. Therefore Slater's condition and consequently strong duality holds \cite{Boyd02}. The dual problem \eqref{eq:dual} can be written as
\begin{align*}
\min_{\substack{\bm{\lambda}, \bm{\lambda}^\prime \in \real^P\\ \bm{\lambda},\bm{\lambda}^\prime\ge 0}} \, \max_{\substack{\dual\in \real^n \\ \bm{\alpha}_i,\bm{\gamma}_i \in {\real^n}\\\bm{\alpha}_i, \bm{\gamma}_i\ge 0, \, \forall i \\ \bm{\alpha}_i^\prime,\bm{\gamma}_i^\prime \in {\real^n}\\\bm{\alpha}_i^\prime, \bm{\gamma}_i^\prime\ge 0, \, \forall i }} -&\mathcal{L}^*(\dual)  + \sum_{i=1}^P \lambda_i \big(\beta - \| \data^T \diag(S_i) \dual + \data^T \hat{\diag}(S_i)(\bm{\alpha}_i + \bm{\gamma}_i )-\data^T \bm{\gamma}_i \|_2 \big)\\
  &+ \sum_{i=1}^P \lambda_i^\prime \big(\beta - \| -\data^T \diag(S_i) \dual + \data^T \hat{\diag}(S_i)(\bm{\alpha}_i^\prime + \bm{\gamma}_i^\prime )-\data^T \bm{\gamma}_i^\prime\|_2 \big). 
\end{align*}
Next, we introduce variables $\vec{r}_1,\ldots,\vec{r}_P, \vec{r}_1^\prime,\ldots,\vec{r}_P^\prime \in\real^d$ and represent the dual problem \eqref{eq:dual} as
\begin{align*}
\min_{\substack{\bm{\lambda},\bm{\lambda}^\prime \in \real^P\\ \bm{\lambda},\bm{\lambda}^\prime\ge 0}} \, \max_{\substack{\dual\in \real^n \\ \bm{\alpha}_i,\bm{\gamma}_i \in {\real^n}\\\bm{\alpha}_i, \bm{\gamma}_i\ge 0, \,\forall i  \\ \bm{\alpha}_i^\prime,\bm{\gamma}_i^\prime \in {\real^n}\\\bm{\alpha}_i^\prime, \bm{\gamma}_i^\prime\ge 0, \, \forall i }}\, \min_{\substack{\vec{r}_i \in \real^d,\, \|\vec{r}_i\|_2\le 1\\ \vec{r}_i^\prime \in \real^d,\, \|\vec{r}_i^\prime\|_2\le 1\\ \forall i } } -\mathcal{L}^*(\dual)  &+ \sum_{i=1}^P \bm{\lambda}_i \big(\beta + \vec{r}_i^T \data^T \diag(S_i) \dual + \vec{r}_i^T \data^T \hat{\diag}(S_i)(\bm{\alpha}_i + \bm{\gamma}_i )-\vec{r}_i^T\data^T \bm{\gamma}_i  \big)\\
 &+ \sum_{i=1}^P \lambda_i^\prime \big(\beta - {\vec{r}_i^\prime}^T \data^T \diag(S_i) \dual +\vec{r}_i^{\prime^T} \data^T \hat{\diag}(S_i) (\bm{\alpha}_i^\prime + \bm{\gamma}_i^\prime )-{\vec{r}_i^\prime}^T\data^T \bm{\gamma}_i^\prime  \big)\,.
\end{align*}
%
We note that the objective is concave in $\dual,\bm{\alpha}_i,\bm{\gamma}_i$ and convex in $\vec{r}_i, \vec{r}_i^\prime$, $\forall i \in [P]$. Moreover, the constraint sets $\|\vec{r}_i\|_2\le 1,\, \|\vec{r}_i^\prime\|_2\le 1, \,\forall i$ are convex and compact. Invoking Sion's minimax theorem \cite{sion_minimax} for the inner $\max \min$ problem, we may express the strong dual of the problem \eqref{eq:dual} as
\begin{align*}
\min_{\substack{\bm{\lambda},\bm{\lambda}^\prime \in \real^P\\ \bm{\lambda},\bm{\lambda}^\prime\ge 0}} \,\min_{\substack{\vec{r}_i \in \real^d,\, \|\vec{r}_i\|_2\le 1\\ \vec{r}_i^\prime \in \real^d,\, \|\vec{r}_i^\prime\|_2\le 1} } \max_{\substack{\dual\in \real^n \\  \bm{\alpha}_i,\bm{\gamma}_i \in {\real^n}\\\bm{\alpha}_i, \bm{\gamma}_i\ge 0, \, \forall i  \\ \bm{\alpha}_i^\prime,\bm{\gamma}_i^\prime \in {\real^n}\\\bm{\alpha}_i^\prime, \bm{\gamma}_i^\prime\ge 0, \, \forall i }}\,  -\mathcal{L}^*(\dual) &+ \sum_{i=1}^P\lambda_i \big(\beta + \vec{r}_i^T \data^T \diag(S_i) \dual + \vec{r}_i^T \data^T \hat{\diag}(S_i)(\bm{\alpha}_i + \bm{\gamma}_i )-\vec{r}_i^T\data^T \bm{\gamma}_i  \big)\\
  &+ \sum_{i=1}^P \lambda_i^\prime \big(\beta  - {\vec{r}_i^\prime}^T \data^T \diag(S_i) \dual +\vec{r}_i^{\prime^T} \data^T \hat{\diag}(S_i) (\bm{\alpha}_i^\prime + \bm{\gamma}_i^\prime )-{\vec{r}_i^\prime}^T\data^T \bm{\gamma}_i^\prime   \big)\,.
\end{align*}

Computing the maximum with respect to $\dual,\bm{\alpha}_i,\bm{\gamma}_i,\bm{\alpha}_i^\prime,\bm{\gamma}_i^\prime$, $\forall i \in [P]$, analytically we obtain the strong dual to \eqref{eq:dual} as
\begin{align*}
&\min_{\substack{\bm{\lambda},\bm{\lambda}^\prime \in \real^P\\ \bm{\lambda},\bm{\lambda}^\prime\ge 0}} \,\min_{\substack{\vec{r}_i \in \real^d,\, \|\vec{r}_i\|_2\le 1\\\vec{r}_i^\prime \in \real^d,\, \|\vec{r}_i^\prime\|_2\le 1 \\  (2\hat{\diag}(S_i)-\vec{I}_n)\data\vec{r}_i\ge 0\\  (2\hat{\diag}(S_i)-\vec{I}_n)\data\vec{r}_i^\prime\ge 0}}
  \mathcal{L}\left(\sum_{i=1}^P \lambda_i \diag(S_i) \data \vec{r}_i^\prime
 -\lambda_i^\prime \diag(S_i) \data \vec{r}_i ,\labelvec\right)
 + \beta \sum_{i=1}^P (\lambda_i+\lambda_i^\prime).
\end{align*}
Now we apply a change of variables and define $\weight_i = \lambda_i \vec{r}_i$ and $\weight_i^{\prime}=\lambda_i^\prime \vec{r}_i^\prime$, $\forall i \in [P]$. Note that we can take $\vec{r}_i=0$ when $\lambda_i=0$ without changing the optimal value. We obtain
\begin{align*}
&\hspace{-0.5cm}\min_{\substack{\weight_i,\weight_i^{\prime} \in P_{S_i}\\ \|\weight_i\|_2\le \lambda_i\\
 \|\weight_i^{\prime}\|_2\le \lambda_i^\prime \\ \bm{\lambda},\bm{\lambda}^\prime\ge 0}}
  \mathcal{L}\left(\sum_{i=1}^P \diag(S_i) \data (\weight_i^{\prime^*}-\weight_i),\labelvec\right) 
+ \beta \sum_{i=1}^P (\lambda_i+\lambda_i^\prime).
\end{align*}
The variables $\lambda_i$, $\lambda_i^\prime$, $\forall i \in [P]$ can be eliminated since $\lambda_i=\|\weight_i\|_2$ and $\lambda_i^\prime=\|\weight_i^{\prime^*}\|_2$ are feasible and optimal. Plugging in these expressions, we get
\begin{align*}
&\min_{\substack{\weight_i,\weight_i^{\prime}\in P_{S_i}}}
  \mathcal{L}\left(\sum_{i=1}^P \diag(S_i) \data (\weight_i^{\prime}-\weight_i),\labelvec\right)  + \beta \sum_{i=1}^P (\|\weight_i\|_2+\|\weight^{\prime}_i\|_2)\,,
\end{align*}
which is identical to \eqref{eq:twolayerconvexprogram}, and proves that the objective values are equal. 
Constructing $\{{\firstw_j}^*,{\secondw_j}^*\}_{j=1}^{m^*}$ as stated in the theorem, and plugging in \eqref{eq:twolayer_objective_generic}, we obtain the value
%
\begin{align*}
    p^*\le \mathcal{L}\left(\sum_{j=1}^{m^*} \act(\data {\firstw_j}^*){\secondw_j}^*,\labelvec\right) \frac{1}{2} 
    &+ \frac{\beta}{2} \sum_{i=1, \weight^{\prime^*}_i\neq 0}^{P} \left( \Big\|\frac{\weight^{\prime^*}_i}{\sqrt{\|\weight^{\prime^*}_i\|_2}}\Big\|_2^2 + \Big\|\sqrt{\|\weight^{\prime^*}_i\|_2} \Big\|_2^2 \right) \nonumber \\ 
    &+ \frac{\beta}{2} \sum_{i=1, \weight^{*}_i\neq 0}^{P} \left(\Big\|\frac{\weight^{*}_i}{\sqrt{\|\weight^{*}_i\|_2}}\Big\|_2^2  + \Big\|\sqrt{\|\weight^{*}_i\|_2} \Big\|_2^2\right)
,\end{align*}
which is identical to the objective value of the convex program \eqref{eq:twolayerconvexprogram}. Since the value of the convex program is equal to the value of it's dual $d^*$ in \eqref{eq:dual}, and $p^*\ge d^*$, we conclude that $p^*=d^*$, which is equal to the value of the convex program \eqref{eq:twolayerconvexprogram} achieved by the prescribed parameters.
\qed

\section{Proof of Theorem \ref{theo:subsampled_GD}}
\label{sec:subsampled_GD}
Here, we first prove that Clarke stationary points of the nonconvex training problem of two-layer networks found by first order methods such as SGD/GD correspond to the global optimum of a version of our convex program based trichotomy arrangements. We then generalize this result to our convex program with standard dichotomy arrangements in \eqref{eq:twolayerconvexprogram}.\\

\noindent\textbf{Clarke Stationary Points and Convex Program with Trichotomies}\\
We first note that \cite{wang2022hidden} provided a similar result however their analysis is valid only for ReLU activations and convex programs with trichotomy arrangements. On the other hand, our proof extends to arbtirary piecewise linear activations and in the next section we generalize this result to our convex program with standard dichotomy arrangements (or diagonal matrices $\diag$).

For piecewise linear activations, we define the hyperplane arrangements matrices $\diag$ based on dichotomies as
\begin{align}\label{eq:diagonal_matrix_dichotomy}
\diag_{ii} := \begin{cases} 1 & \mbox{ if } \datavec_i^T\firstw \geq  0\\
\kappa & \mbox{ otherwise}  \end{cases}, 
\end{align}
whereas trichotomy arrangement matrix $\vec{T}$ is defined as 
\begin{align} \label{eq:diagonal_matrix_trichotomy}
\vec{T}_{ii} := \begin{cases} 1 & \mbox{ if } \datavec_i^T\firstw >0\\
0 & \mbox{ if } \datavec_i^T\firstw =  0\\
-1 & \mbox{ otherwise}  \end{cases}.
\end{align}

Due to the nondifferentiability of the piecewise linear activations, we next review the definition of the Clarke subdifferential \cite{clarke1975generalized} of a given function $f$. Let $ D \subset \mathbb{R}^d$ be the set of points at which \( f \) is differentiable. We assume that \( D \) has (Lebesgue) measure \( 1 \), meaning that \( f \) is differentiable \emph{almost everywhere}.
The Clarke subdifferential of $f$ at $\vec{x}$ is then defined as
\begin{align*} 
  \partial_C f(\vec{x}) = \mathrm{Conv} \left\{\lim_{k \to \infty} \nabla f(\vec{x}_k) \mid \lim_{k \to \infty }\vec{x}_k \to \vec{x},\,\vec{x}_k \in D\right\}.
\end{align*}
Then, we say that $\vec{x} \in \mathbb{R}^d$ is Clarke
stationary with respect to $f$ if $\vec{0} \in   \partial_C f(\vec{x})$.

Based on the definition above, we now consider a nonconvex neural networks model with piecewise linear activations, i.e., $f_{\theta}(\data)=\sum_{j=1}^m\relu{\data\firstw_j}\secondw_j$, and aim to optimize the parameters through the weight decay regularized objective function in \eqref{eq:twolayer_objective_generic}. From the definition of Clarke stationary point, for $j \in [m]$ with $\firstw_j \neq \vec{0}$, we have
\begin{align}\label{eq:clarke_stationary}
\begin{split}
     &-\beta \firstw_j \in \partial_{\firstw_j}\mathcal{L} \left( \sum_{j=1}^m\actt{\data\firstw_j}\secondw_j, \labelvec\right)\\
    &-\beta \secondw_j = \vec{g}^T\actt{\data\firstw_j}
\end{split}, 
\end{align}
where
\begin{align}
    \vec{g}\defn \nabla_{f}\mathcal{L} \bigg( \underbrace{\sum_{j=1}^m\actt{\data\firstw_j}\secondw_j}_{= f}, \labelvec\bigg). 
\end{align}
We note that \eqref{eq:clarke_stationary} formulates the stationarity conditions of the nonconvex training problem and \cite{kornowski2023tempered} proved that running GD to minimize this objective converges to a point, where these stationarity conditions are satified. Then, the first stationary condition in \eqref{eq:clarke_stationary} implies that there exists $\bm{\delta}_j \in [-\kappa,1]^n$ such that
\begin{align*}
   -\beta \firstw_j = \secondw_j\left(\data^T\diag_j\vec{g}+\data^T \vec{S}_j \mathrm{diag}(\bm{\delta}_j)\vec{g} \right),
\end{align*}
where $\diag_j$ is defined in \eqref{eq:diagonal_matrix_dichotomy} and $\vec{S}_j = \mathrm{diag}(\mathbbm{1}[\data \firstw_j = 0])$. Assuming $\firstw_j \neq \vec{0}$ and $\secondw_j \neq 0 $, the equality above implies that 
\begin{align}\label{eq:stationary_cond1}
    -\beta \frac{\firstw_j}{\secondw_j} =\data^T\diag_j\vec{g}+\data^T \vec{S}_j \mathrm{diag}(\bm{\delta}_j)\vec{g} .
\end{align}
Additionally, from the second stationary condition in \eqref{eq:clarke_stationary}, we have
\begin{align}\label{eq:stationary_cond2}
    -\beta \secondw_j &= \vec{g}^T\diag_j \data \firstw_j \nonumber\\
    &= {\firstw_j}^T \data^T\diag_j \vec{g} \nonumber\\
    &={\firstw_j}^T \left( \data^T\diag_j \vec{g} +\data^T\vec{S}_j \mathrm{diag}(\bm{\delta}_j) \vec{g}\right)\nonumber\\
    &={\firstw_j}^T\left( -\beta \frac{\firstw_j}{\secondw_j}\right)\nonumber\\
    &=-\beta \frac{\|\firstw_j\|_2^2}{\secondw_j}.
\end{align}
Thus, we have $\vert\secondw_j\vert=\|\firstw_j\|_2$ and from \eqref{eq:stationary_cond1}
\begin{align}
    \|\data^T\diag_j\vec{g}+\data^T \vec{S}_j \mathrm{diag}(\bm{\delta}_j)\vec{g}\|_2 = \beta.
\end{align}
Now, given the following subsampled convex program with trichotomy arrangement
\begin{align}
      \min_{ \weight  \in \mathcal{C}(\data)}  \mathcal{L}\left(\sum_{i=1}^{\tilde{P}}\actt{\vec{T}_i}\data(\weight_i-\weight_{i+\tilde{P}}) ,\labelvec \right)  +\beta \sum_{i=1}^{2\tilde{P}}\|\weight_i\|_{2}  ,
      \label{eq:convex_program_trichotomy}
\end{align}
where $\mathcal{C}(\data)$ are convex constraint enforcing weights to satisfy the trichotomy arrangement patterns in \eqref{eq:diagonal_matrix_trichotomy}, the KKT conditions are given by: for $i \in [\tilde{P}]$, there exists $\bm{\zeta}_i \geq 0$ and $\bm{\xi}_i$
\begin{equation}
    \begin{aligned}\label{eq:KKT_tri}
      &\data^T \left(\actt{\vec{T}_i}\dual+\vec{T}_i\bm{\zeta}_i + \tilde{\vec{S}}_i \bm{\xi}_i \right) + \beta \frac{\weight_i}{\|\weight_i\|_2}=0,  &&\text{ if } \weight_i\ \neq \vec{0}\\
     &\left \|\data^T \left(\actt{\vec{T}_i}\dual+\vec{T}_i\bm{\zeta}_i + \tilde{\vec{S}}_i \bm{\xi}_i \right) \right\|_2 \leq \beta,  &&\text{ if } \weight_i\ = \vec{0}\\
    &\data^T \left(-\actt{\vec{T}_i}\dual+\vec{T}_i\bm{\zeta}_{i+\tilde{P}} + \tilde{\vec{S}}_i \bm{\xi}_{i+\tilde{P}} \right) + \beta \frac{\weight_{i+\tilde{P}}}{\|\weight_{i+\tilde{P}}\|_2}=0,  &&\text{ if } \weight_{i+\tilde{P}}\ \neq \vec{0}\\
     &\left \|\data^T \left(-\actt{\vec{T}_i}\dual+\vec{T}_i\bm{\zeta}_{i+\tilde{P}} + \tilde{\vec{S}}_i \bm{\xi}_{i+\tilde{P}} \right) \right\|_2 \leq \beta,  &&\text{ if } \weight_{i+\tilde{P}}\ = \vec{0}  
\end{aligned}
\end{equation}
where $\tilde{\vec{S}}_i$ is a diagonal matrix satisfying that $\tilde{\vec{S}}_{jj}=1$ if $\vec{T}_{i,jj}=0$ and $\tilde{\vec{S}}_{jj}=0$ otherwise. Also, $\dual \in \mathbb{R}^n$ is defined as $\dual = \nabla \mathcal{L}\left(\sum_{i=1}^{\tilde{P}}\actt{\vec{T}_i}\data(\weight_i-\weight_{i+\tilde{P}}) ,\labelvec \right)$.

Due to the one to one mapping in Proposition \ref{prop:mapping}, we have $\vec{g}=\dual$. Also, taking $\bm{\zeta}_i=0$, $\bm{\xi}=\mathrm{diag}(\bm{\delta}_j)\vec{g}$, $\bm{\zeta}_{i+\tilde{P}}=0$, $\bm{\xi}_{i+\tilde{P}}=-\mathrm{diag}(\bm{\delta}_j)\vec{g}$, $\diag_j=\actt{\vec{T}_i}$, and $\vec{S}_j=\tilde{\vec{S}}_i$ satisfies the KKT condition in \eqref{eq:KKT_tri}.  Therefore, the Clarke stationary points of the nonconvex training objective in \eqref{eq:clarke_stationary} is a global optimum of the subsampled convex program in \eqref{eq:convex_program_trichotomy}.
\qed\\

\noindent\textbf{Extension to Our Convex Program with Dichotomies}\\
In order to establish a similar proof for our hyperplane arrangement matrices based on dichotomies, here, we show that the optimal solutions to the subsampled convex programs based on dichotomies and trichomoties coincide, i.e., the maximizers of the dual constraints for each case are the same.

We start with stating the dual constraint (dc) of \eqref{eq:twolayer_dual_generic} for each case with $\tilde{P}$ sampled arrangements as follows
\begin{align}\label{eq:dc_trichotomy}
    \begin{split}
        &dc \defn \max_{k \in \tilde{P} } \max_{\firstw \in \ball_2 \cap \mathcal{C}_k}\left\vert \dual^T \diag_k \data \firstw \right\vert \\
        &dc^t \defn  \max_{k \in \tilde{P} } \max_{\firstw \in \ball_2  \cap \mathcal{C}_k^t}\left\vert \dual^T \actt{\vec{T}_k} \data \firstw \right\vert
    \end{split} \; ,  
\end{align}
where
\begin{align}\label{eq:arrangement_cons}
\begin{split}
&\mathcal{C}_k \defn \{\weight \in \mathbb{R}^d:  \datavec_i^T \weight \geq 0, \forall i \in \{i:\diag_{k,ii}=1\}, \datavec_i^T \weight<0,
\text{otherwise}   \}\\
    &\mathcal{C}_k^t \defn \{\weight \in \mathbb{R}^d: \vec{T}_{k,ii} \datavec_i^T \weight > 0, \forall i \in \{i:\vec{T}_{k,ii} \in \{\pm1\} \}, \datavec_i^T \weight=0, \text{otherwise}   \}
\end{split}.
\end{align}
We remark that $\mathcal{C}_k$ is a relaxation of $\mathcal{C}_k^t$ since $\mathcal{C}_k^t$ enforces certain entries to be exactly zero due to trichotomies. 

We first note that if there are no zero entries in the optimal $\vec{T}_k$ for $dc^t$ then the same solution will be optimal for $dc$ since both problems will be exactly identical in that case. If the optimal $\vec{T}_k$ has a zero entry then we need to check if it matches to the solution of $dc$. To do so, we need to show that
\begin{align} \label{eq:dichotomy_v2}
    \argmax_{\firstw \in \ball_2  \cap \mathcal{C}_k^t}\left\vert \dual^T \actt{\vec{T}_k} \data \firstw \right\vert \in \left\{   \argmax_{\firstw \in \ball_2  \cap \mathcal{C}_i}\left\vert \dual^T \diag_l \data \firstw \right\vert,  \argmax_{\firstw \in \ball_2  \cap \mathcal{C}_j}\left\vert \dual^T \diag_j \data \firstw \right\vert \right\},
\end{align}
where $\diag_l$ and $\diag_j$ are dichotomies that include the zero index in $\vec{T}_k$, say $\datavec_i^T\firstw=0$, in the nonnegative ($\datavec_i^T\firstw\geq 0$) and nonpositive ($\datavec_i^T\firstw\leq 0$) sides of the hyperplane. Other than that, all the entries of $\diag_l, \diag_j$, and $\vec{T}_k$ are the same, i.e., $\forall i \in [n]$,
\begin{align*}
\diag_{l,ii} := \begin{cases} \actt{\vec{T}_{k,ii} }& \mbox{ if } \datavec_i^T\firstw \neq  0\\
1 & \mbox{ otherwise}  \end{cases}, \quad
\diag_{j,ii} := \begin{cases} \actt{\vec{T}_{k,ii} }& \mbox{ if } \datavec_i^T\firstw \neq  0\\
0 & \mbox{ otherwise}  \end{cases}. 
\end{align*}
If one of the dichotomy problems in \eqref{eq:dichotomy_v2} achieves the same optimum when $\datavec_i^T\firstw=0$, then we can claim that there is an optimal dichotomy arrangement corresponding to the optimal trichotomy arrangement $\vec{T}_k$. If not, then this means that both dichotomy problems in \eqref{eq:dichotomy_v2} achieve the optimum when $\datavec_i^T\firstw> 0$ and $\datavec_i^T\firstw< 0$, respectively. However, this cannot be true. To illustrate this, let us first denote the solutions to each problem as $\firstw_1$ and $\firstw_2$ such that $\datavec_i^T\firstw_1> 0$ and $\datavec_i^T\firstw_2< 0$. Since the objective function is linear, we can find a linear interpolation between $\firstw_1$ and $\firstw_2$ as $\firstw_0\defn t \firstw_1 + (1-t) \firstw_2$, where $t \in [0,1]$, such that $\datavec_i^T\firstw_0= 0$. Then, since $h(\firstw)\defn \datavec_i^T\firstw$ is a linear function, the interpolation between them cannot achieve a value that is strictly less than both, i.e., $h(\firstw_0)\geq \min\{h(\firstw_1),h(\firstw_2)\}$. Therefore, we have a contradiction due to the assumption that both dichotomies achieves optimal solution without zero entries. This concludes the proof.
\qed


\section{Proof of Corollary \ref{cor:mainconvex_bias}}
To derive the convex program for a network with bias term, we first define a new variable by concatenating the bias and weights as $\hat{\vec{w}}_j^{(1)}:=[\firstw_j;\bias_j]$. Then the rest of the derivations directly follows from the proof of Theorem \ref{theo:mainconvex} when we replace $\firstw$ with $\hat{\vec{w}}^{(1)}=[\firstw;\bias]$.
\qed

\section{Proof of Theorem \ref{theo:lowrank_approx}}
\label{sec:proofcor_lowrank_approx}

\begin{lem}\label{lemma:lowrank_approx}
Given an $L$-Lipschitz convex loss $\mathcal{L}(\cdot,\labelvec)$ and an $R$-Lipschitz activation function $\phi(\cdot)$, consider the following nonconvex optimization problem with $\hat{\data}_k$
\begin{align*}
\begin{split}
    &(\firstwmath,\secondwvech) \in \argmin_{\theta \in \Theta_s} \mathcal{L}(\actt{\hat{\data}_k \firstwmat } \secondwvec,\labelvec)+\beta \|\secondwvec\|_1  
\end{split}
\end{align*}
and the objective value with the original data $\data$ evaluated at any optimum $(\firstwmath,\secondwvech)$
\begin{align*}
       p_k:=\mathcal{L}(\actt{\data \firstwmath } \secondwvech,\labelvec)+\beta\|\secondwvech\|_1.
\end{align*}
Then, we have the following approximation guarantee
\begin{align*}
    p^* \leq  p_k \leq p^*  \left(1+\frac{L R\sigma_{k+1}}{\beta}\right)^2.
\end{align*}
\end{lem}
\noindent\textbf{Proof of Lemma \ref{lemma:lowrank_approx}}
\label{sec:proofthm_lowrank_approx}
We start with defining the optimal parameters for the original and rank-$k$ approximation of the rescaled problem in \eqref{eq:twolayer_objective_generic_l1} as
\begin{align}\label{eq:low_rank_defs}
\begin{split}
       &({\firstwmat}^*,{\secondwvec}^*) \in \argmin_{\theta \in \Theta_s} \mathcal{L}(\actt{\data \firstwmat } \secondwvec,\labelvec)+\beta \|\secondwvec\|_1 \\
    &(\firstwmath,\secondwvech) \in \argmin_{\theta \in \Theta_s} \mathcal{L}(\actt{\hat{\data}_k \firstwmat } \secondwvec,\labelvec)+\beta \|\secondwvec\|_1  
\end{split}
\end{align}
and the objective value achieved by the parameters trained using $\hat{\data}_k$ as
\begin{align*}
       p_k:=\mathcal{L}(\actt{\data \firstwmath } \secondwvech,\labelvec)+\beta\|\secondwvech\|_1.
\end{align*}
Then, we have
\begin{align}
    p^*&=\mathcal{L}(\actt{\data{\firstwmat}^* } {\secondwvec}^*,\labelvec)+\beta \|{\secondwvec}^*\|_1\nonumber\\ &\stackrel{(i)}{\leq}  \mathcal{L}(\actt{\data \firstwmath } \secondwvech,\labelvec)+\beta \|\secondwvech\|_1 =p_k\nonumber\\
    &\stackrel{(ii)}{\leq} \mathcal{L}(\actt{\hat{\data}_k \firstwmath } \secondwvech,\labelvec)+(\beta+LR\sigma_{k+1}) \|\secondwvech\|_1 \nonumber\\
     &\leq  \left( \mathcal{L}(\actt{\hat{\data}_k \firstwmath } \secondwvech,\labelvec)+\beta \|\secondwvech\|_1 \right) \left(1+\frac{L R\sigma_{k+1}}{\beta} \right) \nonumber\\
     &\stackrel{(iii)}{\leq}  \left( \mathcal{L}(\actt{\hat{\data}_k {\firstwmat}^* } {\secondwvec}^*,\labelvec)+\beta \|{\secondwvec}^*\|_1 \right) \left(1+\frac{L R\sigma_{k+1}}{\beta} \right) \nonumber\\
     &\stackrel{(iv)}{\leq}  \left( \mathcal{L}(\actt{\data {\firstwmat}^* } {\secondwvec}^*,\labelvec)+\beta \|{\secondwvec}^*\|_1 \right) \left(1+\frac{L R\sigma_{k+1}}{\beta} \right)^2 \nonumber\\
      &=  p^* \left(1+\frac{L R\sigma_{k+1}}{\beta} \right)^2, \label{eq:rankk_proof}
\end{align}
where $(i)$ and $(iii)$ follow from the optimality definitions of the original and approximated problems in \eqref{eq:low_rank_defs}. In addition, $(ii)$ and $(iv)$ follow from the relations below
{\small
\begin{align*}
    \mathcal{L}(\actt{\data \firstwmath } \secondwvech,\labelvec) &=    \mathcal{L}(\actt{\data \firstwmath } \secondwvech-\actt{\hat{\data}_k \firstwmath } \secondwvech+\actt{\hat{\data}_k \firstwmath } \secondwvech,\labelvec)\\
    &\stackrel{(1)}{\leq}  \mathcal{L}(\actt{\data \firstwmath } \secondwvech-\actt{\hat{\data}_k \firstwmath } \secondwvech,\labelvec)+ \mathcal{L}(\actt{\hat{\data}_k \firstwmath } \secondwvech,\labelvec)\\
    &\stackrel{(2)}{\leq}  L\left\|\actt{\data \firstwmath } \secondwvech-\actt{\hat{\data}_k \firstwmath } \secondwvech \right\|_2+ \mathcal{L}(\actt{\hat{\data}_k \firstwmath } \secondwvech,\labelvec)\\
    &= L \left\|\sum_{j=1}^m\left(\actt{\data \firstwh_j } -\actt{\hat{\data}_k \firstwh_j}\right) \secondwh_j \right\|_2+ \mathcal{L}(\actt{\hat{\data}_k \firstwmath } \secondwvech,\labelvec)\\
    &\stackrel{(3)}{\leq}  L \sum_{j=1}^m\left\|\actt{\data \firstwh_j } -\actt{\hat{\data}_k \firstwh_j} \right\|_2\left\vert\secondwh_j \right \vert+ \mathcal{L}(\actt{\hat{\data}_k \firstwmath } \secondwvech,\labelvec)\\
    &\leq  L \max_{j \in [m]}\left\|\actt{\data \firstwh_j } -\actt{\hat{\data}_k \firstwh_j} \right\|_2 \|\secondwvech \|_1+ \mathcal{L}(\actt{\hat{\data}_k \firstwmath } \secondwvech,\labelvec)\\
    &\stackrel{(4)}{\leq}  LR \max_{j \in [m]}\|\firstwh_j\|_2\left\|\data -\hat{\data}_k  \right\|_2 \|\secondwvech \|_1+ \mathcal{L}(\actt{\hat{\data}_k \firstwmath } \secondwvech,\labelvec)\\
    &\stackrel{(5)}{=}  LR \sigma_{k+1}\|\secondwvech \|_1+ \mathcal{L}(\actt{\hat{\data}_k \firstwmath } \secondwvech,\labelvec),
\end{align*}}
where we use the convexity and $L$-Lipschitz property of the loss function, convexity of $\ell_2$-norm, $R$-Lipschitz property of the activation, and $\max_{j}\|\firstwh_j\|_2=1$ from the rescaling in Lemma \ref{lemma:scaling} for $(1), (2), (3), (4),$ and (5), respectively.

Based on \eqref{eq:rankk_proof}, we have
\begin{align*}
    p^* \leq  p_k \leq p^*  \left(1+\frac{L R\sigma_{k+1}}{\beta}\right)^2.
\end{align*}

\qed

Based on the approximation bound provided by Lemma \ref{lemma:lowrank_approx}, we next show that the complexity of solving the convex reformulations can be reduced via rank-$k$ approximations. Note that due to the rank-$k$ data matrix $\hat{\data}_k$, the number of hyperplane arrangements in the corresponding convex formulation \eqref{eq:twolayerconvexprogram} is significantly reduced. We formalize this in the next corollary. 

We first restate the exact convex program as follows
\begin{align*}
&p^*=\min_{ \weight  \in \mathcal{C}(\data)}\,  \mathcal{L}\left(\dataf\weight ,\labelvec \right)  +\beta \sum_{i=1}^{2P}\|\weight_i\|_{2}\\
&=\min_{ \weight  \in \mathcal{C}(\data)}\,  \mathcal{L}\left(\sum_{i=1}^P\diag_i\data(\weight_i-\weight_{i+P}) ,\labelvec \right)  +\beta \sum_{i=1}^{2P}\|\weight_i\|_{2}.
\end{align*}
In addition to this, we define two rank-$k$ approximated versions based on Theorem \ref{theo:lowrank_approx}
\begin{align}
   \hat{\weight}^{(k)}& \in \argmin_{ \weight  \in \mathcal{C}(\hat{\data}_k)}\,  \mathcal{L}\left(\sum_{i=1}^{\hat{P}}\diag_i^k \hat{\data}_k (\weight_i-\weight_{i+\hat{P}}) ,\labelvec \right)  +\beta \sum_{i=1}^{2\hat{P}}\|\weight_i\|_{2} \label{eq:cvx_rank1} \\
      \weight^{(k)}& \in \argmin_{ \weight  \in \mathcal{C}(\hat{\data}_k)}\,  \mathcal{L}\left(\sum_{i=1}^{\hat{P}}\diag_i^k\data(\weight_i-\weight_{i+\hat{P}}) ,\labelvec \right)  +\beta \sum_{i=1}^{2\hat{P}}\|\weight_i\|_{2}  ,\label{eq:cvx_rank2}
\end{align}
where $\diag_i^k$ denotes the set of arrangements sampled from rank-$k$ data matrix $\hat{\data}_k $. Note that the difference between \eqref{eq:cvx_rank1} and \eqref{eq:cvx_rank2} is that we use rank-$k$ data for sampling arrangements of both problems while using the full rank data only for \eqref{eq:cvx_rank2}.

Let us first denote the objective values evaluated at $\hat{\weight}^{(k)}$ and $\weight^{(k)}$ using the original data $\data$ as $\hat{p}_{\mathrm{cvx}-k}$ and $p_{\mathrm{cvx}-k}$, respectively. Then from Lemma \ref{lemma:lowrank_approx}, we can use $\hat{\weight}^{(k)}$ to achieve the following approximation guarantee
\begin{align}\label{eq:theorem_final}
    p^* \leq \hat{p}_{\mathrm{cvx}-k}\leq p^*  \left(1+\frac{L R\sigma_{k+1}}{\beta}\right)^2.
\end{align}
Moreover, since \eqref{eq:cvx_rank2} utilizes the full rank data, the network output can span a larger output space and therefore the corresponding optimal objective value of the minimization problem is smaller, i.e., $p_{\mathrm{cvx}-k} \leq \hat{p}_{\mathrm{cvx}-k}$. However, $p_{\mathrm{cvx}-k} \geq p^*$ since $p_{\mathrm{cvx}-k} $ has smaller number of arrangements due to using rank-$k$ matrix for hyperplane arrangement sampling. 

Combining these observations with \eqref{eq:theorem_final} yields
\begin{align*}
    p^*\leq p_{\mathrm{cvx}-k} \leq \hat{p}_{\mathrm{cvx}-k} \leq p^*  \left(1+\frac{L R\sigma_{k+1}}{\beta}\right)^2.
\end{align*}
\qed
\section{Proof of Theorem \ref{theo:effcient_sampling}}
Suppose that we randomly sample binary vectors $\vec{d}\in \{0,1\}^n$, which denotes the diagonal entries of $\diag$. Then, we define the probability of $\vec{d}$ being the $i^{th}$ arrangement as $\theta_i$, i.e., $p_i:=\mathbb{P}[\mathrm{diag}(\vec{d})=\diag_i]$. Next, we compute an event where we miss at least one arrangement among $P$ possible ones. Let this event be denoted as $\mathcal{A}$, which is defined as follows
\begin{align*}
    \mathbb{P}[\mathcal{A}]=\mathbb{P}\left[\bigcup_{i=1}^P\{\text{miss $\diag_i$}\} \right] \leq \sum_{i=1}^P\mathbb{P}[\mathrm{diag}(\vec{d}) \neq  \diag_i] = \sum_{i=1}^P (1-\theta_i)^{\tilde P} \leq P (1-\theta_{min})^{\tilde P},
\end{align*}
where the first inequality follows from the union bound, $\tilde P$ denotes the number of arrangements we sample, and $\theta_{min}:= \min_i \theta_i$. Then, to be able to sample all arrangements with probability $1-\epsilon$, we choose $\tilde P $ such that 
\begin{align*}
P (1-\theta_{min})^{\tilde P} \leq \epsilon \implies \tilde P \geq \frac{\log\left( P / \epsilon\right)}{\log(1/(1-\theta_{min}))} .
\end{align*}
Next, we use the following identity $\log(1/(1-x)) \geq  x$ given $x<1$, to obtain an upper bound for the RHS of the inequality above
\begin{align*}
 \frac{\log\left( P / \epsilon\right)}{\log(1/(1-\theta_{min}))}\geq \frac{\log\left( P / \epsilon\right)}{\theta_{min}}= \frac{P\log\left( P / \epsilon\right)}{\bar \theta},
\end{align*}
where $\bar \theta := P \theta_{min}$. Therefore, the threshold for the number of hyperplane arrangements we need to sample simplifies to
\begin{align*}
    \tilde P \geq \frac{P\log\left( P / \epsilon\right)}{\bar \theta} = \mathcal{O}\left(k \left(\frac{n}{k}\right)^k \log\left(\frac{n}{k}\right)\right).
\end{align*}
Note that this threshold is a polynomial function of all problem parameters, i.e., the number of samples $n$ and feature dimension $d$, since $P=\mathcal{O}((n/k)^k)$ given a fixed rank $k$ based on Remark \ref{rem:rank_approx} and $\bar\theta$ is a constant factor. 
\qed

\section{Proof of Corollary \ref{cor:main_lp}}
We first replace $\|\firstw\|_2\le 1$ with $\|\firstw\|_p\le 1$. Then, the rest of the derivations directly follows from the proof of Theorem \ref{theo:mainconvex} and yield the claimed group $\ell_p$ regularized convex program in \eqref{eq:twolayerconvexprogram_lp}.
\qed

\section{Proof of Theorem \ref{theo:main_spikefree}}
\label{sec:proofthm_spikefree}
Following Theorem \ref{theo:mainconvex}, we have the following dual constraint 
\begin{align*} 
    \max_{\firstw\in \ball_2} \, \left \vert \dual^T \act(\data \firstw) \right \vert 
    &=\max_{\substack{S \subseteq [n]\\ S \in \mathcal{S} }}\max_{\substack{\firstw\in \ball_2 \cap P_S}} \, \left\vert \dual^T  \diag(S)\data \firstw\right \vert \\
    &=\max_{\substack{S \subseteq [n]\\ S \in \mathcal{S} }}\max_{\substack{\firstw\in \ball_2\\\data\firstw\geq 0}} \, \left\vert \dual^T  \data \firstw\right \vert,
\end{align*}
where the second equality follows from the definition of spike-free matrices. We then apply the same steps in Theorem \ref{theo:mainconvex} for a case with $P=1$ and $\diag_1=\vec{I}_n$ to achieve the convex program claimed in \eqref{eq:twolayerconvexprogram_spikefree}.
\qed

\section{Proof of Theorem \ref{theo:mainconvex_vector}}
\label{sec:proofthm_vector}

As in Theorem \ref{theo:mainconvex}, we start with the dual of the scaled primal problem in \eqref{eq:twolayer_objective_generic_vector_scaled}, which is formulated as
\begin{align}\label{eq:twolayer_objective_generic_vector_dual}
  & d_v^*  =\min_{\dualmat \in \mathbb{R}^{n \times C}} - \mathcal{L}^*(\dualmat) \mbox{ s.t.}\, \max_{\firstw:\,\|\firstw\|_2\le 1} \, \left\| \dualmat^T \act(\data \firstw) \right\|_2 \le \beta\,. 
\end{align}
Now, let us focus on the dual constraint as follows
\begin{align*} 
    \max_{\firstw \in \ball_2} \, \left\| \dualmat^T \act(\data \firstw) \right\|_2 &= \max_{\firstw, \vec{g} \in \ball_2} \, \vec{g}^T \dualmat^T \act(\data \firstw)\\
    &=\max_{\substack{S \subseteq [n]\\ S \in \mathcal{S} }}\max_{\substack{\firstw, \vec{g} \in \ball_2\\\firstw \in P_S}} \, \vec{g}^T \dualmat^T \diag(S)\data \firstw\\
    &=\max_{\substack{S \subseteq [n]\\ S \in \mathcal{S} }}\max_{\substack{\firstw, \vec{g} \in \ball_2\\\firstw \in P_S}} \, \left\langle \dualmat, \diag(S)\data \firstw\vec{g}^T \right\rangle \\
    &=\max_{\substack{S \subseteq [n]\\ S \in \mathcal{S} }}\max_{\substack{ \vec{Z}=\firstw\vec{g}^T\\ \firstw \in P_S\\ \firstw, \vec{g} \in \ball_2}} \, \left\langle \dualmat, \diag(S)\data \vec{Z} \right\rangle \\
  &=\max_{\substack{S \subseteq [n]\\ S \in \mathcal{S} }}\max_{\substack{\vec{Z} \in \mathcal{K} }} \, \left\langle \dualmat, \diag(S)\data \vec{Z} \right\rangle,
\end{align*}
where $\mathcal{K}:=\mathrm{Conv}\{\vec{u}\vec{g}^T\,:\, \firstw \in P_S,\, \|\firstw\|_2, \|\vec{g}\|_2 \leq 1 \}$. We also define a new convex norm over the set $\mathcal{K}$ as 
\begin{align*}
    \|\vec{Z}\|_{\mathcal{C}}: =\min_{t\ge 0\,} t \mbox{ s.t. } \weightmat \in t\, \mathcal{K}.
\end{align*}
Then, the dual problem \eqref{eq:twolayer_objective_generic_vector_dual} can be equivalently written as
\begin{align*}
   d_v^*= \min_{\dualmat \in \mathbb{R}^{n \times C}} - \mathcal{L}^*(\dualmat) \mbox{ s.t.}\, \max_{\vec{Z}: \vec{Z} \in \mathcal{K}_i} \, \left\langle \dualmat, \diag_i\data \vec{Z} \right\rangle \le \beta\, \forall i \in [P]. 
\end{align*}
where $\mathcal{K}_i:=\mathrm{Conv}\{\firstw \vec{g}^T\,:\, \firstw \in P_{S_i},\, \|\firstw\|_2\leq 1,\, \|\vec{Z}\|_*\leq 1 \}$ with the corresponding norm $\|\cdot\|_{\mathcal{C}_i}$. We then write the Lagrangian of the above problem form as follows
\begin{align} \label{eq:dual_vector}
 d_v^*= \max_{\substack{\dualmat\in \real^{n \times C}  }}\,\min_{\substack{\bm{\lambda}\in \real^P\\ \bm{\lambda}\ge 0}} \min_{\substack{\vec{Z}_i \in \mathcal{K}_i, \forall i} } -\mathcal{L}^*(\dualmat)  &+ \sum_{i=1}^P \lambda_i \big(\beta - \left\langle \dualmat, \diag_i\data \vec{Z}_i \right\rangle \big)\,.
\end{align}
%
Invoking Sion's minimax theorem \cite{sion_minimax} for the $\max \min$ problems, we may express the strong dual of the problem \eqref{eq:twolayer_objective_generic_vector_dual} as
\begin{align*}
 d_v^*= \min_{\substack{\bm{\lambda}\in \real^P\\ \bm{\lambda}\ge 0}} \min_{\substack{\vec{Z}_i \in \mathcal{K}_i,\forall i } }\, \max_{\substack{\dualmat\in \real^{n \times C}  }} -\mathcal{L}^*(\dualmat)  &+ \sum_{i=1}^P \lambda_i \big(\beta - \left\langle \dualmat, \diag_i\data \vec{Z}_i \right\rangle \big)\,.
\end{align*}
Computing the maximum with respect to $\dualmat$, analytically we obtain the strong dual to \eqref{eq:twolayer_objective_generic_vector_dual} as
\begin{align*}
&d_v^*=\min_{\substack{\bm{\lambda} \in \real^P\\ \bm{\lambda}\ge 0}} \,  \min_{\substack{\vec{Z}_i \in \mathcal{K}_i,\forall i } }
  \mathcal{L}\left(\sum_{i=1}^P \lambda_i \diag(S_i) \data \vec{Z}_i
 ,\labelvec\right)
 + \beta \sum_{i=1}^P \lambda_i.
\end{align*}
Now we apply a change of variables and define $\weightmat_i = \lambda_i \vec{Z}_i$, $\forall i \in [P]$. Thus, we obtain
\begin{align*}
&d_v^*=\min_{\substack{\weightmat_i \in \lambda_i \mathcal{K}_i\\  \bm{\lambda} \ge 0}}
  \mathcal{L}\left(\sum_{i=1}^P \diag(S_i) \data \weightmat_i^{*},\labelvec\right) 
+ \beta \sum_{i=1}^P \lambda_i.
\end{align*}
The variables $\lambda_i$, $\forall i \in [P]$ can be eliminated since $\lambda_i=\|\weightmat_i\|_{\mathcal{C}_i}$ is feasible and optimal. Plugging in these expressions, we get
\begin{align}\label{eq:convex_program_vector_proof}
&d_v^* =\min_{\substack{\weightmat_i \in \mathbb{R}^{d \times C}}}
  \mathcal{L}\left(\sum_{i=1}^P \diag(S_i) \data \weightmat_i,\labelvec\right)  + \beta \sum_{i=1}^P \|\weightmat_i\|_{\mathcal{C}_i}\,,
\end{align}
which is identical to the objective value of the convex program \eqref{eq:twolayerconvexprogram_vector}. Since the value of the convex program is equal to the value of it's dual $d_v^*$ in \eqref{eq:dual_vector}, and $p_v^*\ge d_v^*$, we conclude that $p_v^*=d_v^*$, which is equal to the value of the convex program \eqref{eq:twolayerconvexprogram_vector} achieved by the prescribed parameters.

\qed

\section{Proof of Theorem \ref{theo:mainconvex_vector_l1}}
\label{sec:proofthm_vector_l1}

As in Theorem \ref{theo:mainconvex_vector}, we start with scaling the primal problem in \eqref{eq:twolayer_objective_generic_vector_l1} as
\begin{align}\label{eq:twolayer_objective_generic_vector_l1_scaled}
     p_{v1}^*:=\min_{\theta \in \Theta_s} \mathcal{L}(f_{\theta}(\data),\labelvec)+\beta \sum_{j=1}^m\|\secondwvec_j\|_1 \,.
\end{align}
which has the following dual with respect to $\secondwvec_j$
\begin{align}\label{eq:twolayer_objective_generic_vector_l1_dual}
  &p_{v1}^*= d_{v1}^*  =\min_{\dualmat \in \mathbb{R}^{n \times C}} - \mathcal{L}^*(\dualmat) \mbox{ s.t.}\, \max_{\firstw \in \ball_2} \, \left \vert \dual_l^T \act(\data \firstw) \right \vert \le \beta\, \forall l \in [C]. 
\end{align}
Now, let us rewrite the dual constraint as follows
\begin{align*} 
    \max_{\firstw\in \ball_2} \, \left \vert \dual_l^T \act(\data \firstw) \right \vert 
    &=\max_{\substack{S \subseteq [n]\\ S \in \mathcal{S} }}\max_{\substack{\firstw\in \ball_2 \cap  P_S}} \, \left\vert \dual_l^T  \diag(S)\data \firstw\right \vert .
\end{align*}
Then, the dual problem \eqref{eq:twolayer_objective_generic_vector_l1_dual} can be equivalently written as
\begin{align*}
   d_{v1}^*= \min_{\dualmat \in \mathbb{R}^{n \times C}} - \mathcal{L}^*(\dualmat) \mbox{ s.t.}\, \max_{ \vec{Z}_i \in \mathcal{K}_i} \, \left\vert \dual_l^T \diag_i\data \firstw \right\vert \le \beta\, \forall i \in [P],\, \forall l \in [C]. 
\end{align*}
The rest of the proofs directly follow from Theorem \ref{theo:mainconvex}, which yield the convex problem in \eqref{eq:twolayerconvexprogram_vector_l1}.
\qed

\section{Constructing hyperplane arrangements in polynomial time}\label{sec:hyperplane_arrangments_appendix}
{We first define the set of all hyperplane arrangements for the data matrix $\data$ as
\begin{align*}
\mathcal{H} := \bigcup \big \{ \{\sign(\data \firstw)\} \,:\, \firstw \in \real^d \big \}\,.
\end{align*}
By definition, $\mathcal{H}$ is bounded, i.e., $\exists N_H \in \mathbb{N} <\infty$ such that $\vert \mathcal{H} \vert \le N_H$. We now define the collection of sets that correspond to positive signs for each element in $\mathcal{H}$ as
\begin{align*}
\mathcal{S} := \big\{ \{ \cup_{h_i=1} \{i\}  \} \,:\, \vec{h} \in \mathcal{H} \big\},
\end{align*}
which is also an alternative representation of the sign patterns in $\mathcal{H}$. Using these definitions, we introduce a new diagonal matrix representation $\diag(S)\in \real^{n\times n}$ as
\begin{align*} 
\diag(S)_{ii} := \begin{cases} 1 & \mbox{ if } i \in S\\
\kappa & \mbox{ otherwise}  \end{cases}.
\end{align*}
Therefore, the output of the activation function can be equivalently written as $\act(\data \firstw)=\diag(S)\data \firstw $ provided that $(2\diag(S)-\vec{I}_n)\data \firstw \geq 0$.

}
We now consider the number of all distinct sign patterns $\mbox{sign}(\data \vec{z})$ for all possible choices $z\in \real^{d}$. Note that this number is the number of regions in a partition of $\real^d$ by hyperplanes passing through the origin, and are perpendicular to the rows of $\data$. We now show that the dimension $d$ can be replaced with $\mbox{rank}(\data)$ without loss of generality. Suppose that the data matrix $\data$ has rank $r$. We may express $\data = \vec{U}\bm{\Sigma} \vec{V}^T$ using its Singular Value Decomposition in compact form, where $\vec{U}\in\real^{n\times r}, \bm{\Sigma}\in \real^{r\times r}, \vec{V}^T \in \real^{r\times d}$. For any vector $z\in\real^d$ we have $\data \vec{z} = \vec{U}\bm{\Sigma} \vec{V}^T\vec{z}=\vec{U}\vec{z}^\prime$  for some $\vec{z}^\prime \in \real^r$.  Therefore, the number of distinct sign patterns $\mbox{sign}(\data \vec{z})$ for all possible $\vec{z}\in \real^d$ is equal to the number of distinct sign patterns $\mbox{sign}(\vec{U}\vec{z}^\prime)$ for all possible $\vec{z}^\prime \in \real^r$.

Consider an arrangement of $n$ hyperplanes $\in \real^r$, where $n \ge r$. Let us denote the number of regions in this arrangement by $P_{n,r}$. In \cite{ojha2000enumeration,cover1965geometrical} it's shown that this number satisfies
\begin{align*}
    P_{n,r} \le 2\sum_{k=0}^{r-1}{n-1 \choose k}\,.
\end{align*}
For hyperplanes in general position, the above inequality is in fact an equality.
In \cite{edelsbrunner1986constructing}, the authors present an algorithm that enumerates all possible hyperplane arrangements $\mathcal{O}(n^r)$ time, which can be used to construct the data for the convex program \eqref{eq:twolayerconvexprogram}.
%

\section{Dual problem for \eqref{eq:2layer_regularized_cost_l1}}\label{sec:twolayer_dualform_appendix}
The following lemma proves the dual form of \eqref{eq:2layer_regularized_cost_l1}.
\begin{lem}\label{lem:twolayer_dual}
The dual form of the following primal problem
\begin{align*}
     \min_{\firstw_j \in \ball_2} \min_{\{\secondw_j\}_{j=1}^m} \mathcal{L}\left(\sum_{j=1}^m \act(\data \firstw_j ) \secondw_j,\labelvec \right) +\beta \sum_{j=1}^m \vert \secondw_j \vert\,,
\end{align*}
is given by the following
\begin{align*}
     \min_{\firstw_j \in \ball_2} \max_{ \substack{\dual \in \real^n \,\mbox{\scriptsize s.t.} \\ \vert \dual^T\act(\data \firstw_j)\vert \le \beta  }} -\mathcal{L}^*(\dual)\,.
\end{align*}
\end{lem}
[\textbf{Proof of Lemma \ref{lem:twolayer_dual}}]
Let us first reparametrize the primal problem as follows
\begin{align*}
     \min_{\firstw_j \in \ball_2} \min_{\vec{r},\secondw_j} \mathcal{L}(\vec{r},\labelvec) +\beta \sum_{j=1}^m \vert \secondw_j \vert\, \text{ s.t. } \vec{r}=\sum_{j=1}^m \act(\data \firstw_j ) \secondw_j  ,
\end{align*}
which has the following Lagrangian
\begin{align*}
    L(\dual,\vec{r},\firstw_j,\secondw_j)=\mathcal{L}(\vec{r},\labelvec) +\beta \sum_{j=1}^m \vert \secondw_j \vert-\dual^T\vec{r} +\dual^T\sum_{j=1}^m \act(\data \firstw_j ) \secondw_j.
\end{align*}
Then, minimizing over $\vec{r}$ and $\secondwvec$ yields the proposed dual form.

%

\section{Dual problem for \eqref{eq:SDP} }\label{sec:sdp_appendix}
Let us first reparameterize the primal problem as follows
\begin{align*}
    &\max_{\vec{M},\dual} -\mathcal{L}^*(\dual)\mbox{ s.t.  } \sigma_{\max}\left(\vec{M}\right)\le \beta, \; \vec{M}=[\data_1^T \dual\, ...\, \data_K^T \dual ].
\end{align*}
Then the Lagrangian is as follows
\begin{align*}
    L(\lambda,\vec{Z},\vec{M},\dual)&=-\mathcal{L}^*(\dual)+\lambda\left(\beta-\sigma_{\max}\left(\vec{M}\right)\right)+\bold{tr}(\vec{Z}^T \vec{M})-\bold{tr}(\vec{Z}^T[\data_1^T \dual \ldots \data_K^T \dual ])\\
    &=-\mathcal{L}^*(\dual)+\lambda\left(\beta-\sigma_{\max}\left(\vec{M}\right)\right)+\bold{tr}(\vec{Z}^T \vec{M})-\dual^T \sum_{k=1}^K \data_k\vec{z}_k,
\end{align*}
where $\lambda \geq 0$ and $\bold{tr}$ denotes the trace operation. Then maximizing over $\vec{M}$ and $\dual$ yields the following dual form
\begin{align*}
    &\min_{\vec{z}_k\in\real^d ,\forall k \in [K] } \mathcal{L}\left(\sum_{k=1}^K \data_k\vec{z}_k,\labelvec\right)+\beta\Big \|[\vec{z}_1,\ldots,\vec{z}_K] \Big \|_{*},
\end{align*}
where $\Big\|[\vec{z}_1,\ldots,\vec{z}_K] \Big \|_{*}=\|\vec{Z}\|_*=\sum_i \sigma_i(\vec{Z})$ is the $\ell_1$-norm of singular values, i.e., nuclear norm \cite{recht2010guaranteed}.  
%


\section{Dual problem for \eqref{eq:linear_cnn} }\label{sec:circular_linear_cnn_appendix}
Let us denote the eigenvalue decomposition of $\firstwmat_j$ as $\firstwmat
_j=\vec{F} \diag_j \vec{F}^H$, where $\vec{F} \in \mathbb{C}^{d \times d}$ is the Discrete Fourier Transform matrix and $\diag_j\in \mathbb{C}^{d \times d}$ is a diagonal matrix. Then, applying the scaling in Lemma \ref{lemma:scaling} and then taking the dual as in Lemma \ref{lem:twolayer_dual} yields
\begin{align*}
    \max_\dual -\mathcal{L}^*(\dual) \text{ s.t. } \|\dual^T \data \vec{F} \diag \vec{F}^H \|_2 \leq \beta,\; \forall \diag: \|\diag \|_F^2 \leq d,
\end{align*}
which can be equivalently written as
\begin{align*}
    \max_\dual -\mathcal{L}^*(\dual)  \text{ s.t. } \|\dual^T \hat{\data} \diag \|_2 \leq \beta,\; \forall \diag: \|\diag \|_F^2 \leq d .
\end{align*}
Since $\diag$ is diagonal, $\|\diag \|_F^2 \leq d$ is equivalent to $\sum_{i=1}^d D_{ii}^2 \leq 1$. Therefore, the problem above can be further simplified as
\begin{align*}
    \max_\dual -\mathcal{L}^*(\dual)  \text{ s.t. } \|\dual^T \hat{\data}\|_{\infty} \leq \frac{\beta}{\sqrt{d}}\;  .
\end{align*}
Then, taking the dual of this problem gives the following
\begin{align*}
    \min_{\vec{z} \in \mathbb{C}^d} \mathcal{L}\left(\hat{\data} \vec{z}
,\labelvec \right)+ \frac{\beta}{\sqrt{d}} \|\vec{z}\|_1.
\end{align*}



\section{Semi-infinite strong duality \label{appendix_semi_infinite_duality}}
Note that the semi-infinite problem \eqref{eq:2layer_regularized_cost_innerdual} is convex. We first show that the optimal value is finite. For $\beta>0$, it is clear that $\dual=0$ is strictly feasible, and achieves $0$ objective value. Note that the optimal value $p^*$ satisfies $p^*\le \|\labelvec\|_2^2$ since this value is achieved when all the parameters are zero.
Consequently, Theorem 2.2 of \cite{shapiro2009semi} implies that strong duality holds, i.e., $p^*=d^*_{\infty}$, if the solution set of the semi-infinite problem in \eqref{eq:2layer_regularized_cost_innerdual} is nonempty and bounded. Next, we note that the solution set of \eqref{eq:2layer_regularized_cost_innerdual} is the Euclidean projection of $\labelvec$ onto the polar set $(\rectset \cup -\rectset)^\circ$ which is a convex, closed and bounded set since $\act(\data\firstw)$ can be expressed as the union of finitely many convex closed and bounded sets.
\qed

\section{Semi-infinite strong gauge duality \label{appendix_semi_infinite_duality_gauge}}
Now we prove strong duality for \eqref{eq:supportfun}. We invoke the semi-infinite optimality conditions for the dual \eqref{eq:supportfun}, in particular we apply Theorem 7.2 of \cite{semiinfinite_goberna} and use the standard notation therein. We first define the set
\begin{align*}
    \mathbf{K}=\mathbf{cone}\left\{ \left( \begin{array}{c}s\, \act(\data \firstw) \\ 1 \end{array}  \right), \firstw \in \ball_2, s\in\{-1,+1\}; \left(\begin{array}{c} 0 \\ -1\end{array}\right) \right\}\,.
\end{align*}
Note that $\mathbf{K}$ is the union of finitely many convex closed sets, since $\act(\data\firstw)$ can be expressed as the union of finitely many convex closed sets. Therefore the set $\mathbf{K}$ is closed. By Theorem 5.3 of \cite{semiinfinite_goberna}, this implies that the set of constraints in \eqref{eq:dual} forms a Farkas-Minkowski system. By Theorem 8.4 of \cite{semiinfinite_goberna}, primal and dual values are equal, given that the system is consistent. Moreover, the system is discretizable, i.e., there exists a sequence of problems with finitely many constraints whose optimal values approach to the optimal value of \eqref{eq:dual}.
\qed

\section{Neural Gauge function and equivalence to minimum-norm networks}\label{sec:gauge_appendix}
Consider the gauge function
\begin{align*}
    p^g=&\min_{r\ge0}\, r \mbox{ s.t. } r \labelvec \in \mbox{conv}(\rectset\cup-\rectset)
\end{align*}
and its dual representation in terms of the support function of the polar of $\mbox{conv}(\rectset\cup-\rectset)$
\begin{align*}
    d^g=\max_{\dual} \dual^T \labelvec  \mbox{ s.t. } \dual\in (\rectset\cup-\rectset)^\circ.
\end{align*}
Since the set $\rectset\cup-\rectset$ is a closed convex set that contains the origin, we have $p^g=d^g$ \cite{Rockafellar} and $\left(\mbox{conv}(\rectset\cup-\rectset)\right)^\circ = (\rectset\cup-\rectset)^\circ$. The result in Section \ref{appendix_semi_infinite_duality} implies that the above value is equal to the semi-infinite dual value, i.e., $p^d=p^g_{\infty}$, where 
\begin{align*}
    p^g_{\infty} := \min_{\bm{\mu}}\, \|\bm{\mu}\|_{TV}\mbox{ s.t. } \int_{\firstw \in \ball_2} \act(\data \firstw)d\bm{\mu}(\firstw)=\labelvec\,.
\end{align*}
By Caratheodory's theorem, there exists optimal solutions the above problem consisting of $m^*$ Dirac deltas \cite{Rockafellar,rosset2007}, and therefore
\begin{align*}
    p^g_{\infty} = \min_{{\firstw_j \in \ball_2},\secondw_j}\, \sum_{j=1}^{m^*} \vert\secondw_j\vert \mbox{ s.t. } \sum_{j=1}^{m^*} \act(\data \firstw_j)\secondw_j=\labelvec\,,
\end{align*}
where we define $m^*$ as the number of Dirac delta's in the optimal solution to $p^g_{\infty}$. If the optimizer is non-unique, we define $m^*$ as the minimum cardinality solution among the set of optimal solutions.
Now consider the non-convex problem
\begin{align*}
    &\min_{\{\firstw_j,\secondw_j\}_{j=1}^m} \|\secondwvec\|_1 \mbox{ s.t. } \sum_{j=1}^m \act(\data \firstw_j) \secondw_j =\labelvec,\;\firstw_j \in \ball_2\,.
\end{align*}
Using the standard parameterization for $\ell_1$-norm we get
\begin{align*}
    &\min_{\substack{\{\firstw_j\}_{j=1}^m \\ \vec{s}\ge0 \\ \vec{t}\ge0}} \sum_{j=1}^m (t_j + s_j) \mbox{ s.t.} \sum_{j=1}^m \act(\data \firstw_j) t_j - \act(\data \firstw_j)s_j=\labelvec,\; \firstw_j \in \ball_2\,,\forall j.
\end{align*}
Introducing a slack variable $r\in\reals_+$, an equivalent representation can be written as
\begin{align*}
    &\min_{\substack{\{\firstw_j\}_{j=1}^m\\\vec{s},\vec{t},r\ge0}}\, r \mbox{ s.t. } \sum_{j=1}^m \act(\data \firstw_j) t_j - \act(\data \firstw_j)s_j=\labelvec,\; \sum_{j=1}^m (t_j + s_j )= r,\;\firstw_j \in \ball_2\,,\forall j.
\end{align*}
Note that $r>0$ as long as $\labelvec\neq \vec{0}$. Rescaling variables by letting $t_j^\prime = t_j/r$, $s_j^\prime = s_j/r$ in the above program, we obtain
%
\begin{align}
    &\min_{\substack{\{\firstw_j\}_{j=1}^m \\\vec{s}^\prime, \vec{t}^\prime,r\ge0}} r \label{eq:gaugenonconvex} \mbox{ s.t.} \sum_{j=1}^m \left( \act(\data \firstw_j) t^\prime_j - \act(\data \firstw_j)s^\prime_j\right)=r\labelvec,\; \sum_{j=1}^m (t_j^\prime+ s_j^\prime) = 1,\firstw_j \in \ball_2,\forall j \nonumber\,.
\end{align}
Suppose that $m\ge m^*$. It holds that
\begin{align}
    & \exists \vec{s}^\prime,\vec{t}^\prime\ge0\, ,\{\firstw_j\}_{j=1}^m \mbox{ s.t. }  \sum_{j=1}^m (t_j^\prime+ s_j^\prime) = 1,\,\|\firstw_j\|_2\le 1,\,\forall j,\,\sum_{j=1}^m (\data \firstw_j)t^\prime_j - \act(\data \firstw_j)s^\prime_j=r\labelvec \, \nonumber \\
    &\iff r \labelvec \in \mbox{conv}(\rectset\cup-\rectset).
\end{align}
We conclude that the optimal value of \eqref{eq:gaugenonconvex} is identical to the gauge function $p_g$.

\section{Alternative proof of the semi-infinite strong duality \label{appendix_semi_infinite_duality2}}
It holds that $p^* \ge d^*$ by weak duality in \eqref{eq:2layer_regularized_cost_innerdual}.
Theorem \ref{theo:mainconvex} proves that the objective value of \eqref{eq:dual} is identical to the value of \eqref{eq:twolayer_objective_generic} as long as $m\ge m^*$. Therefore we have $p^*=d^*$.
\qed

\section{Finite dimensional strong duality results for Theorem \ref{theo:mainconvex}}
\label{sec:finitestrongdualconst}
\begin{lem}
\label{lem:finitestrongdualconst}
Suppose $\diag(S)$, $\hat{\diag}(S)$, $\hat{\diag}(S^c)$ are fixed diagonal matrices as described in the proof of Theorem \ref{theo:mainconvex}, and $\data$ is a fixed matrix. The dual of the convex optimization problem
\begin{align*}
&\max_{\substack{\firstw \in \ball_2\\ \hat{\diag}(S)\data \firstw \ge 0 \\ (\vec{I}_n-\hat{\diag}(S^c)\data )\firstw \le 0 }}
\dual^T\diag(S) \data \firstw 
\end{align*}
is given by
\begin{align*}
&\qquad \min_{\substack{\bm{\alpha},\bm{\gamma} \in \real^{n } \\ \bm{\alpha},\bm{\gamma} \ge 0}}
\| \data^T \diag(S) \dual + \data^T \hat{\diag}(S) (\bm{\alpha} + \bm{\gamma} )-\data^T \bm{\gamma} \|_2\,
\end{align*}
and strong duality holds.
\end{lem}
Note that the linear inequality constraints specify valid hyperplane arrangements. Then there exists strictly feasible points in the constraints of the maximization problem. Standard finite second order cone programming duality implies that strong duality holds \cite{boyd_convex} and the dual is as specified.
\qed


\end{document}